\documentclass{article} 
\usepackage{iclr2027_conference,times}

\usepackage{amsmath,amsfonts,bm}

\def\eqref#1{equation~\ref{#1}}

\def\1{\bm{1}}

\DeclareMathAlphabet{\mathsfit}{\encodingdefault}{\sfdefault}{m}{sl}
\SetMathAlphabet{\mathsfit}{bold}{\encodingdefault}{\sfdefault}{bx}{n}

\usepackage{hyperref}
\usepackage{url}
\usepackage{graphicx}
\usepackage{booktabs}
\usepackage{tabularx}
\usepackage{array}
\usepackage{amssymb}
\usepackage[table]{xcolor}
\usepackage{multirow}
\usepackage{xcolor}
\usepackage{enumitem}
\usepackage{rotating}
\usepackage{longtable}
\usepackage{pdflscape}
\usepackage{listings}
\usepackage{adjustbox}
\usepackage[most]{tcolorbox}
\usepackage{caption}

\definecolor{benchblue}{HTML}{335AB4}
\definecolor{benchgray}{HTML}{777777}
\definecolor{benchrow}{HTML}{EEF3FA}

\newcommand{\eg}{\emph{e.g.}}
\newcommand{\ie}{\emph{i.e.}}
\newcommand{\etc}{\emph{etc.}}

\newtcolorbox{promptbox}[1]{
    breakable,
    colback=gray!5,
    colframe=gray!40,
    boxrule=0.4pt,
    arc=2pt,
    left=7pt,
    right=7pt,
    top=5pt,
    bottom=5pt,
    fonttitle=\small\bfseries,
    fontupper=\small,
    coltitle=black,
    colbacktitle=gray!10,
    title={#1},
    before skip=8pt,
    after skip=8pt
}

\title{MoI-Bench: A Benchmark for Image Generation and Editing on Mobile Devices}

\author{
    \textbf{Kailai Feng\textsuperscript{1,*},
    Haoxiang Rao\textsuperscript{2,*},
    Yuxiang Wei\textsuperscript{3,\dag},
    Bo Chen}\\
    \textbf{Junsong Li\textsuperscript{4},
    Yang Pan,
    Chengkun Cai\textsuperscript{1},
    Jagmohan Chauhan\textsuperscript{1,\dag}}\\[6pt]
    \textsuperscript{1}University College London
    \quad
    \textsuperscript{2}Nanjing University\\
    \textsuperscript{3}The Hong Kong Polytechnic University
    \quad
    \textsuperscript{4}East China Normal University\\[4pt]
    \textsuperscript{*}Equal contribution.
    \quad
    \textsuperscript{\dag}Co-corresponding authors.\\[6pt]
    Project Page: \url{https://carlofkl.github.io/MoI-Bench-project/}
}

\iclrfinalcopy 

\begin{document}

\maketitle

\begin{abstract}
Developing on-device image models requires considering practical user needs, task types and device resource constraints.
However, existing benchmarks provide limited guidance on identifying capability gaps between on-device models and server-side models.
We therefore introduce \textbf{MoI-Bench}, a benchmark grounded in mobile user needs, comprising 64 subtasks and 1150 samples.
It defines 16 \textbf{\textit{operations}}, such as removal and replacement, and eight visual \textbf{\textit{target}} categories covering content such as portraits and text.
These form two orthogonal axes whose applicable combinations define specific tasks, such as removing facial makeup.
To better guide model improvement, we further propose \textbf{MoI-Eval}, which introduces task-specific \textbf{\textit{criteria}} covering instruction following, visual quality and detail preservation.
Based on these criteria, we construct sample-level evaluation questions and aggregate their scores into overall quality and capability profiles along the \textit{operation}, \textit{target} and \textit{criteria} axes.
We evaluate representative on-device and server-side models and measure mobile deployment costs.
The results show that this decoupled \textit{Operation $\times$ Target} design provides finer-grained guidance for improving compact models.
For example, it highlights the need to prioritize specific capability gaps (\eg, text-replacement) over broad application-level goals (\eg, improving text editing).
Unlike conventional benchmarks that rank models using a single score, we introduce a two-dimensional \textit{Quality-Deployability} capability map to assess models for practical mobile use.
The map shows that current on-device models can achieve efficient generation or editing but expanding their task coverage deserves greater attention than further improving deployment efficiency.
Further progress requires improving editing quality, particularly identity preservation and text rendering, while broadening task coverage.
When resource constraints limit the capabilities achievable on-device, task-aware on-device--cloud routing provides a complementary direction for future research.
Our benchmark data and inference and evaluation toolkits are available \href{https://github.com/Carlofkl/MoI-Bench}{here}.

\end{abstract}

\section{INTRODUCTION}
\label{sec:intro}

\begin{figure}
    \centering
    \includegraphics[width=\linewidth]{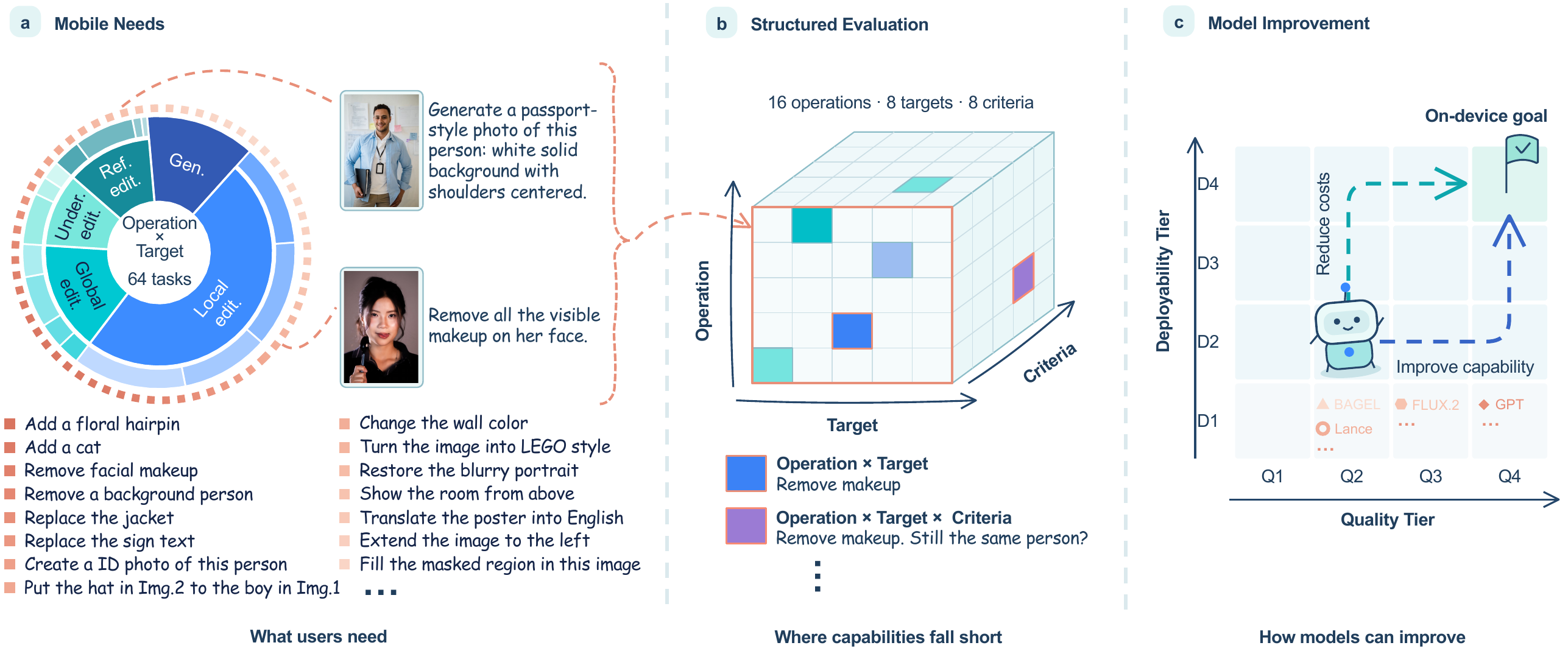}
    \caption{Overview of our benchmark and evaluation framework: (a) tasks grounded in mobile user needs; (b) structured capability evaluation along operation, target and criteria axes; and (c) model improvement guided by the Quality-Deployability map.}
    \label{fig:intro}
    \vspace{-10pt}
\end{figure}

Advances in large diffusion models have improved image generation and editing capabilities~\cite{cai-etal-2025-bayesian,feng2026vitaglyph,liu2026instancecontrol}, with increasing use in mobile photo-editing applications.
Cloud-based services such as GPT Image 2.5~\cite{openai2026gptimage25} and Seedream~5.0~\cite{bytedance2026seedream5} bring substantial inference costs and raise privacy concerns over personal image uploads.
These concerns have motivated the development of compact on-device models, such as SnapGen, Mobile-O, DreamLite and so on~\cite{shaker2026mobile,duan2026moebius,deng2026blazeedit,chen2025snapgen}.
Evaluating these models requires determining which mobile tasks they support, how well their outputs satisfy user requirements, and whether inference fits device resource budgets.
However, existing benchmarks such as GenEval~\cite{ghosh2023geneval} and ImgEdit-Bench~\cite{ye2026imgedit} primarily assess general generation and editing capabilities.
They provide limited guidance on connecting mobile application needs, task performance and on-device deployment costs.
This makes it difficult for developers to select models for specific requests and for researchers to identify capability gaps relative to larger models.

A request such as removing text from a picture specifies both an action and the visual content it concerns.
We define the requested action, such as removal or replacement, as the \textit{\textbf{operation}}, and the type of visual content to generate or modify as the \textit{\textbf{target}}.
Integrating more general application categories such as \texttt{Text Editing} can extend coverage of mobile scenarios but does not always make the task coverage explicit.
For example, \texttt{Text Editing} can include both text removal and text replacement, so a high category score of \texttt{Text Editing} does not establish proficiency in both operations.
Conversely, removing text and removing a person share the same operation applied to different targets.
Grouping scores only by application category makes it difficult to compare removal capability across targets and between on-device and server-side models.
We therefore introduce \textbf{MoI-Bench}, a \textbf{Mo}bile \textbf{I}mage generation and editing \textbf{Benchmark} grounded in a user survey of mobile application needs.
Fig.~\ref{fig:intro} illustrates how we connect these needs to structured capability evaluation and model improvement.
We define most operations following prior works~\cite{jiang2026geditbench, jia2026compbench}, then use a user survey to identify priority target categories for mobile applications.
It defines 16 operations and eight target categories as two orthogonal axes of an \textit{Operation $\times$ Target} Task Matrix.
Applicable combinations are manually reviewed, yielding 64 subtasks comprising 1150 samples.
This organization supports comparisons between operations for the same target and between targets for the same operation, helping developers assess the requests relevant to their applications.

Task performance alone does not explain which output requirements a model fails to satisfy.
When asked to remove a specified person from a group portrait, a model may remove the wrong person, alter the remaining people's identities or leave visible artifacts.
Then we propose \textbf{MoI-Eval}, which associates each subtask with applicable criteria (\ie, instruction following, visual quality and detail preservation) in a sparse \textit{Operation $\times$ Target $\times$ Criteria} Rubric Cube.
Based on these criteria, we develop detailed questions for each sample to evaluate models' outputs and aggregate their scores into overall quality $Q$ and capability profiles along \textit{operation}, \textit{target} and \textit{criteria}.
Joint \textit{Operation $\times$ Target} profiles show performance on specific requests, while criterion profiles help researchers identify whether improvements are needed in instruction following, visual quality or preservation of existing details.

For mobile applications, output quality must also be considered alongside deployment costs.
Even high-quality models are unsuitable for on-device use if their memory requirements exceed device capacity.
We therefore retain quality $Q$ and deployability $D$ as separate dimensions in a \textit{Quality-Deployability} capability map.
The quality dimension considers the overall score $Q$, task coverage and criterion-group scores, while the deployability dimension considers end-to-end latency, peak memory and deployed model size.
Based on these measurements, we define quality tiers and deployability tiers and use them to construct a \textit{Quality-Deployability} capability map.
The map positions models by capability and resource requirements, helping developers select suitable models and researchers identify priorities for broader task coverage, higher output quality or lower deployment costs.

We evaluate 19 generative models and measure text-to-image deployment costs on iOS and Android devices.
Our results first show that MoI-Bench provides finer-grained and clearer guidance for model improvement.
For example, Ovis-U1~\cite{wang2025ovis} achieves comparable performance with FLUX.2 [klein]~\cite{blackforestlabs2026flux2klein} on text removal but falls far behind on text replacement.
This challenges the assumption that compact models perform poorly across all text-related tasks and suggests that improving specific tasks (\eg, text replacement) should take priority over text editing in general.
This distinction also helps users select models for their intended purposes.
Criterion profiles further distinguish on-device models' development priorities: DreamLite~\cite{feng2026dreamlite} shows weaknesses in visual integrity and coherence, whereas Ovis-U1 has a lower identity-preservation score.
Additionally, our findings suggest that broadening task coverage deserves greater attention than further optimizing mobile deployment efficiency.
Models such as BlazeEdit~\cite{deng2026blazeedit} and Moebius~\cite{duan2026moebius} support only a single editing task, while even more general editing models such as DreamLite~\cite{feng2026dreamlite} lack support for multiple input images.
Mobile-O and DreamLite both reach D3 on the iPhone~17~Pro, but their reference quality tiers are Q1 and Q2, respectively.
Further progress therefore requires improving editing quality and expanding task coverage while preserving resource efficiency.
The task-dependent capability gaps also motivate exploring on-device--cloud routing, with requests assigned according to model capability and deployment constraints.

In summary, our contributions are:
\begin{itemize}

    \item We introduce \textbf{MoI-Bench}, a benchmark for on-device image models, built on an \textit{Operation $\times$ Target} Task Matrix. Its 1,150 samples span 64 subtasks designed around mobile application needs, covering image generation and four types of editing operations.

    \item We propose \textbf{MoI-Eval}, which derives overall quality $Q$ and capability profiles along the operation, target, and criteria axes from a sparse \textit{Operation $\times$ Target $\times$ Criteria} Rubric Cube. It further introduces, to the best of our knowledge, the first unified \textit{Quality-Deployability} capability map for benchmarking models instead of a single score.

    \item We benchmark representative generative models and further examine \textit{Quality-Deployability} trade-offs for on-device models through measurements on iOS and Android devices. We release the license-audited benchmark annotations, inference code and evaluation toolkit.
    
\end{itemize}

\section{Related Work}
\paragraph{On-device Diffusion Models.}
Mobile diffusion models face constraints in storage, memory, computation and operator support.
Existing work addresses these constraints through graph optimization and compression~\cite{chen2023speed,choi2023squeezing}, pruning and distillation~\cite{li2023snapfusion,zhao2024mobilediffusion,castells2024edgefusion,chen2025snapgen,chavhan2026nanoflux,hu2026snapgen++,du2026elasticdit, feng2024multi}, and low-bit quantization~\cite{zhao2024mixdq,li2024svdquant}.
Recent systems improve generation efficiency and extend capabilities to understanding and editing~\cite{shaker2026mobile,feng2026dreamlite,deng2026blazeedit,kodavanti2026edgedit,duan2026moebius}.
These advances motivate a unified evaluation of model quality and mobile deployability.

\paragraph{Benchmarks for image generation and editing.}
Generation benchmarks assess prompt alignment, compositional reasoning and visual correctness~\cite{saharia2022photorealistic,hu2023tifa,huang2025t2i,ghosh2023geneval,bakr2023hrs,lee2023holistic,li2024genai,wu2024conceptmix,wang2025unigenbench++}.
Editing benchmarks evaluate instruction following and content preservation across increasingly diverse tasks~\cite{zhang2023magicbrush,ju2023direct,ma2024i2ebench,qian2025gie,liu2025step1x,jia2026compbench,wu2026kris,ye2026imgedit,yi2026vde,jiang2026geditbench}.
Unified suites cover generation, editing and reference-conditioned tasks~\cite{ku2024imagenhub,pan2025ice,hua2025mmigbench,sani2026imagenworld}.
MoI-Bench organizes tasks by operation and target for task-specific analysis and, to our knowledge, is the first benchmark to jointly evaluate the quality and deployability of on-device image diffusion models.

\section{MoI-Bench}
\label{sec:bmk}

\begin{table*}[t]

\centering
\caption{
The \textit{Operation $\times$ Target} Task Matrix of MoI-Bench.
Each cell describes the corresponding task.
Merged cells indicate tasks shared across target categories;
N/A marks combinations excluded from the benchmark.
}
\label{tab:moge_bench}
\includegraphics[width=\textwidth]{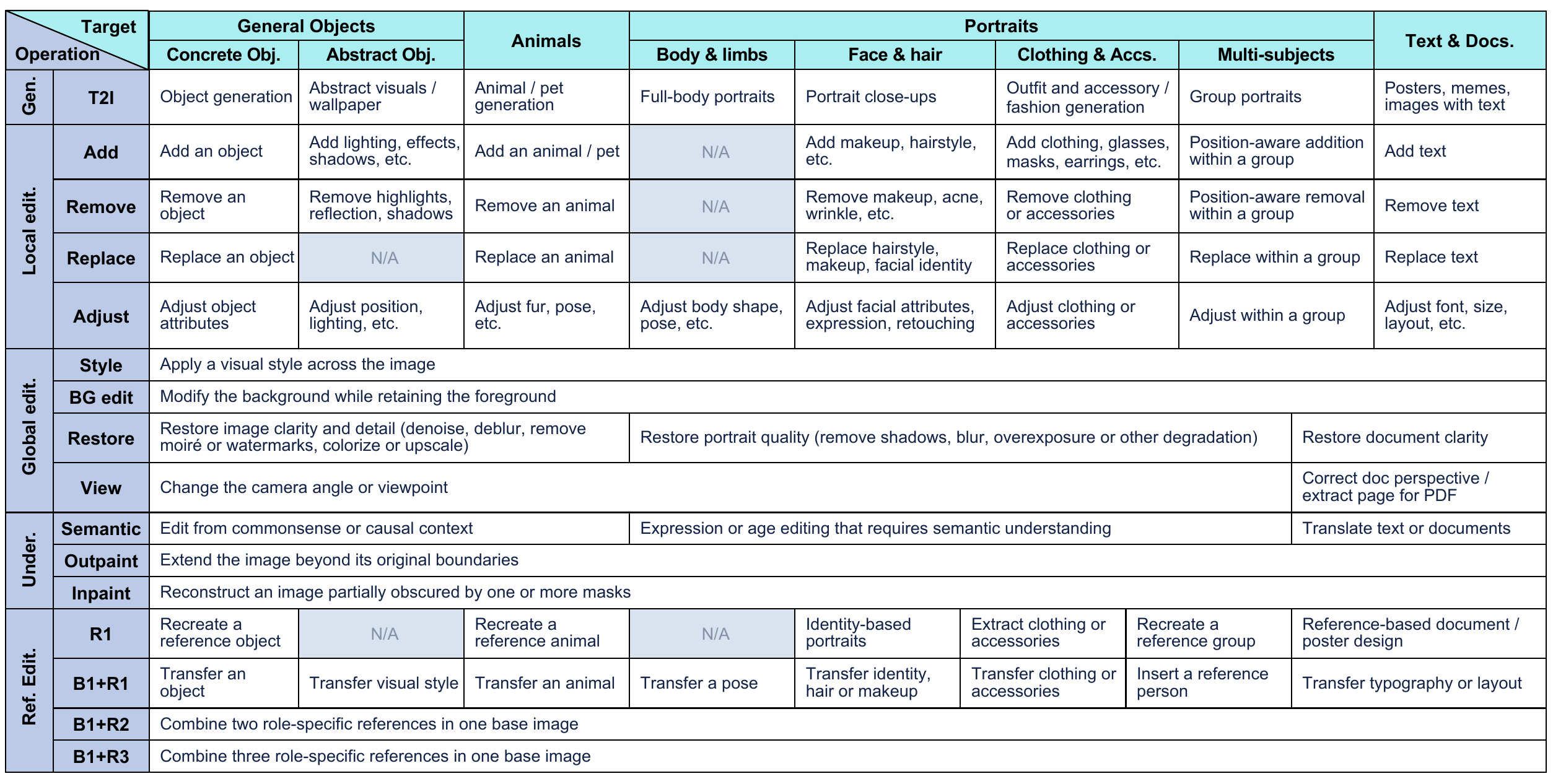}
\end{table*}

MoI-Bench organizes tasks through an \textit{Operation $\times$ Target} Task Matrix.
Operations specify what the model should do, and targets identify the visual content to generate or modify.
Applicable combinations define subtasks of the form ``\texttt{Operation} on \texttt{Target}''.

To better align the benchmark with real-world mobile user needs, we survey 1,000 participants to guide task scope and target coverage.
As shown in Fig.~\ref{fig:survey_bmk_sta}(a), portrait retouching, object removal, photo restoration and text editing are the most common desired uses.
These preferences motivate an emphasis on image editing, while text-to-image generation is retained to assess models' basic image-generation capabilities.
The reported use cases guide our selection of targets and operations, whose applicable combinations define subtasks such as blemish adjustment and photobomb removal.
User priorities motivate criteria for instruction following, identity preservation and visual quality.
We refine these requirements into task-specific criteria and sample-level evaluation questions.
Preferences for quick results, low storage use and avoiding device slowdown further motivate our deployment measurements of latency, model size and peak memory.
Survey details are provided in the Appendix~\ref{app:survey}.

\subsection{Operation × Target Task Matrix}


\paragraph{Operation taxonomy.} 
Following prior taxonomies~\cite{yu2025anyedit,ye2026imgedit}, we organize 16 operation units into five groups (Tab.~\ref{tab:moge_bench}):

\begin{itemize}
    \item \textit{Generation.} Creating an image from a prompt without an input image. 
    Tasks specify the desired subjects, attributes, composition and style.
    Generation conditioned on a reference image is assigned to Reference Editing.
    
    \item \textit{Local Editing.} 
    Adding, removing, replacing or adjusting specified content while preserving unrelated regions.
    Addition introduces new content, while removal deletes selected content and reconstructs the exposed region.
    Replacement substitutes content, such as replacing a hairstyle or a word.
    Adjustment changes attributes of existing content, such as color, size, position or pose.
    
    \item \textit{Global Editing.} 
    Changing overall appearance or scene context through stylization, background modification, restoration or camera-view changes.
    Stylization applies a visual style across the image, while background modification changes the surroundings and preserves the foreground subject.
    Restoration addresses degradation such as blur, noise or overexposure, including localized defects.
    Camera-view changes alter the viewing perspective rather than individual object positions.
    
    \item \textit{Understanding Editing.} 
    Inferring the required change or completing content from semantic and visual context.
    Semantic editing requires interpreting an intended outcome rather than an explicitly prescribed visual change, such as showing what happens to ice cream left in the sun.
    Inpainting reconstructs missing content within a supplied mask using the surrounding image.
    Outpainting extends content beyond the original image boundaries while maintaining scene consistency.

    \item \textit{Reference Editing.} 
    Using reference images to specify desired content or attributes, such as identity, pose, clothing or style.
    Following prior work~\cite{oshima2026multibanana}, we organize these tasks by input configuration.
    A single reference without a base image guides the generation of a new image.
    A base image with one reference supports transferring specified content or attributes into the base image.
    Configurations with two or three references combine information from multiple sources, with each reference assigned a role in the instruction.

\end{itemize}

\paragraph{Target taxonomy.}
To reflect mobile application needs, we include Animals, Portraits, and Text \& Documents alongside General Objects.
Targets are categorized by the primary visual content to generate or modify, rather than every entity present in the image.
These four groups contain eight target categories.

\begin{itemize}
    \item \textit{General Objects.} 
    This group contains Concrete Objects and Abstract Objects.
    Concrete Objects include vehicles, furniture and everyday items.
    Abstract Objects cover visual effects and relations, such as lighting, shadows and reflections.
    
    \item \textit{Animals.}
    This category covers animals, including pets, their appearance and their accessories.
    Tasks include generating animals, modifying their pose or fur, and adding or removing accessories such as collars.
    
    \item \textit{Portraits.} 
    This group contains four categories: Body \& Limbs, Face \& Hair, Clothing \& Accessories, and Multi-Subjects.
    Body \& Limbs covers body shape, limb configuration and pose.
    Face \& Hair covers facial features, expressions, makeup, hairstyles and identity.
    Clothing \& Accessories covers garments and worn items, such as coats, glasses and jewelry.
    Multi-Subjects covers groups of people, including group-level changes and edits to selected individuals within a group.
    Examples include adding or removing a person and changing the pose of a specified individual while preserving the others.

    \item \textit{Text \& Documents.} 
    This category covers text content and its visual presentation in documents, posters, signs and other text-bearing images.
    Tasks include adding, removing, replacing or translating text, as well as modifying typography and layout.
    It also includes restoring document clarity and correcting page perspective.

\end{itemize}

\paragraph{Task coverage.}
We construct subtasks by pairing operations with target categories.
For example, \texttt{Replacement} $\times$ \texttt{Text \& Documents} replaces specified text while retaining the surrounding content and layout.
We also manually review the applicability of each \textit{operation $\times$ target} combination.
For example, \texttt{Addition} $\times$ \texttt{Body \& Limbs} is marked N/A because adding anatomical parts falls outside our portrait-editing scope.
Operations are adapted to practical target-specific needs (\eg, camera-view editing becomes perspective rectification for documents rather than viewpoint changes in general scenes).
We also group target categories under a shared subtask when they follow the same task definition (\eg, stylization of the whole image).
Reference editing with two or three reference images also uses shared subtasks to combine information from multiple sources.
Each merged cell counts as one subtask, yielding 64 subtasks in total (Tab.~\ref{tab:moge_bench}).
Detailed task descriptions and examples are provided in the Appendix~\ref{app:tasks}.

\subsection{Dataset Construction and Quality Control}

\paragraph{Sample construction.}
Following ImgEdit-Bench~\cite{ye2026imgedit}, we construct samples through subtask-level allocation, image selection and instruction refinement.
Within each subtask, we first identify the specific transformations and target variations to cover and assign an initial sample quota.
For example, $(B1 + R1) \times$ \texttt{Face\&Hair} includes hairstyle, makeup and facial identity transfer.
We allocate approximately five samples to each transformation, giving 15 samples for this subtask.
These quotas guide image collection and are adjusted after reviewing image and instruction diversity

Subsequently, we retrieve input and reference images from Pexels and Wikimedia Commons.
Images are selected to provide the visual information needed for the requested task, such as visible facial features for makeup editing or readable text for text replacement.
For reference editing, we review the base and reference images together to ensure that the specified transfer is feasible and that each image has a clear role.
For inpainting and restoration, we apply predefined masks or degradations to the source images.
All evaluated models receive the same processed inputs.

Given the subtask and selected images, Gemini 3.1 Pro~\cite{gemini31pro2026} proposes instruction templates.
We manually select and adapt these templates to each sample, specifying the requested change, its target and any content that should remain unchanged.
For tasks with multiple input images, instructions also identify the role of each reference.
Text-to-image samples contain only prompts describing the desired image, without input images.
Two annotators cross-review each sample for instruction clarity, task feasibility and consistency with the input images.
They also check for duplicates and review visual and instruction diversity within each subtask.
The final subtask sample counts are provided in Tab.~\ref{tab:app-task-examples} in the Appendix~\ref{app:tasks}.

\paragraph{Quality control.}
Following prior work~\cite{ye2026imgedit,jiang2026geditbench,zhang2023magicbrush}, two annotators cross-review each sample for instruction clarity, task feasibility and image--instruction consistency.
They check whether the instruction clearly identifies the requested change and its target, and whether the input images provide the information needed to perform the task.
For reference editing, they inspect all inputs together and check that the role of each reference is clearly specified.
For text-to-image generation, they review the prompt for clarity and consistency with the assigned subtask.
We also check for duplicate samples and review visual and instruction diversity within each subtask.
This review informs instruction refinement and adjustments to sample allocation.

\subsection{Benchmark Statistics and Comparison}
\begin{figure}[t]
    \centering
    \includegraphics[width=\linewidth]{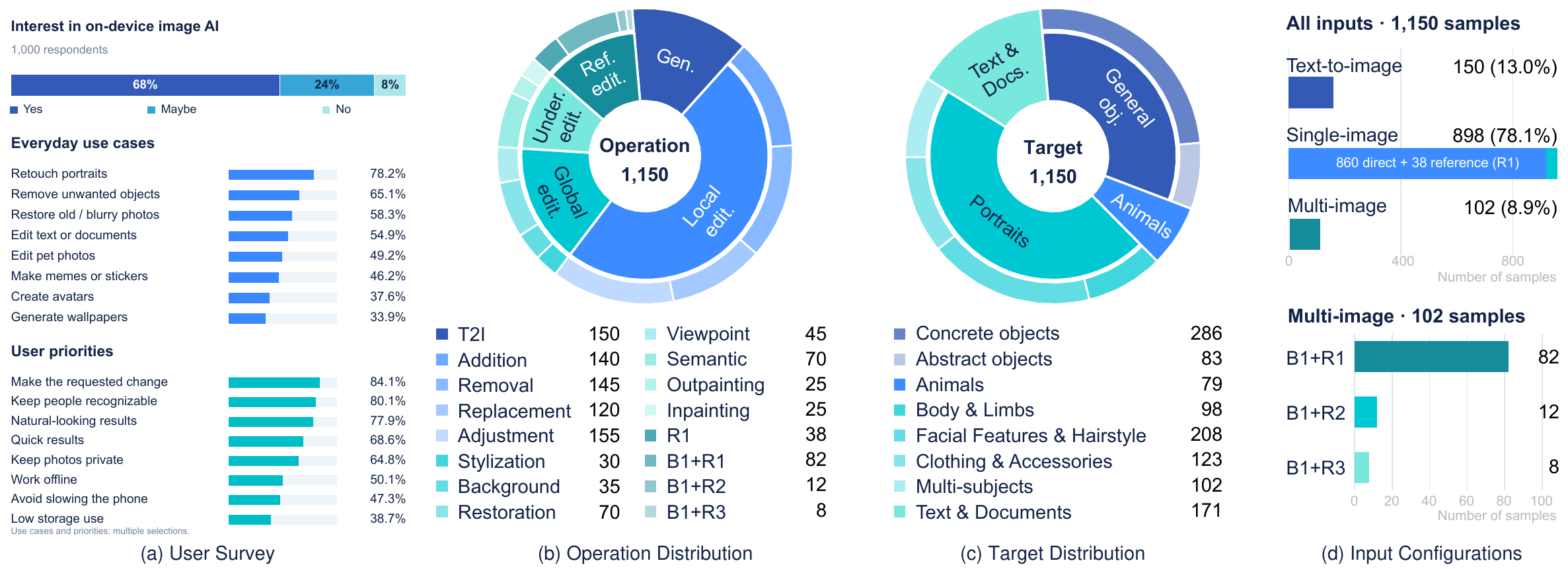}
    \caption{
    (a) Survey responses on interest in on-device image AI, everyday use cases and user priorities.
    (b--d) Sample distributions by operation, target and input configuration.
    Each sample is assigned to one target category according to the primary visual content to generate or modify.
    }
    \vspace{-10pt}
    \label{fig:survey_bmk_sta}
\end{figure}

\begin{table*}[t!]
\centering
\caption{
Task coverage across generation and editing benchmarks.
Except for T2I, task columns describe editing tasks and may overlap.
$\checkmark$: documented coverage;
$\times$: outside the stated scope.
Act., Pos., Impl. denote action, object-position or spatial-relation changes, implicit editing respectively;
Max. Refs counts reference images excluding the base image and MT denotes multi-turn editing.
$^{a}$ImgEdit-Bench includes training and evaluation prompts.
}
\label{tab:benchmark_comparison}

\small
\setlength{\tabcolsep}{1pt}
\renewcommand{\arraystretch}{1.0}

\def\Y{\ensuremath{\checkmark}}
\def\N{\ensuremath{\times}}
\def\U{---}

\begin{tabular}{@{}lr*{10}{c}@{}}
\toprule
\multirow{2}{*}{Benchmark}
& \multirow{2}{*}{Size}
& \multicolumn{8}{c}{Task coverage}
& \multirow{2}{*}{Max.~Refs}
& \multirow{2}{*}{MT} \\
\cmidrule(lr){3-10}
& &
T2I & Loc. & Glob. & Act. & Pos. & View & Impl. & Multi-obj.
& & \\
\midrule

\multicolumn{12}{@{}l}{\textit{Text-to-image benchmarks}} \\

GenEval~\cite{ghosh2023geneval}
& 553
& \Y & \N & \N & \N & \N & \N & \N & \N
& 0 & \N \\

GenAI-Bench~\cite{li2024genai}
& 1,600
& \Y & \N & \N & \N & \N & \N & \N & \N
& 0 & \N \\

Qwen-Image-Bench~\cite{li2026qwen}
& 1,000
& \Y & \N & \N & \N & \N & \N & \N & \N
& 0 & \N \\

\midrule
\multicolumn{12}{@{}l}{\textit{General image-editing benchmarks}} \\

AnyEdit-Test~\cite{yu2025anyedit}
& 1,250
& \N & \Y & \Y & \Y & \Y & \Y & \Y & \N
& 0 & \N \\

ImgEdit-Bench~\cite{ye2026imgedit}
& $811$
& \N & \Y & \Y & \Y & \N & \N & \Y & \Y
& 0 & \Y \\

GEdit-Bench~\cite{liu2025step1x}
& 606
& \N & \Y & \Y & \Y & \N & \N & \N & \N
& 0 & \N \\

KontextBench~\cite{labs2025flux}
& 1,026
& \N & \Y & \Y & \Y & \Y & \Y & \Y & \Y
& 1 & \N \\

GEditBench v2~\cite{jiang2026geditbench}
& 1,200
& \N & \Y & \Y & \Y & \Y & \Y & \Y & \Y
& 1 & \N \\

CompBench~\cite{jia2026compbench}
& ${>}3{,}000$
& \N & \Y & \Y & \Y & \Y & \Y & \Y & \Y
& 0 & \Y \\

Edit-Compass~\cite{bai2026edit}
& 2,388
& \N & \Y & \Y & \Y & \Y & \Y & \Y & \Y
& 2 & \N \\

\midrule
\multicolumn{12}{@{}l}{\textit{Reference-conditioned benchmarks}} \\

OmniContext~\cite{wu2026omnigen2}
& 400
& \N & \Y & \Y & \Y & \Y & \N & \N & \Y
& 2 & \N \\

MultiBanana~\cite{oshima2026multibanana}
& 3,769
& \N & \Y & \Y & \Y & \Y & \Y & \Y & \Y
& 8 & \N \\

MICON-Bench~\cite{wu2026micon}
& 1,043
& \N & \Y & \Y & \Y & \Y & \N & \Y & \Y
& 2 & \N \\

\midrule
\multicolumn{12}{@{}l}{\textit{Specialized editing benchmarks}} \\

RefEdit-Bench~\cite{pathiraja2025refedit}
& 200
& \N & \Y & \Y & \Y & \Y & \N & \N & \Y
& 0 & \N \\

TextEditBench~\cite{gui2025texteditbench}
& 1,196
& \N & \Y & \N & \N & \Y & \U & \Y & \U
& 0 & \N \\

\midrule
\multicolumn{12}{@{}l}{\textit{Generation and editing benchmarks}} \\

ICE-Bench~\cite{pan2025ice}
& 6,538
& \Y & \Y & \Y & \Y & \N & \N & \N & \Y
& 1 & \N \\

\rowcolor{gray!12}
\textbf{MoI-Bench}
& \textbf{1,150}
& \Y & \Y & \Y & \Y & \Y & \Y & \Y & \Y
& 3 & \N \\

\bottomrule
\end{tabular}
\end{table*}

\paragraph{Task Distribution}
MoI-Bench contains 1150 samples across 64 subtasks defined by the \textit{Operation $\times$ Target} Task Matrix (Tab.~\ref{tab:moge_bench}).
Fig.~\ref{fig:survey_bmk_sta}(b--c) summarizes the sample distribution along the operation and target axes.
Local editing forms the largest operation group, with 560 samples covering addition, removal, replacement and adjustment.

Some subtasks span multiple target categories. 
For target statistics, we assign each sample to a single category according to the primary visual content to generate or modify.
Portraits form the largest target group, consistent with the emphasis on portrait-related needs identified in our user survey.

\paragraph{Input Configurations}
As shown in Fig.~\ref{fig:survey_bmk_sta}(d), MoI-Bench includes 150 text-to-image samples, 898 single-image samples and 102 multi-image samples.
Single-image samples comprise 860 direct-editing cases and 38 reference-only cases.
Multi-image samples use one base image with one, two or three reference images, comprising 82, 12 and 8 cases, respectively.
We limit reference inputs to three images to keep the benchmark focused on mobile use under resource constraints.

\paragraph{Comparison with existing benchmarks.}
We summarize task categories used in prior benchmarks to define the comparison dimensions in Tab.~\ref{tab:benchmark_comparison}.
These dimensions cover text-to-image generation, local and global editing, pose and spatial changes, viewpoint changes, implicit editing and multi-object editing.
We also compare input configurations, including the maximum number of reference images and support for multi-turn editing.
MoI-Bench covers text-to-image generation and editing tasks ranging from local modifications to implicit and multi-object edits, with up to three reference images.
Its distinguishing features are the \textit{Operation $\times$ Target} organization, task-specific quality criteria, and joint assessment of quality and mobile deployability.
We exclude multi-turn editing because our user survey indicates lower demand for this interaction pattern in mobile applications.

\section{MoI-Eval}
\label{sec:eval}

\subsection{Quality Rubric Cube}

We extend the \textit{Operation $\times$ Target} Task Matrix with a criteria axis to form a \textit{Quality Rubric Cube}.
Each applicable cell specifies the evaluation requirements for an \textit{operation $\times$ target} combination under one criterion.
%

\paragraph{Evaluation criteria.}
Following ImgEdit-Bench~\cite{ye2026imgedit}, we organize evaluation into three criterion groups: instruction following, visual quality and detail preservation.
These groups contain eight criteria that form the third axis of the rubric cube:

\begin{itemize}
    \item \textit{Instruction Following (IF)}: It assesses whether the output satisfies the requested operation and its constraints.
    \textit{Fidelity} checks whether the intended outcome is achieved, including specified attributes, relations and reference conditions.
    \textit{Selection} checks whether the operation targets the correct region or instance, such as the leftmost person in a group rather than another person.

    \item \textit{Visual Quality (VQ)}: It assesses the visual quality of the generated or edited image.
    \textit{Integrity} checks whether generated or modified content has complete shapes and is free from distortions, malformed parts and artifacts.
    \textit{Coherence} checks consistency within the output and between edited regions and their surroundings, including boundaries, textures and lighting.
    \textit{Presentation} assesses whole-image quality, including readability, composition and physical plausibility.
 
    \item \textit{Detail Preservation (DP)}: It assesses whether content and attributes outside the requested change remain preserved. 
    \textit{Context} checks non-target objects, animals and background details.
    \textit{Structure} checks framing and spatial relations, including poses and positions that should remain unchanged.
    Given the demand for identity preservation in our user survey, \textit{Identity} explicitly checks human identity and appearance features not targeted by the instruction.
    Preservation requirements exclude changes requested by the user; for example, hairstyle replacement should preserve facial identity but not the original hairstyle.

\end{itemize}

\paragraph{Task-specific rubrics.}
For each subtask, we define task-specific requirements and scoring rules for the applicable criteria.
Each rubric entry $(o,t,c)$ corresponds to operation $o$, target $t$ and criterion $c$.
For example, \texttt{Removal} $\times$ \texttt{Multi-Subjects} $\times$ \texttt{Identity} checks whether the remaining people's identities are preserved after removing the specified person.

We then manually mark each entry as always applicable, sample-dependent, or not applicable.
For text-to-image generation, all three \texttt{DP} criteria are inapplicable because no source image is provided.
For object editing, \texttt{Identity} is sample-dependent and applies only when the input contains people whose identities should be preserved.
These applicability labels define the sparse rubric cube.
For each sample, we include always-applicable entries and resolve sample-dependent entries using its instruction and input images.
Non-applicable entries are excluded from evaluation.

\paragraph{Sample-grounded questions.}
Following question-based evaluation methods (\eg, TIFA~\cite{hu2023tifa}), we specify evaluation requirements as questions about whether the output satisfies them.
For sample $s$, we construct a question set $\mathcal{Q}_{s,c}$ for each applicable criterion:
\begin{equation}
\mathcal{Q}_{s,c} = G(R_{o_s,t_s,c}, p_s, \mathcal{I}_s), \qquad c \in \mathcal{C}_s,
\end{equation}
where $o_s$ and $t_s$ denote the operation and target category of sample $s$, $R_{o_s,t_s,c}$ is the rubric for criterion $c$, $p_s$ is the instruction, and $\mathcal{I}_s$ contains the input images.
$\mathcal{I}_s$ is empty for text-to-image generation. $\mathcal{C}_s$ contains the criteria applicable to sample $s$.
$G$ denotes question construction using a VLM (\ie, Gemini-3.1-pro~\cite{gemini31pro2026}), followed by manual review.
For example, removing the leftmost person from a group portrait yields an \textit{Identity} question: ``Do the remaining people retain their original identities?''
Two annotators cross-review each sample to confirm rubric applicability and ensure that its questions match the instruction and input images without duplicating requirements. 
Each question includes task-specific conditions for full, partial and failed outcomes, constructed without access to model outputs.
The finalized questions and scoring conditions are shared across all evaluated models and judges.

\subsection{Quality Scoring and Capability Profiles}
\label{sec:quality}

We aggregate question-level judgments into overall quality $Q$ and capability profiles by operation, target and evaluation criterion.

\paragraph{Scoring and aggregation.}

We use GPT-4o~\cite{openai2024gpt4o} as the primary VLM judge to evaluate each question against the model output, using the instruction and input images as needed.
We additionally evaluate with Gemini-3.1-pro~\cite{gemini31pro2026} under the same protocol.
We adopt the three-level score mapping of Qwen-Image-Bench~\cite{li2026qwen}, normalized to $\{0, 0.6, 1\}$.
For binary requirements, only scores of 0 and 1 apply.
Under our rubrics, these scores indicate unmet, partially met and fully met requirements, respectively.
For sample $s$, let $z_{s,c}$ denote the score for criterion $c$ and $v_{s,k}$ the score for criterion group $k \in \{\mathrm{IF}, \mathrm{VQ}, \mathrm{DP}\}$.
\begin{equation}
    z_{s,c} = \frac{1}{|\mathcal{Q}_{s,c}|} \sum_{q \in \mathcal{Q}_{s,c}} r_{s,q}, 
    \qquad
    v_{s,k} = \frac{1}{|\mathcal{C}_{s,k}|} \sum_{c \in \mathcal{C}_{s,k}} z_{s,c},
\end{equation}
where $r_{s,q}$ is the question-level score and $\mathcal{C}_{s,k}$ contains the applicable criteria in group $k$ for sample $s$.
Non-applicable criteria are excluded, and applicable criteria receive equal weight within each group regardless of question counts.
Following prior uses of geometric aggregation~\cite{ku2024viescore,kamath2025geneval}, we compute the sample quality score $u_s$ as a weighted geometric mean of its criterion-group scores.
\begin{equation}
u_s = \prod_{k \in \mathcal{K}_s}
  v_{s,k}^{\widetilde{w}_{s,k}},
\qquad
\widetilde{w}_{s,k} = \frac{w_{s,k}} {\sum_{\ell \in \mathcal{K}_s} w_{s,\ell}},
\end{equation}
where $\mathcal{K}_s$ contains the applicable criterion groups and $w_{s,k}$ is a predefined weight for criterion group $k$, set by task operation and fixed across models.
Geometric aggregation limits compensation for poor performance in one criterion group by high scores in others.
Only IF and VQ contribute to text-to-image generation.
Weight settings are provided in Appendix~\ref{app:aggregation}, and sensitivity abalysis are reported in Appendix~\ref{app:weight-sensitivity}.

Overall quality is the mean score over the retained sample set $\mathcal{S}_m$ for model $m$:
\begin{equation}
Q_m = \frac{1}{|\mathcal{S}_m|}\sum_{s \in \mathcal{S}_m} u_s.
\end{equation}
Starting from all 1150 samples, we exclude confirmed generation or judge execution errors.
Unsupported tasks receive $u_s=0$ and remain in $\mathcal{S}_m$.
Successfully generated and evaluated outputs retain their scores regardless of quality.
Each retained sample contributes equally to $Q_m$, while subtasks contribute in proportion to their retained sample counts.

\paragraph{Capability profiles and Quality tiers.}

\begin{table*}[t]
\centering
\caption{
Quality and deployability tiers.
All listed requirements must be satisfied, with higher tiers taking precedence.
$C_{0.6}$ denotes the proportion of subtasks with mean quality scores of at least $0.6$.
The group requirement applies to each of IF, VQ and DP.
Q4 and D4 are the highest tiers.
D2--D4 require successful generation of valid outputs on the target device.
--- indicates no additional constraint.
}
\label{tab:quality_tiers}

\footnotesize
\setlength{\tabcolsep}{1pt}
\renewcommand{\arraystretch}{1}

\resizebox{0.75\textwidth}{!}{%
\begin{tabular}{@{}lccc@{\hspace{10pt}}lccc@{}}
\toprule
\multicolumn{4}{c}{Quality}
& \multicolumn{4}{c}{Deployability} \\
\cmidrule(lr){1-4}
\cmidrule(lr){5-8}
Tier & Overall $Q$ & Coverage $C_{0.6}$ & Each group
& Tier & Size (MiB) & E2E (s) & Memory (MiB) \\
\midrule

Q4 & $\geq 0.90$ & $\geq 90\%$ & $\geq 0.90$
& D4 & $\leq 1{,}024$ & $\leq 2$ & $\leq 2{,}048$ \\

Q3 & $\geq 0.70$ & $\geq 60\%$ & $\geq 0.60$
& D3 & $\leq 2{,}048$ & $\leq 30$ & $\leq 4{,}096$ \\

Q2 & $\geq 0.30$ & --- & ---
& D2 & --- & --- & --- \\

Q1 & $< 0.30$ & --- & ---
& D1 & \multicolumn{3}{c}{Unavailable Deployment} \\

\bottomrule
\end{tabular}%
}
\end{table*}

We aggregate the underlying evaluation scores along the operation, target and criteria axes to characterize model capabilities beyond overall quality $Q$.
Operation and target profiles average sample quality scores $u_s$ within each category, using the target assignments defined in Sec.~\ref{sec:bmk}.
Joint \textit{Operation $\times$ Target} profiles further reveal performance on specific tasks and connect model capabilities to concrete mobile application needs.
For example, a user who wants to remove text from a photograph can consult the Removal $\times$ Text \& Documents score, while replacing the text requires examining the corresponding Replacement score.
This distinction supports model selection for individual requests rather than relying on a broad text-editing score.

We also average $v_{s,k}$ and $z_{s,c}$ over their applicable samples to obtain profiles for the three criterion groups and eight individual criteria.
These profiles help researchers identify the capabilities each model needs to improve within instruction following, visual quality and detail preservation.
For example, successful person removal may still involve unintended changes to the remaining people's identities.
Examining Fidelity and Identity separately reveals this difference between completing the requested change and preserving existing details.
All profiles follow the same sample inclusion rules as overall quality.

To guide model development towards real mobile user needs, we further define quality tiers that jointly consider output quality, task coverage and performance across criterion groups.
A model in the highest tier should produce high-quality outputs across a broad range of tasks and meet minimum requirements in each criterion group.
Overall quality alone does not establish this breadth, especially when subtasks contain different numbers of samples.
We therefore complement $Q$ with task coverage, defined as the proportion of subtasks whose mean quality reaches a specified threshold:
\begin{equation}
C_{\tau}
=
\frac{1}{|\mathcal{T}|}
\sum_{j \in \mathcal{T}}
\mathbf{1}\!\left[\bar{u}_j \geq \tau\right],
\end{equation}
where $\bar{u}_j$ is the mean quality of subtask $j$, $\mathcal{T}$ contains the 64 subtasks, and $\tau=0.6$.
Unlike the sample-averaged score $Q$, coverage weights all subtasks equally.
It measures how broadly a model meets the quality threshold, rather than simply counting the tasks it supports.

We assign each model the highest qualifying tier from Q1 to Q4 using the joint requirements for $Q$, $C_{0.6}$ and benchmark-level IF, VQ and DP scores in Tab.~\ref{tab:quality_tiers}.
Q4 is the highest tier.
These requirements distinguish high average quality from broad task coverage and consistent performance across criterion groups, helping identify whether model development should prioritize output quality, task coverage or specific quality requirements.

\subsection{Quality-Deployability Map}

The quality score $Q$ measures output quality but does not account for the time and resources required for mobile inference.
We assess deployability using end-to-end latency, peak memory and deployed model size.
These metrics capture waiting time, runtime memory usage and storage requirements in mobile use.
We therefore present quality and deployability on separate axes in the \textit{Quality-Deployability Map}.

\paragraph{Deployability measurement.}
Mobile deployment involves model-specific quantization, operator adaptation and runtime optimization.
Differences in model architecture and device support make it difficult to apply an identical optimization procedure across different models.
We therefore evaluate the resource costs of deployed configurations from a mobile user's perspective: storage requirements, waiting time and runtime memory usage.
Our protocol accommodates model-specific deployment optimizations and measures the resulting deployment costs using consistent metrics.

A deployment configuration specifies model weights, component precisions, inference settings and runtime.
We report three metrics:
\begin{itemize}

    \item \textit{Model Size.} 
    Total storage required for standalone inference, including the backbone, image decoder, required encoders and supporting inference files.
    Shared files are counted once; duplicate artifacts and development files are excluded.

    \item \textit{Inference Latency.} 
    End-to-end time covering prompt processing, text encoding, generation, image decoding and post-processing.
    Initial model loading, warm-up and result-file writing are excluded.
    Loading and offloading performed during generation remain part of the measured latency.

    \item \textit{Peak Memory.} 
    Maximum sampled application memory within each inference run, averaged across runs.
    We use application physical footprint on iOS and process proportional set size (PSS) on Android, reporting the two platform-specific measurements separately.

\end{itemize}
Storage and memory are reported in MiB.
Failed and unmeasured configurations are distinguished.
Devices, workloads, precision settings and measurement procedures are detailed in the experimental setup.

\paragraph{Deployability tiers.}
Alongside quality tiers, we define deployability tiers D1--D4 using the joint requirements in Tab.~\ref{tab:quality_tiers}.
D1 denotes unavailable local deployment under the tested conditions, including deployment failure, invalid output, cloud- or API-only access, or a model that is too large for the target device.
D2 requires successful execution, while D3 and D4 impose progressively tighter latency, storage and memory limits.
D4 is the highest tier, designed to represent responsive mobile inference with modest resource requirements.
These tiers assess the deployment outcome achieved by a model and its implementation on a specified device, allowing improvements from hardware and runtime optimization to be reflected in its rating.
Thresholds are shared across devices, models and precision settings.

For the baseline comparison, we do not introduce bespoke operator adaptations or additional model-specific performance optimizations beyond those required to execute the models.
Among the evaluated models, DreamLite is the only model for which we successfully deployed editing on-device.
We therefore use text-to-image generation as the common deployment task supported by all evaluated on-device models.

\paragraph{Tier threshold calibration.}
We first manually define initial thresholds for quality and deployability tiers.
For quality tiers, we compare assignments obtained from GPT-4o and Gemini~3.1~Pro, identify disagreements, and iteratively adjust the shared thresholds to improve agreement while preserving meaningful distinctions between capability levels.
For deployability tiers, we compare assignments for matched deployment configurations on two devices with comparable hardware capabilities and calibrate the shared thresholds to reduce cross-device discrepancies.
The resulting thresholds are fixed for all reported comparisons and listed in Tab.~\ref{tab:quality_tiers}.
Sensitivity analyses are provided in the supplementary material.

\paragraph{Q-D Map construction.}

Evaluating an on-device model requires considering both quality and deployability.
High quality cannot compensate for unsuccessful deployment, while low latency cannot compensate for insufficient task capability.
We therefore retain these dimensions as separate axes rather than combining them into a single score.
The \textit{Quality-Deployability} capability map places reference quality tiers from Sec.~\ref{sec:quality} on the horizontal axis and device-specific deployability tiers on the vertical axis.
The upper-right region represents the goal of an on-device flagship model, combining high quality, broad task coverage and efficient mobile execution.
The lower-right region represents high-quality models with limited mobile deployability, including server-side models accessed through cloud APIs.
The map thus indicates whether model development should prioritize quality and task coverage, deployment efficiency, or both.

A complementary scatter plot presents continuous quality scores against end-to-end mobile latency.
Color indicates task coverage, and bubble area represents total parameter count.
Models without successful mobile inference are displayed separately without latency coordinates.
Together, these views show each model's position and the measured differences behind its tier assignment.

\section{Experiments}
\label{sec:exp}

\subsection{Experimental Setup}

\paragraph{Evaluation Models.}
We evaluate 19 representative models, including 16 open-weight models and three closed-source models.
Given the limited availability of lightweight models, we include widely used and recently released larger models as quality references.
The evaluated models are listed in Tab.~\ref{tab:quality_overall}.
We access GPT Image 2, Nano Banana 2 and Seedream 5.0 Pro through their APIs.

\paragraph{Quality Evaluation.}

We run open-weight models on NVIDIA A100 GPUs using reference implementations and released checkpoints.
Inference follows the default hyperparameters unless otherwise specified.
We generate one output per sample and use seed 0 where the implementation supports seed control.
We use GPT-4o~\cite{openai2024gpt4o} as the primary judge and independently evaluate the same outputs with Gemini~3.1~Pro~\cite{gemini31pro2026}.
Both judges use identical sample-specific questions and scoring conditions under the protocol in Sec.~\ref{sec:eval}.
Overall results from both judges are presented in Tab.~\ref{tab:quality_overall}.
Complete prompts are provided in the Appendix~\ref{app:questions}

\paragraph{Deployment Configurations.}
We evaluate models with fewer than 4B total parameters on an iPhone~17~Pro and a Xiaomi~15~Pro.
To simplify and standardize deployment, we use 4-bit quantization for external text encoders and retain the original checkpoint precision for the remaining components.
For a subset of models, a comparison with the original configurations shows only minor quality differences and unchanged quality tiers after text-encoder quantization (Appendix~\ref{sec:text-encoder-quantization}).
The iOS implementation uses Core ML for image-generation components and MLX for external text encoders.
The Android implementation uses ONNX Runtime with the Qualcomm QNN HTP backend for image-generation components and CPU execution for text encoders.
DreamLite is the only evaluated model for which we successfully deployed editing on-device, so we use text-to-image generation as the common deployment task.
Beyond changes required for execution, we introduce no additional model-specific operator or runtime optimizations.
Each successful configuration receives one warm-up run followed by three measured runs.
We report mean end-to-end latency and mean per-run peak memory, sampled every 25\,ms.
Models that cannot be deployed on the target device serve as quality references and receive D1.
Results for these configurations are presented in the Tab.~\ref{tab:deployment-results}.

\subsection{Overall Quality Analysis}

\paragraph{Quantitative Evaluation}
\begin{table*}[t]
\centering
\caption{
Overall quality and task coverage evaluated by GPT-4o and Gemini~3.1~Pro.
Tasks indicates support for generation (Gen.), editing (Edit), or both (Both); Both does not imply support for all subtasks.
Type denotes the image generation mechanism: diffusion-based (Diff., including flow matching), autoregressive (AR).
$C_{0.6}$ denotes the proportion of subtasks with mean quality scores of at least $0.6$.
Criterion-group scores are reported as (IF, VQ, DP).
Both judges yield the same quality tiers under the thresholds in Tab.~\ref{tab:quality_tiers}; Q4 is the highest tier.
Within each model group, models are ordered by ascending quality tier and then GPT-4o overall quality.
}
\label{tab:quality_overall}

\small
\setlength{\tabcolsep}{4pt}
\renewcommand{\arraystretch}{1.10}

\resizebox{\textwidth}{!}{%
\begin{tabular}{@{}lccccccccc@{}}
\toprule
\multirow{2}{*}{Model}
& \multirow{2}{*}{Tasks}
& \multirow{2}{*}{Type}
& \multicolumn{3}{c}{GPT-4o}
& \multicolumn{3}{c}{Gemini 3.1 Pro}
& \multirow{2}{*}{Q-tier} \\
\cmidrule(lr){4-6}
\cmidrule(lr){7-9}
& & &
$Q$ & $C_{0.6}$ & (IF, VQ, DP) & $Q$ & $C_{0.6}$ & (IF, VQ, DP)
& \\
\midrule

\multicolumn{10}{@{}l}{\textit{Closed-source Models}} \\
\addlinespace[2pt]

Nano Banana 2~\cite{google2026nanobanana2}
& Both & --
& 0.95 & 1.00 & (0.95, 0.97, 0.97)
& 0.93 & 1.00 & (0.94, 0.96, 0.96)
& Q4 \\

Seedream 5.0 Pro~\cite{bytedance2026seedream5}
& Both & --
& 0.95 & 1.00 & (0.95, 0.97, 0.98)
& 0.93 & 1.00 & (0.93, 0.97, 0.96)
& Q4 \\

GPT Image 2~\cite{openai2026gptimage25}
& Both & --
& 0.97 & 1.00 & (0.97, 0.98, 0.98)
& 0.96 & 1.00 & (0.96, 0.98, 0.96)
& Q4 \\

\midrule
\multicolumn{10}{@{}l}{\textit{Open-weight Reference Models}} \\
\addlinespace[2pt]

LongCat-Image~\cite{team2025longcat}
& Gen. & Diff.
& 0.12 & 0.13 & (0.12, 0.12, 0.00)
& 0.12 & 0.13 & (0.12, 0.12, 0.00)
& Q1 \\

Lumina-DiMOO-8B~\cite{xin2025lumina}
& Both & Diff.
& 0.43 & 0.23 & (0.48, 0.47, 0.62)
& 0.31 & 0.14 & (0.42, 0.36, 0.45)
& Q2 \\

InternVL-U-4B~\cite{tian2026internvl}
& Both & Diff.
& 0.51 & 0.36 & (0.61, 0.55, 0.48)
& 0.39 & 0.22 & (0.57, 0.48, 0.33)
& Q2 \\

VIBE-Image-Edit~\cite{alekseenko2026vibe}
& Edit & Diff.
& 0.59 & 0.56 & (0.64, 0.60, 0.78)
& 0.53 & 0.47 & (0.58, 0.55, 0.73)
& Q2 \\

OmniGen2~\cite{wu2026omnigen2}
& Both & Diff.
& 0.62 & 0.55 & (0.66, 0.70, 0.73)
& 0.54 & 0.36 & (0.62, 0.66, 0.65)
& Q2 \\

BAGEL-7B-MoT~\cite{deng2025bagel}
& Both & Diff.
& 0.64 & 0.58 & (0.69, 0.68, 0.77)
& 0.60 & 0.47 & (0.67, 0.64, 0.77)
& Q2 \\

LongCat-Image-Edit
& Edit & Diff.
& 0.67 & 0.67 & (0.66, 0.70, 0.86)
& 0.63 & 0.61 & (0.64, 0.70, 0.83)
& Q2 \\

Lance-3B~\cite{fu2026lance}
& Both & Diff.
& 0.67 & 0.67 & (0.69, 0.74, 0.76)
& 0.60 & 0.52 & (0.64, 0.69, 0.65)
& Q2 \\

DeepGen 1.0~\cite{wang2026deepgen}
& Both & Diff.
& 0.69 & 0.70 & (0.74, 0.74, 0.70)
& 0.57 & 0.44 & (0.69, 0.66, 0.52)
& Q2 \\

FLUX.2 [klein] 4B~\cite{blackforestlabs2026flux2klein}
& Both & Diff.
& 0.83 & 0.91 & (0.82, 0.88, 0.95)
& 0.79 & 0.78 & (0.79, 0.85, 0.93)
& Q3 \\

\midrule
\multicolumn{10}{@{}l}{\textit{On-device Models} (Total Params. $<4$B)} \\
\addlinespace[2pt]

Janus-Pro-1B~\cite{chen2025janus}
& Gen. & AR
& 0.08 & 0.08 & (0.09, 0.08, 0.00)
& 0.08 & 0.08 & (0.09, 0.08, 0.00)
& Q1 \\

Mobile-O-0.5B~\cite{shaker2026mobile}
& Gen. & Diff.
& 0.10 & 0.11 & (0.10, 0.10, 0.00)
& 0.09 & 0.09 & (0.10, 0.10, 0.00)
& Q1 \\

SANA-Sprint-0.6B~\cite{chen2025sana}
& Gen. & Diff.
& 0.10 & 0.11 & (0.10, 0.10, 0.00)
& 0.10 & 0.11 & (0.10, 0.10, 0.00)
& Q1 \\

SANA-600M~\cite{xie2024sana}
& Gen. & Diff.
& 0.10 & 0.11 & (0.11, 0.11, 0.00)
& 0.10 & 0.11 & (0.11, 0.10, 0.00)
& Q1 \\

DreamLite~\cite{feng2026dreamlite}
& Both & Diff.
& 0.62 & 0.64 & (0.65, 0.64, 0.76)
& 0.52 & 0.42 & (0.62, 0.52, 0.62)
& Q2 \\

Ovis-U1-3B~\cite{wang2025ovis}
& Both & Diff.
& 0.66 & 0.59 & (0.68, 0.70, 0.72)
& 0.54 & 0.38 & (0.62, 0.64, 0.51)
& Q2 \\

\bottomrule
\end{tabular}%
}
\end{table*}

Tab.~\ref{tab:quality_overall} reports overall quality and task coverage under both judges.
The three closed-source models reach Q4 and meet the task-quality threshold on all 64 subtasks, while FLUX.2 [klein] leads the open-weight models at Q3.
Among models below 4B total parameters, DreamLite and Ovis-U1 reach Q2.
Ovis-U1 achieves higher overall quality, but DreamLite has higher task coverage under both judges: 0.64 versus 0.59 with GPT-4o, and 0.42 versus 0.38 with Gemini.
Generation-only models (\ie, LongCat-Image, SANA, \etc) remain at Q1 because unsupported editing tasks contribute zero; their overall scores therefore reflect limited coverage of this editing-focused benchmark, not generation quality alone.

\textit{Higher average quality does not necessarily mean that a model meets the quality threshold on more tasks, making task coverage a complementary consideration for model selection.}

\paragraph{Qualitative Evaluation}

\begin{figure}[t]
    \centering
    \includegraphics[width=\linewidth]{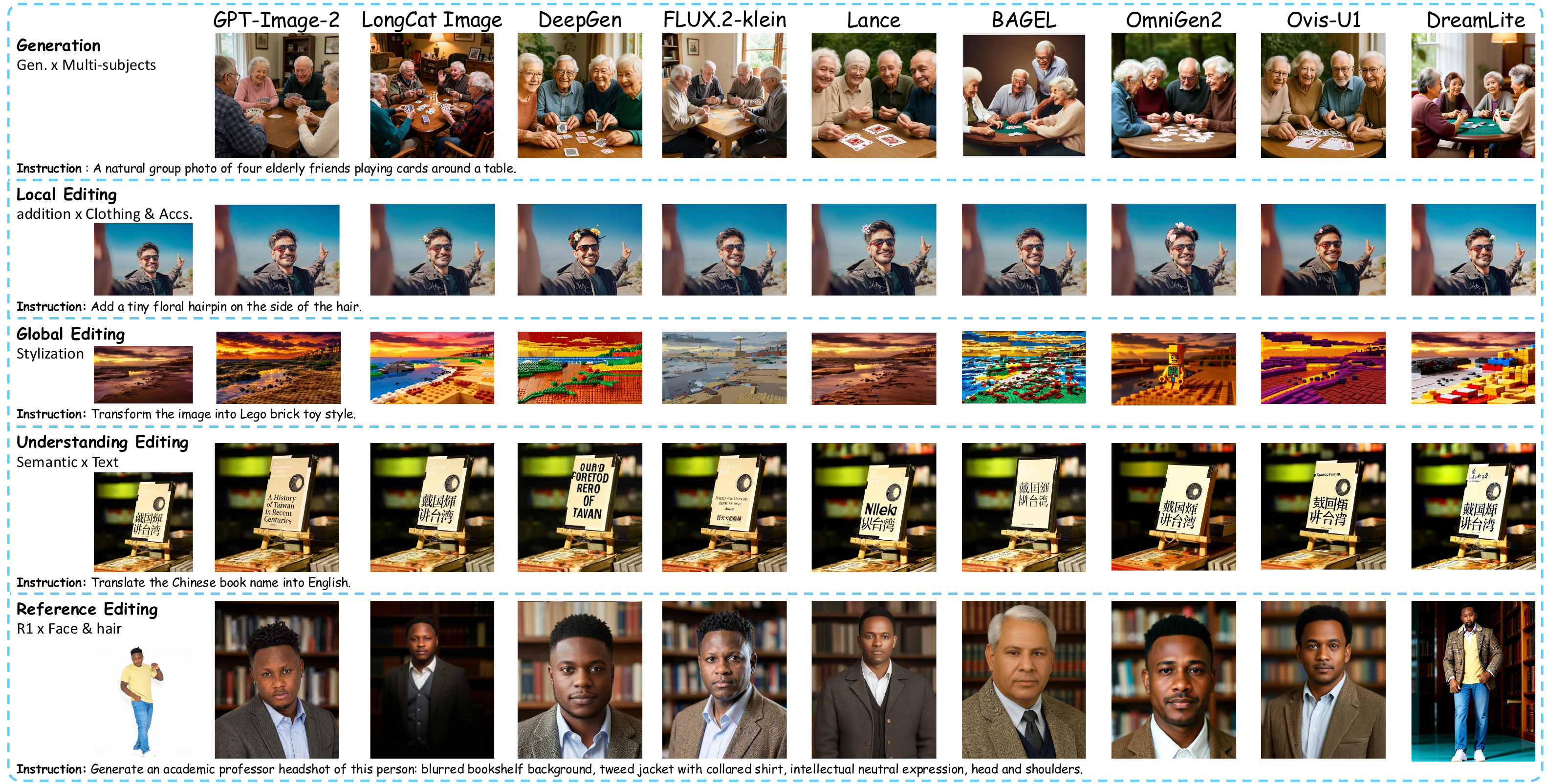}
    \caption{
    Qualitative comparison of representative image generation and editing models on MoI-Bench across the five operation groups. Zoom in for better quality.
    }
    \label{fig:qualitative_com}
\end{figure}

Fig.~\ref{fig:qualitative_com} compares representative outputs across the five operation groups.
DreamLite and Ovis-U1 produce plausible group portraits and visible Lego-style transformations, but neither fully translates the Chinese book title into English, whereas GPT Image 2 produces a readable English title.
In reference editing, DreamLite retains a full-body composition despite the headshot instruction, while Ovis-U1 produces the requested framing but changes the person's appearance.
Additional qualitative results are provided in the Appendix~\ref{app:qualitative}.

\textit{These examples suggest that compact models need more precise instruction following and better preservation of reference details alongside their generation and stylization capabilities.}

\vspace{-5pt}
\subsection{Capability Analysis}

\begin{figure}[t]
    \centering
    \includegraphics[width=\linewidth]{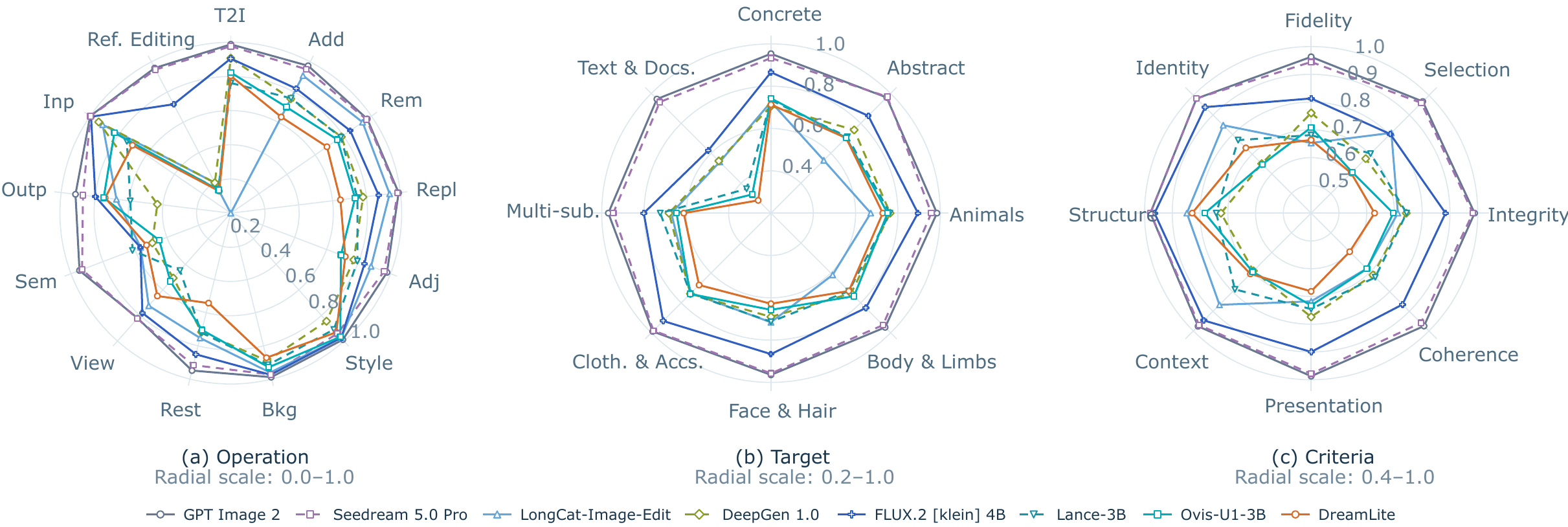}
    \caption{Capability profiles of representative models across (a) operations, (b) targets and (c) evaluation criteria, scored by GPT-4o. Higher scores indicate better performance. Radial scales differ across subplots.}
    \label{fig:capability_radars}
    \vspace{-10pt}
\end{figure}

Fig.~\ref{fig:capability_radars} summarizes the operation, target and criterion profiles of representative models.
The profiles show that model strengths vary across these dimensions.
We examine these differences below through detailed breakdowns and joint Operation $\times$ Target analysis.

\vspace{-10pt}
\paragraph{Operation-level performance.}

\begin{table*}[t]
\centering
\caption{
Operation-level quality evaluated by GPT-4o.
Scores average sample quality over the retained samples for each operation. 
Unsupported tasks contribute zero.
Add., Rem., Repl. and Adj. denote addition, removal, replacement and adjustment.
Style, Bkg., Rest. and View denote stylization, background modification, restoration and viewpoint changes.
Sem., Outp. and Inp. denote semantic editing, outpainting and inpainting.
R1 uses one reference image and B1+R1--B1+R3 use one base image and one to three references.
}
\label{tab:capability-operation}
\small
\setlength{\tabcolsep}{1.5pt}
\renewcommand{\arraystretch}{1.10}
\resizebox{\textwidth}{!}{%
\begin{tabular}{@{}l*{16}{c}@{}}
\toprule
\multirow{2}{*}{Model}
& Generation & \multicolumn{4}{c}{Local Editing}
& \multicolumn{4}{c}{Global Editing}
& \multicolumn{3}{c}{Understand. Editing}
& \multicolumn{4}{c}{Ref. Editing} \\
\cmidrule(lr){2-2}\cmidrule(lr){3-6}\cmidrule(lr){7-10}\cmidrule(lr){11-13}\cmidrule(lr){14-17}
& T2I & Add. & Rem. & Repl. & Adj.
& Style & Bkg. & Rest. & View
& Sem. & Outp. & Inp.
& R1 & B1+R1 & B1+R2 & B1+R3 \\
\midrule
\multicolumn{17}{@{}l}{\textit{Closed-source Models}} \\
\addlinespace[2pt]
Nano Banana 2 & 0.99 & 0.94 & 0.95 & 0.98 & 0.96 & 0.99 & 0.97 & 0.92 & 0.80 & 0.95 & 0.95 & 0.99 & 0.96 & 0.93 & 0.88 & 0.80 \\
Seedream 5.0 Pro & 0.98 & 0.95 & 0.96 & 0.98 & 0.96 & 0.94 & 0.97 & 0.92 & 0.83 & 0.93 & 0.87 & 1.00 & 0.97 & 0.94 & 0.90 & 0.99 \\
GPT Image 2 & 0.99 & 0.97 & 0.97 & 0.99 & 0.98 & 0.99 & 0.99 & 0.95 & 0.82 & 0.95 & 0.92 & 1.00 & 0.97 & 0.96 & 0.87 & 0.99 \\
\midrule
\multicolumn{17}{@{}l}{\textit{Open-weight Reference Models}} \\
\addlinespace[1pt]
Lumina-DiMOO-8B & 0.86 & 0.53 & 0.45 & 0.45 & 0.41 & 0.74 & 0.38 & 0.26 & 0.23 & 0.30 & 0.35 & 0.38 & 0.18 & 0.04 & 0.00 & 0.00 \\
InternVL-U-4B & 0.87 & 0.56 & 0.39 & 0.53 & 0.54 & 0.87 & 0.75 & 0.44 & 0.20 & 0.45 & 0.69 & 0.57 & 0.60 & 0.00 & 0.00 & 0.00 \\
VIBE-Image-Edit & 0.00 & 0.79 & 0.87 & 0.82 & 0.79 & 0.96 & 0.90 & 0.75 & 0.33 & 0.58 & 0.57 & 0.87 & 0.55 & 0.00 & 0.00 & 0.00 \\
OmniGen2 & 0.83 & 0.64 & 0.64 & 0.67 & 0.66 & 0.88 & 0.90 & 0.41 & 0.30 & 0.40 & 0.42 & 0.67 & 0.54 & 0.46 & 0.44 & 0.24 \\
LongCat-Image-Edit & 0.00 & 0.91 & 0.94 & 0.94 & 0.88 & 0.97 & 0.96 & 0.75 & 0.72 & 0.56 & 0.67 & 0.91 & 0.74 & 0.00 & 0.00 & 0.00 \\
Lance-3B & 0.76 & 0.76 & 0.78 & 0.75 & 0.79 & 0.91 & 0.91 & 0.72 & 0.45 & 0.62 & 0.59 & 0.74 & 0.58 & 0.00 & 0.00 & 0.00 \\
DeepGen 1.0 & 0.91 & 0.76 & 0.79 & 0.78 & 0.77 & 0.84 & 0.89 & 0.72 & 0.51 & 0.49 & 0.43 & 0.94 & 0.74 & 0.00 & 0.00 & 0.00 \\
FLUX.2 [klein] 4B & 0.90 & 0.82 & 0.85 & 0.87 & 0.84 & 0.97 & 0.97 & 0.85 & 0.78 & 0.57 & 0.80 & 0.99 & 0.78 & 0.71 & 0.61 & 0.70 \\
\midrule
\multicolumn{17}{@{}l}{\textit{On-device Models}} \\
\addlinespace[2pt]
Mobile-O-0.5B & 0.74 & 0.00 & 0.00 & 0.00 & 0.00 & 0.00 & 0.00 & 0.00 & 0.00 & 0.00 & 0.00 & 0.00 & 0.00 & 0.00 & 0.00 & 0.00 \\
SANA-600M & 0.80 & 0.00 & 0.00 & 0.00 & 0.00 & 0.00 & 0.00 & 0.00 & 0.00 & 0.00 & 0.00 & 0.00 & 0.00 & 0.00 & 0.00 & 0.00 \\
DreamLite & 0.80 & 0.64 & 0.68 & 0.65 & 0.72 & 0.93 & 0.87 & 0.54 & 0.65 & 0.53 & 0.74 & 0.70 & 0.54 & 0.00 & 0.00 & 0.00 \\
Ovis-U1-3B & 0.82 & 0.70 & 0.76 & 0.73 & 0.69 & 0.97 & 0.93 & 0.70 & 0.53 & 0.45 & 0.75 & 0.83 & 0.56 & 0.00 & 0.00 & 0.00 \\
\bottomrule
\end{tabular}%
}
\end{table*}

Tab.~\ref{tab:capability-operation} shows that compact models are competitive on some operations but remain limited on others.
DreamLite and Ovis-U1 achieve stylization scores of 0.93 and 0.97, respectively, close to FLUX.2 [klein]'s 0.97, while larger gaps remain in replacement and restoration.
Their relative strengths also vary: Ovis-U1 performs better on removal and inpainting, whereas DreamLite scores higher on adjustment and viewpoint changes.
Both models support single-reference tasks but lack support for the evaluated base-plus-reference configurations, indicating a coverage limitation beyond output quality.

\textit{Current compact models still lack support for many functions. Broader task support can be an important direction for future development.}

\paragraph{Target-level performance.}

\begin{table*}[t]
\centering
\caption{
Target-level quality evaluated by GPT-4o across the eight target categories.
Scores average sample quality using the sample-level target assignments.
Unsupported tasks contribute zero, and confirmed execution errors are excluded.
Scores are rounded to two decimal places.
}
\label{tab:capability-target}
\small
\setlength{\tabcolsep}{1.5pt}
\renewcommand{\arraystretch}{1.10}
\resizebox{\textwidth}{!}{%
\begin{tabular}{@{}l*{8}{c}@{}}
\toprule
\multirow{2}{*}{Model}
& \multicolumn{2}{c}{General Objects}
& \multirow{2}{*}{Animals}
& \multicolumn{4}{c}{Portraits}
& \multirow{2}{*}{\shortstack{Text\&Docs.}} \\
\cmidrule(lr){2-3}\cmidrule(lr){5-8}
& Concrete & Abstract &
& \shortstack{Body\&Limbs}
& \shortstack{Face\&Hair}
& \shortstack{Cloth.\&Accs.}
& \shortstack{Multi-subjects} & \\
\midrule
\multicolumn{9}{@{}l}{\textit{Closed-source Models}} \\
\addlinespace[2pt]
Nano Banana 2 & 0.95 & 0.95 & 0.99 & 0.94 & 0.95 & 0.97 & 0.94 & 0.93 \\
Seedream 5.0 Pro & 0.93 & 0.98 & 0.96 & 0.95 & 0.96 & 0.99 & 0.95 & 0.94 \\
GPT Image 2 & 0.95 & 0.97 & 0.98 & 0.96 & 0.96 & 0.99 & 0.97 & 0.96 \\
\midrule
\multicolumn{9}{@{}l}{\textit{Open-weight Reference Models}} \\
\addlinespace[1pt]
Lumina-DiMOO-8B & 0.46 & 0.67 & 0.53 & 0.37 & 0.41 & 0.48 & 0.42 & 0.27 \\
InternVL-U-4B & 0.52 & 0.55 & 0.56 & 0.62 & 0.55 & 0.67 & 0.29 & 0.38 \\
VIBE-Image-Edit & 0.67 & 0.51 & 0.61 & 0.56 & 0.67 & 0.68 & 0.63 & 0.34 \\
OmniGen2 & 0.64 & 0.68 & 0.76 & 0.68 & 0.60 & 0.73 & 0.69 & 0.34 \\
LongCat-Image-Edit & 0.73 & 0.55 & 0.67 & 0.61 & 0.72 & 0.73 & 0.67 & 0.54 \\
Lance-3B & 0.73 & 0.71 & 0.76 & 0.72 & 0.71 & 0.74 & 0.72 & 0.36 \\
DeepGen 1.0 & 0.71 & 0.76 & 0.76 & 0.74 & 0.69 & 0.74 & 0.68 & 0.55 \\
FLUX.2 [klein] 4B & 0.87 & 0.85 & 0.89 & 0.83 & 0.87 & 0.92 & 0.80 & 0.62 \\
\midrule
\multicolumn{9}{@{}l}{\textit{On-device Models}} \\
\addlinespace[2pt]
Mobile-O-0.5B & 0.07 & 0.21 & 0.15 & 0.16 & 0.09 & 0.07 & 0.12 & 0.05 \\
SANA-600M & 0.06 & 0.24 & 0.18 & 0.15 & 0.09 & 0.08 & 0.11 & 0.07 \\
DreamLite & 0.71 & 0.70 & 0.73 & 0.72 & 0.63 & 0.68 & 0.61 & 0.29 \\
Ovis-U1-3B & 0.74 & 0.70 & 0.75 & 0.76 & 0.66 & 0.74 & 0.65 & 0.33 \\
\bottomrule
\end{tabular}%
}
\end{table*}

Tab.~\ref{tab:capability-target} shows that text and documents remain challenging for both compact and larger open-weight models.
DreamLite and Ovis-U1 score 0.29 and 0.33 in this category, compared with 0.61--0.76 across portrait categories.
The gap also exists among stronger reference models. 
For example, FLUX.2 [klein] scores 0.62 on text and documents, compared with 0.94 for Seedream 5.0 Pro and 0.96 for GPT Image 2.

\textit{Current on-device models underperform on text handling, facial tasks, and multi-person tasks.}

\paragraph{Operation $\times$ Target analysis.}

\begin{figure}
    \centering
    \includegraphics[width=\linewidth]{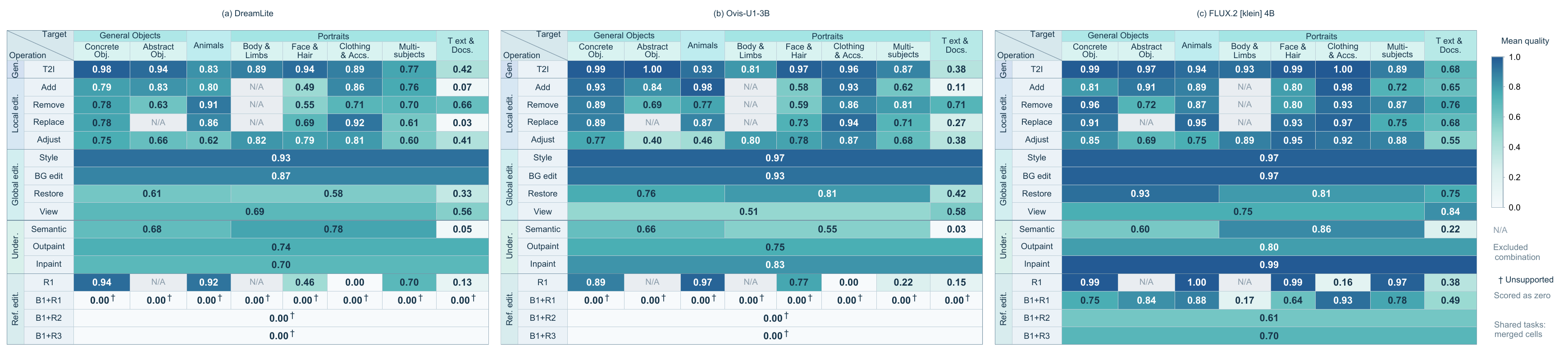}
    \caption{
    Operation $\times$ Target quality profiles evaluated by GPT-4o.
    Cells report mean sample quality for each task.
    N/A denotes excluded combinations, and $\dagger$ marks unsupported tasks scored as zero.
    Merged cells indicate shared tasks.
    }
    \label{fig:oxt_heatmap}
\end{figure}

Fig.~\ref{fig:oxt_heatmap} reveals task-specific differences that operation-level and target-level averages obscure.
Although text and documents receive low aggregate scores, DreamLite and Ovis-U1 achieve 0.66 and 0.71 on text removal, close to FLUX.2 [klein]'s 0.76.
Their scores fall to 0.03 and 0.27 on text replacement, and 0.07 and 0.11 on text addition.

\textit{These results suggest that limited text-generation capability is the main weakness. }
And \textit{improving replacement therefore requires identifying which content types remain difficult, rather than treating the operation as uniformly weak.}

In reference editing, model performance also varies across targets.
For R1 tasks involving facial features and hairstyles, Ovis-U1 outperforms DreamLite (0.77 versus 0.46), but the ordering reverses for multiple subjects (0.22 versus 0.70).
A single reference-editing or portrait score would conceal this distinction.
Moreover, both models perform well on R1 tasks involving concrete objects and animals, yet neither supports the evaluated base-plus-reference configurations.
Their reference-editing limitations thus involve both quality on particular targets and missing input configurations.

\textit{The joint analysis helps developers distinguish task-specific quality gaps from missing functionality and supports model selection according to both the requested operation and its target.}

\paragraph{Criterion-level analysis.}

\begin{table*}[t]
\centering
\caption{
Capability profiles for the eight evaluation criteria, evaluated by GPT-4o.
IF, VQ and DP denote instruction following, visual quality and detail preservation.
Scores follow the existing criterion-profile aggregation and include zeros for unsupported tasks.
Scores are rounded to two decimal places.
}
\label{tab:capability-criteria}
\small
\setlength{\tabcolsep}{1.5pt}
\renewcommand{\arraystretch}{1.10}
\resizebox{0.8\textwidth}{!}{%
\begin{tabular}{@{}l*{8}{c}@{}}
\toprule
\multirow{2}{*}{Model}
& \multicolumn{2}{c}{IF}
& \multicolumn{3}{c}{VQ}
& \multicolumn{3}{c}{DP} \\
\cmidrule(lr){2-3}\cmidrule(lr){4-6}\cmidrule(lr){7-9}
& Fidelity & Selection & Integrity & Coherence & Presentation
& Context & Structure & Identity \\
\midrule
\multicolumn{9}{@{}l}{\textit{Closed-source Models}} \\
\addlinespace[2pt]
Nano Banana 2 & 0.94 & 0.95 & 0.98 & 0.96 & 0.98 & 0.98 & 0.96 & 0.99 \\
Seedream 5.0 Pro & 0.94 & 0.96 & 0.98 & 0.96 & 0.98 & 0.97 & 0.98 & 0.98 \\
GPT Image 2 & 0.96 & 0.97 & 0.99 & 0.97 & 0.99 & 0.98 & 0.98 & 0.98 \\
\midrule
\multicolumn{9}{@{}l}{\textit{Open-weight Reference Models}} \\
\addlinespace[2pt]
Lumina-DiMOO-8B & 0.48 & 0.42 & 0.46 & 0.43 & 0.53 & 0.62 & 0.65 & 0.53 \\
InternVL-U-4B & 0.67 & 0.46 & 0.55 & 0.53 & 0.57 & 0.42 & 0.52 & 0.54 \\
VIBE-Image-Edit & 0.60 & 0.72 & 0.64 & 0.61 & 0.66 & 0.82 & 0.82 & 0.79 \\
OmniGen2 & 0.68 & 0.58 & 0.72 & 0.64 & 0.73 & 0.71 & 0.75 & 0.74 \\
LongCat-Image-Edit & 0.65 & 0.81 & 0.71 & 0.68 & 0.72 & 0.87 & 0.85 & 0.85 \\
Lance-3B & 0.68 & 0.70 & 0.74 & 0.73 & 0.75 & 0.79 & 0.74 & 0.77 \\
DeepGen 1.0 & 0.76 & 0.68 & 0.74 & 0.71 & 0.77 & 0.69 & 0.72 & 0.65 \\
FLUX.2 [klein] 4B & 0.81 & 0.80 & 0.88 & 0.87 & 0.90 & 0.95 & 0.96 & 0.94 \\
\midrule
\multicolumn{9}{@{}l}{\textit{On-device Models}} \\
\addlinespace[2pt]
Mobile-O-0.5B & 0.10 & 0.00 & 0.10 & 0.11 & 0.10 & 0.00 & 0.00 & 0.00 \\
SANA-600M & 0.11 & 0.00 & 0.10 & 0.11 & 0.10 & 0.00 & 0.00 & 0.00 \\
DreamLite & 0.66 & 0.60 & 0.63 & 0.60 & 0.68 & 0.71 & 0.83 & 0.73 \\
Ovis-U1-3B & 0.71 & 0.61 & 0.70 & 0.68 & 0.73 & 0.70 & 0.78 & 0.65 \\
\bottomrule
\end{tabular}%
}
\end{table*}

Tab.~\ref{tab:capability-criteria} shows that the gap between on-device and larger models varies across quality requirements.
Ovis-U1 approaches LongCat-Image-Edit in Integrity and Coherence but trails in Selection and Identity, while DreamLite approaches it in Structure but trails in Integrity and Coherence.
These comparisons suggest that preserving image structure or producing visually coherent edits does not ensure accurate target selection or identity preservation.
Identity preservation is also a weakness of some larger models: DeepGen scores 0.65, below DreamLite's 0.73, despite its higher Fidelity and visual-quality scores.

\textit{On-device model development should go beyond visual quality to improve where edits are applied and whose identity is preserved, while addressing the visual defects that remain in models such as DreamLite.}

\subsection{Mobile Deployment}
\label{sec:mobile-deployment}

\paragraph{Baseline deployment.}

\begin{table*}[t]
\centering
\caption{
Mobile deployment results on the iPhone~17~Pro and Xiaomi~15~Pro.
Params. includes all parameterized components required for inference, including encoders and image codecs.
Res. and Steps denote output resolution and denoising steps.
Size, E2E and Mem. denote model storage, end-to-end inference latency and peak application memory.
Memory is measured using physical footprint on iOS and process PSS on Android.
$^{\dagger}$The Android Janus-Pro deployment uses precomputed prompt embeddings; its reported latency excludes prompt preparation.
}
\label{tab:deployment-results}

\small
\setlength{\tabcolsep}{3pt}
\renewcommand{\arraystretch}{1.12}

\resizebox{\textwidth}{!}{%
\begin{tabular}{@{}lccc rrr rrr c@{}}
\toprule
\multirow{2}{*}{Model}
& \multirow{2}{*}{Params.}
& \multirow{2}{*}{Res.}
& \multirow{2}{*}{Steps}
& \multicolumn{3}{c}{iPhone 17 Pro}
& \multicolumn{3}{c}{Xiaomi 15 Pro}
& \multirow{2}{*}{\shortstack{D-tier}} \\
\cmidrule(lr){5-7}
\cmidrule(lr){8-10}
& & & &
Size (MiB) & E2E (s) & Mem. (MiB)
& Size (MiB) & E2E (s) & Mem. (MiB)
& \\
\midrule

Janus-Pro-1B
$^{\dagger}$
& 2.1B & $384^2$ & --
& 2975.4 & 58.26 & 1505.4
& 2974.1 & 76.01 & 3253.8
& D2 \\

Mobile-O-0.5B
& 1.7B & $512^2$ & 20
& 1718.3 & 2.81 & 426.8
& 1770.2 & 13.30 & 1757.1
& D3 \\

SANA-Sprint-0.6B
& 3.5B & $1024^2$ & 2
& 2860.9 & 4.49 & 1727.2
& 3069.1 & 25.24 & 3896.7
& D2 \\

SANA-600M
& 3.5B & $1024^2$ & 18
& 2857.2 & 15.33 & 1775.2
& 3065.4 & 139.68 & 3897.6
& D2 \\

DreamLite
& 2.5B & $1024^2$ & 4
& 1684.6 & 3.22 & 1305.7
& 1804.5 & 18.19 & 2390.5
& D3 \\

Ovis-U1-3B
& 3.6B & $1024^2$ & 50
& 4544.8 & 106.12 & 6106.9
& 3499.4 & 2526.25 & 2716.6
& D2 $^\dagger$ \\

\bottomrule
\end{tabular}%
}
\begin{minipage}{\textwidth}
\footnotesize
$^{\dagger}$ indicates visible artifacts in the output images.
\end{minipage}
\end{table*}

Tab.~\ref{tab:deployment-results} compares mobile deployment costs across the two devices.
Mobile-O and DreamLite reach D3, while no evaluated configuration reaches D4.
DreamLite generates at $1024^2$ resolution in 3.22\,s on iOS and 18.19\,s on Android, whereas Mobile-O achieves lower latency at $512^2$.
SANA-Sprint completes inference in under 30\,s on both devices but remains D2 because its storage exceeds the 2\,GiB limit for D3.
Device-dependent latency differences are substantial: SANA-600M takes 15.33\,s on iOS and 139.68\,s on Android.

\textit{Even setting output quality aside, current mobile deployment still falls short of user expectations, highlighting the need to reduce memory usage while maintaining low latency.}
\paragraph{Precision configurations.}

\begin{table*}[t]
\centering
\caption{
Deployment results under different backbone and decoder precisions on the iPhone~17~Pro and Xiaomi~15~Pro.
Size and Mem.\ are measured in MiB, and E2E latency is measured in seconds.
Blank cells indicate results pending verification or measurement; -- denotes confirmed execution failure.
Status describes the inspected outputs: C denotes recognizable outputs and F denotes severe quality degradation or failed execution. 
}
\label{tab:precision-comparison}

\small
\setlength{\tabcolsep}{5pt}
\renewcommand{\arraystretch}{1.12}

\resizebox{\textwidth}{!}{%
\begin{tabular}{@{}lccrrrcrrrc@{}}
\toprule
\multirow{2}{*}{Model}
& \multicolumn{2}{c}{Precision}
& \multicolumn{4}{c}{iPhone 17 Pro}
& \multicolumn{4}{c}{Xiaomi 15 Pro} \\
\cmidrule(lr){2-3}
\cmidrule(lr){4-7}
\cmidrule(lr){8-11}
& Backbone & Decoder
& Size & E2E & Mem. & Status
& Size & E2E & Mem. & Status \\
\midrule

\multirow{3}{*}{Janus-Pro-1B}
& FP16  & W8A16
& 2935.0 & 58.96 & 1502.2 & C
& 2934.0 & 77.59 & 3253.8 & C \\
& W8A16 & FP16
& 1764.4 & 54.54 & 1500.8 & C
& -- & -- & -- & F \\
& W8A16 & W8A16
& 1724.0 & 54.28 & 1503.6 & C
& -- & -- & -- & F \\
\midrule

\multirow{3}{*}{Mobile-O-0.5B}
& FP16  & W8A16
& 1580.1 & 2.84 & 449.2 & C
& 1619.8 & 12.49 & 1767.0 & C \\
& W8A16 & FP16
& 1155.7 & 2.68 & 423.5 & C
& 1209.3& 8.21 & 3149.2 & C \\
& W8A16 & W8A16
& 1017.5 & 2.65 & 424.2 & C
& 1059.5 & 7.11 & 3127.6 & C \\
\midrule

\multirow{3}{*}{SANA-Sprint-0.6B}
& FP16  & W8A16
& 2722.6 & 3.84 & 1851.1 & C
& 2919.5 & 23.36 & {3893.1} & C$^{\dagger}$ \\
& W8A16 & FP16
& 2296.7 & 3.23 & 1744.3 & C
& -- & -- & -- & F \\
& W8A16 & W8A16
& 2158.5 & 3.62 & 1862.5 & C
& -- & -- & -- & F \\
\midrule

\multirow{3}{*}{SANA-600M}
& FP16  & W8A16
& 2706.1 & 11.33 & 4289.3 & C
& 2915.8 & 142.11 & {3901.7} & C$^{\dagger\ddagger}$ \\
& W8A16 & FP16
& 2294.6 & 11.93 & 3760.8 & C
& 2504.1 & 44.55 & {4255.4} & C$^{\dagger}$ \\
& W8A16 & W8A16
& 2143.5 & 11.83 & 4130.6 & C
& 2353.8 & 41.50 & {4307.6} & C$^{\dagger}$ \\
\midrule

\multirow{3}{*}{DreamLite}
& FP16  & W8A16
& 1683.4 & 3.62 & 1306.5 & C
& 1803.3 & 18.14 & {2392.5} & C \\
& W8A16 & FP16
& 1314.2 & 3.44 & 1297.9 & C
& 1439.6 & 17.16 & {2406.2} & C \\
& W8A16 & W8A16
& 1313.1 & 3.42 & 1298.3 & C
& 1438.5 & 16.98 & {2390.3} & C \\
\midrule

\multirow{3}{*}{Ovis-U1-3B}
& FP16  & W8A16
& 4497.68 & 120.52 & 6116.51 & C 
& 3452.5 & 3099.0 & 6413.0 & F \\
& W8A16 & FP16
& -- & -- & -- & F
& -- & -- & -- & F \\
& W8A16 & W8A16
& -- & -- & -- & F
& -- & -- & -- & F \\

\bottomrule
\end{tabular}%
}

\begin{minipage}{\textwidth}
\footnotesize
$^{\dagger}$ Execution sequentially unloads the text encoder and backbone, and uses tiled decoding in an isolated process.
$^{\ddagger}$ The FP16 backbone uses 29 sessions; W8A16 backbones use a single session.
\end{minipage}
\end{table*}

We vary backbone and decoder precision across FP16, W8A16 and W8A8.
W8A16 and W8A8 use 8-bit weights with 16-bit and 8-bit activations respectively.
We focus on FP16 and W8A16 combinations in Tab.~\ref{tab:precision-comparison}, as W8A8 configurations frequently encounter execution failures or severe visual degradation.
Weight quantization generally reduces model storage, but its latency and memory benefits are less consistent.
For example, Mobile-O runs faster with a W8A16 backbone on Android, while its peak memory increases substantially.
Several configurations that produce recognizable outputs on iOS also fail execution or output-quality checks on Android, indicating that precision choices must be validated for each model and platform.

\paragraph{Latency breakdown.}

\begin{table*}[t]
\centering
\caption{
Latency breakdown on the iPhone~17~Pro with FP16 backbones and decoders.
Times are in seconds; -- denotes unavailable results.
}
\label{tab:latency-breakdown}

\small
\setlength{\tabcolsep}{7pt}
\renewcommand{\arraystretch}{1}

\resizebox{\textwidth}{!}{%
\begin{tabular}{@{}lrccrrrr@{}}
\toprule
\multirow{2}{*}{Model}
& \multirow{2}{*}{Cond.}
& \multicolumn{3}{c}{Generation}
& \multirow{2}{*}{Decode}
& \multirow{2}{*}{Other}
& \multirow{2}{*}{Total} \\
\cmidrule(lr){3-5}
& & Steps & Branches & Time/step & & & \\
\midrule

Janus-Pro-1B
& 2.214 & 576 & 2 & 0.0971
& 0.083 & 0.042 & 58.256 \\

Mobile-O-0.5B
& 0.095 & 20 & 2 & 0.1134
& 0.230 & 0.216 & 2.809 \\

SANA-Sprint-0.6B
& 0.281 & 2 & 1 & 0.3920
& 3.154 & 0.270 & 4.488 \\

SANA-600M
& 0.298 & 18 & 2 & 0.6378
& 3.035 & 0.515 & 15.328 \\

DreamLite
& 0.065 & 4 & 1 & 0.7645
& 0.036 & 0.065 & 3.224 \\

Ovis-U1-3B
& 0.818 & 50 & 1 & 2.008
& 4.347 & 0.536 & 106.121 \\

\bottomrule
\end{tabular}%
}
\end{table*}

Tab.~\ref{tab:latency-breakdown} shows that fewer generation steps do not necessarily lead to lower total latency.
Mobile-O completes 20 steps faster than DreamLite completes four, owing to its lower per-step cost and lower output resolutions.
Janus-Pro has a low per-token cost, but generating 576 image tokens accounts for approximately 96\% of its total latency.
For SANA-Sprint, two-step generation takes only 0.78\,s, while decoding takes 3.15\,s and accounts for 70\% of total latency.

\textit{Step distillation and CFG distillation should be prioritized to improve inference efficiency, alongside further optimization of the decoder.}

\subsection{Q-D analysis.}
\begin{figure}
    \centering
    \includegraphics[width=\linewidth]{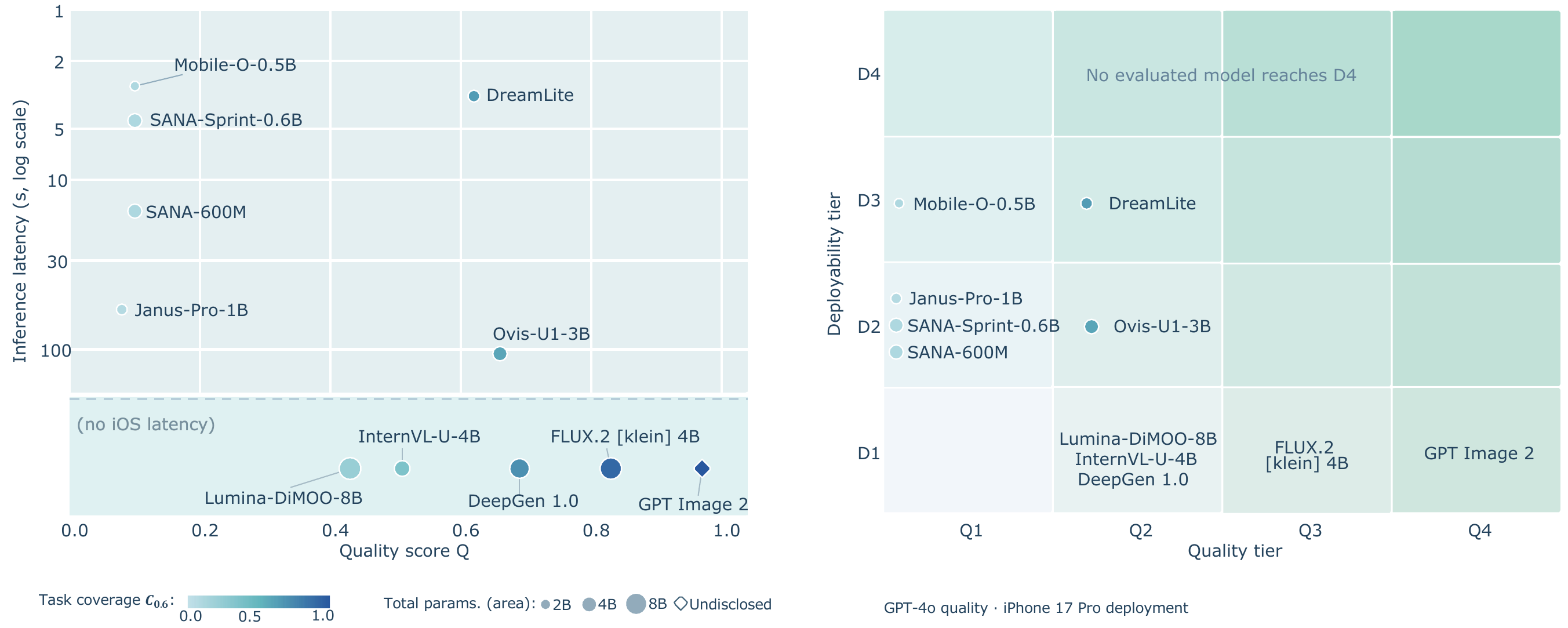}
    \caption{
    Quality and deployability of representative models.
    Left: GPT-4o quality versus iOS inference latency, with color indicating task coverage and bubble area indicating total parameters; diamonds denote undisclosed parameter counts.
    Models without iOS latency measurements are shown separately.
    Right: quality and deployability tiers.
    }
    \label{fig:qd_tier}
\end{figure}

Fig.~\ref{fig:qd_tier} shows that fast mobile inference does not necessarily provide broad task capability.
Mobile-O and DreamLite have similar iOS latency (2.81 and 3.22\,s), yet their task coverage differs substantially (0.11 and 0.64), placing them in Q1--D3 and Q2--D3, respectively.
Mobile-O's low benchmark-wide quality primarily reflects its generation-only scope, highlighting the distinction between efficiently executing a supported task and meeting diverse mobile editing needs.
Ovis-U1 achieves a slightly higher overall quality score than DreamLite (0.66 versus 0.62), but lower task coverage (0.59 versus 0.64), alongside substantially higher latency and peak memory.
\textit{Thus, a higher average quality score alone does not establish a more useful mobile model.}

These results suggest different development priorities for different starting points.
For efficient generation-only models, a key next step is to support editing while retaining their deployment efficiency.
For models such as DreamLite, priorities include broader input support, particularly multiple reference images, and better editing quality in the weak operation--target combinations identified by our capability analysis.
For Ovis-U1, reducing inference latency and memory demand is necessary to make its existing capabilities more accessible on-device.
Step and CFG distillation, decoder optimization and memory-efficient execution are candidate approaches, whose benefits should be evaluated alongside any changes in output quality.
\textit{Moving toward the upper-right region of the map therefore requires both broader, more reliable capabilities and lower deployment costs; the evaluated on-device models currently reach neither Q3 nor D4.}

When model-size constraints limit the capabilities achievable locally, an on-device agent could use the capability profiles to route requests between local models and cloud services.
Routing should depend on the requested operation and target rather than broad application labels.
For example, Ovis-U1 performs comparably to FLUX.2 [klein] on text removal but falls far behind on text replacement, suggesting that these requests may warrant different execution choices.
\textit{A router could retain locally supported requests that meet the required quality and latency budgets, while sending tasks such as difficult text replacement or unsupported multi-image editing to a more capable cloud model.
Evaluating this task-aware routing strategy is a direction for future work.}

\section{Conclusion}

We introduce MoI-Bench and MoI-Eval to connect users' needs for image generation and editing on mobile devices with model evaluation and development.
MoI-Bench organizes 64 subtasks along operation and target axes, while MoI-Eval uses task-specific criteria to assess output quality and construct capability profiles.
For users, these profiles help select on-device or cloud models for specific requests, such as makeup removal or text replacement.
For researchers, they identify unsupported tasks or unmet quality requirements, providing directions for expanding model capabilities and improving quality.
The \textit{Quality $\times$ Deployability} map complements this analysis with mobile latency, storage and memory costs, helping developers assess which capabilities are practical to offer locally.
Together, this framework supports evaluating image models by how well they meet users' needs and guides the development of more capable and accessible image AI across mobile and cloud platforms.

\clearpage
\bibliography{iclr2027_conference}

@inproceedings{chen2023speed,
  title={Speed is all you need: On-device acceleration of large diffusion models via gpu-aware optimizations},
  author={Chen, Yu-Hui and Sarokin, Raman and Lee, Juhyun and Tang, Jiuqiang and Chang, Chuo-Ling and Kulik, Andrei and Grundmann, Matthias},
  booktitle={2023 IEEE/CVF Conference on Computer Vision and Pattern Recognition Workshops (CVPRW)},
  pages={4651--4655},
  year={2023},
  organization={IEEE}
}

@article{choi2023squeezing,
  title={Squeezing large-scale diffusion models for mobile},
  author={Choi, Jiwoong and Kim, Minkyu and Ahn, Daehyun and Kim, Taesu and Kim, Yulhwa and Jo, Dongwon and Jeon, Hyesung and Kim, Jae-Joon and Kim, Hyungjun},
  journal={arXiv preprint arXiv:2307.01193},
  year={2023}
}

@article{li2023snapfusion,
  title={Snapfusion: Text-to-image diffusion model on mobile devices within two seconds},
  author={Li, Yanyu and Wang, Huan and Jin, Qing and Hu, Ju and Chemerys, Pavlo and Fu, Yun and Wang, Yanzhi and Tulyakov, Sergey and Ren, Jian},
  journal={Advances in Neural Information Processing Systems},
  volume={36},
  pages={20662--20678},
  year={2023}
}

@inproceedings{zhao2024mobilediffusion,
  title={Mobilediffusion: Instant text-to-image generation on mobile devices},
  author={Zhao, Yang and Xu, Yanwu and Xiao, Zhisheng and Jia, Haolin and Hou, Tingbo},
  booktitle={European Conference on Computer Vision},
  pages={225--242},
  year={2024},
  organization={Springer}
}

@article{castells2024edgefusion,
  title={Edgefusion: On-device text-to-image generation},
  author={Castells, Thibault and Song, Hyoung-Kyu and Piao, Tairen and Choi, Shinkook and Kim, Bo-Kyeong and Yim, Hanyoung and Lee, Changgwun and Kim, Jae Gon and Kim, Tae-Ho},
  journal={arXiv preprint arXiv:2404.11925},
  year={2024}
}

@inproceedings{chen2025snapgen,
  title={Snapgen: Taming high-resolution text-to-image models for mobile devices with efficient architectures and training},
  author={Chen, Jierun and Hu, Dongting and Huang, Xijie and Coskun, Huseyin and Sahni, Arpit and Gupta, Aarush and Goyal, Anujraaj and Lahiri, Dishani and Singh, Rajesh and Idelbayev, Yerlan and others},
  booktitle={2025 IEEE/CVF Conference on Computer Vision and Pattern Recognition (CVPR)},
  pages={7997--8008},
  year={2025},
  organization={IEEE}
}

@inproceedings{zhao2024mixdq,
  title={Mixdq: Memory-efficient few-step text-to-image diffusion models with metric-decoupled mixed precision quantization},
  author={Zhao, Tianchen and Ning, Xuefei and Fang, Tongcheng and Liu, Enshu and Huang, Guyue and Lin, Zinan and Yan, Shengen and Dai, Guohao and Wang, Yu},
  booktitle={European Conference on Computer Vision},
  pages={285--302},
  year={2024},
  organization={Springer}
}

@article{li2024svdquant,
  title={Svdquant: Absorbing outliers by low-rank components for 4-bit diffusion models},
  author={Li, Muyang and Lin, Yujun and Zhang, Zhekai and Cai, Tianle and Li, Xiuyu and Guo, Junxian and Xie, Enze and Meng, Chenlin and Zhu, Jun-Yan and Han, Song},
  journal={arXiv preprint arXiv:2411.05007},
  year={2024}
}

@inproceedings{chavhan2026nanoflux,
  title={NanoFLUX: Distillation-Driven Compression of Large Text-to-Image Generation Models for Mobile Devices},
  author={Chavhan, Ruchika and Chadwick, Malcolm and Ramos, Alberto Gil Couto Pimentel and Morreale, Luca and Noroozi, Mehdi and Mehrotra, Abhinav},
  booktitle={Forty-third International Conference on Machine Learning},
  year={2026}
}

@article{hu2026snapgen++,
  title={SnapGen++: Unleashing Diffusion Transformers for Efficient High-Fidelity Image Generation on Edge Devices},
  author={Hu, Dongting and Gupta, Aarush and Gabidolla, Magzhan and Sahni, Arpit and Coskun, Huseyin and Li, Yanyu and Idelbayev, Yerlan and Mahmood, Ahsan and Lebedev, Aleksei and Lahiri, Dishani and others},
  journal={arXiv preprint arXiv:2601.08303},
  year={2026}
}

@article{kodavanti2026edgedit,
  title={EdgeDiT: Hardware-Aware Diffusion Transformers for Efficient On-Device Image Generation},
  author={Kodavanti, Sravanth and Arveti, Manjunath and Vajrala, Sowmya and Miriyala, Srinivas and others},
  journal={arXiv preprint arXiv:2603.28405},
  year={2026}
}

@article{du2026elasticdit,
  title={ElasticDiT: Efficient Diffusion Transformers via Elastic Architecture and Sparse Attention for High-Resolution Image Generation on Mobile Devices},
  author={Du, Kunpeng and Xie, Haizhen and Lu, Sen and Yu, Lei and Bao, Binglei and Tang, Huaao and Liu, Chuntao and Wu, Hao and Zhao, Yang and Huang, Zhicai and others},
  journal={arXiv preprint arXiv:2605.15684},
  year={2026}
}

@article{shaker2026mobile,
  title={Mobile-O: Unified Multimodal Understanding and Generation on Mobile Device},
  author={Shaker, Abdelrahman and Heakl, Ahmed and Muhammad, Jaseel and Thawkar, Ritesh and Thawakar, Omkar and Li, Senmao and Cholakkal, Hisham and Reid, Ian and Xing, Eric P and Khan, Salman and others},
  journal={arXiv preprint arXiv:2602.20161},
  year={2026}
}

@article{feng2026dreamlite,
  title={DreamLite: A Lightweight On-Device Unified Model for Image Generation and Editing},
  author={Feng, Kailai and Wei, Yuxiang and Chen, Bo and Pan, Yang and Ye, Hu and Liu, Songwei and Yan, Chenqian and Gao, Yuan},
  journal={arXiv preprint arXiv:2603.28713},
  year={2026}
}

@article{deng2026blazeedit,
  title={BlazeEdit: Generalist Image Editing on Mobile Devices with Image-to-Image Diffusion Models},
  author={Deng, Fei and Xu, Yanwu and Bao, Zhipeng and Zhang, Zhixing and Jia, Haolin and Raveendran, Karthik and Wei, Jianing},
  journal={arXiv preprint arXiv:2605.28067},
  year={2026}
}

@article{saharia2022photorealistic,
  title={Photorealistic text-to-image diffusion models with deep language understanding},
  author={Saharia, Chitwan and Chan, William and Saxena, Saurabh and Li, Lala and Whang, Jay and Denton, Emily L and Ghasemipour, Kamyar and Gontijo Lopes, Raphael and Karagol Ayan, Burcu and Salimans, Tim and others},
  journal={Advances in neural information processing systems},
  volume={35},
  pages={36479--36494},
  year={2022}
}

@inproceedings{hu2023tifa,
  title={Tifa: Accurate and interpretable text-to-image faithfulness evaluation with question answering},
  author={Hu, Yushi and Liu, Benlin and Kasai, Jungo and Wang, Yizhong and Ostendorf, Mari and Krishna, Ranjay and Smith, Noah A},
  booktitle={2023 ieee/cvf international conference on computer vision (iccv)},
  pages={20349--20360},
  year={2023},
  organization={IEEE}
}

@inproceedings{bakr2023hrs,
  title={Hrs-bench: Holistic, reliable and scalable benchmark for text-to-image models},
  author={Bakr, Eslam Mohamed and Sun, Pengzhan and Shen, Xiaoqian and Khan, Faizan Farooq and Li, Li Erran and Elhoseiny, Mohamed},
  booktitle={2023 ieee/cvf international conference on computer vision (iccv)},
  pages={19984--19996},
  year={2023},
  organization={IEEE}
}

@article{huang2025t2i,
  title={T2i-compbench++: An enhanced and comprehensive benchmark for compositional text-to-image generation},
  author={Huang, Kaiyi and Duan, Chengqi and Sun, Kaiyue and Xie, Enze and Li, Zhenguo and Liu, Xihui},
  journal={IEEE Transactions on Pattern Analysis and Machine Intelligence},
  volume={47},
  number={5},
  pages={3563--3579},
  year={2025},
  publisher={IEEE}
}

@article{ghosh2023geneval,
  title={Geneval: An object-focused framework for evaluating text-to-image alignment},
  author={Ghosh, Dhruba and Hajishirzi, Hannaneh and Schmidt, Ludwig},
  journal={Advances in Neural Information Processing Systems},
  volume={36},
  pages={52132--52152},
  year={2023}
}

@article{lee2023holistic,
  title={Holistic evaluation of text-to-image models},
  author={Lee, Tony and Yasunaga, Michihiro and Meng, Chenlin and Mai, Yifan and Park, Joon Sung and Gupta, Agrim and Zhang, Yunzhi and Narayanan, Deepak and Teufel, Hannah and Bellagente, Marco and others},
  journal={Advances in Neural Information Processing Systems},
  volume={36},
  pages={69981--70011},
  year={2023}
}

@article{li2024genai,
  title={Genai-bench: Evaluating and improving compositional text-to-visual generation},
  author={Li, Baiqi and Lin, Zhiqiu and Pathak, Deepak and Li, Jiayao and Fei, Yixin and Wu, Kewen and Ling, Tiffany and Xia, Xide and Zhang, Pengchuan and Neubig, Graham and others},
  journal={arXiv preprint arXiv:2406.13743},
  year={2024}
}

@article{wu2024conceptmix,
  title={Conceptmix: A compositional image generation benchmark with controllable difficulty},
  author={Wu, Xindi and Yu, Dingli and Huang, Yangsibo and Russakovsky, Olga and Arora, Sanjeev},
  journal={Advances in Neural Information Processing Systems},
  volume={37},
  pages={86004--86047},
  year={2024}
}

@article{zhang2023magicbrush,
  title={Magicbrush: A manually annotated dataset for instruction-guided image editing},
  author={Zhang, Kai and Mo, Lingbo and Chen, Wenhu and Sun, Huan and Su, Yu},
  journal={Advances in neural information processing systems},
  volume={36},
  pages={31428--31449},
  year={2023}
}

@article{ju2023direct,
  title={Direct inversion: Boosting diffusion-based editing with 3 lines of code},
  author={Ju, Xuan and Zeng, Ailing and Bian, Yuxuan and Liu, Shaoteng and Xu, Qiang},
  journal={arXiv preprint arXiv:2310.01506},
  year={2023}
}

@article{ma2024i2ebench,
  title={I2ebench: A comprehensive benchmark for instruction-based image editing},
  author={Ma, Yiwei and Ji, Jiayi and Ye, Ke and Lin, Weihuang and Wang, Zhibin and Zheng, Yonghan and Zhou, Qiang and Sun, Xiaoshuai and Ji, Rongrong},
  journal={Advances in Neural Information Processing Systems},
  volume={37},
  pages={41494--41516},
  year={2024}
}

@article{qian2025gie,
  title={Gie-bench: Towards grounded evaluation for text-guided image editing},
  author={Qian, Yusu and Lu, Jiasen and Fu, Tsu-Jui and Wang, Xinze and Chen, Chen and Yang, Yinfei and Hu, Wenze and Gan, Zhe},
  journal={arXiv preprint arXiv:2505.11493},
  year={2025}
}

@article{liu2025step1x,
  title={Step1x-edit: A practical framework for general image editing},
  author={Liu, Shiyu and Han, Yucheng and Xing, Peng and Yin, Fukun and Wang, Rui and Cheng, Wei and Liao, Jiaqi and Wang, Yingming and Fu, Honghao and Han, Chunrui and others},
  journal={arXiv preprint arXiv:2504.17761},
  year={2025}
}

@inproceedings{jia2026compbench,
  title={Compbench: Benchmarking complex instruction-guided image editing},
  author={Jia, Bohan and Huang, Wenxuan and Tang, Yuntian and Qiao, Junbo and Liao, Jincheng and Cao, Shaosheng and Zhao, Fei and Feng, Zhaopeng and Gu, Zhouhong and Yin, Zhenfei and others},
  booktitle={Proceedings of the IEEE/CVF Conference on Computer Vision and Pattern Recognition},
  pages={1112--1122},
  year={2026}
}

@article{wu2026kris,
  title={Kris-bench: Benchmarking next-level intelligent image editing models},
  author={Wu, Yongliang and Li, Zonghui and Hu, Xinting and Ye, Xinyu and Zeng, Xianfang and Yu, Gang and Zhu, Wenbo and Schiele, Bernt and Yang, Ming-Hsuan and Yang, Xu},
  journal={Advances in Neural Information Processing Systems},
  volume={38},
  year={2026}
}

@article{ye2026imgedit,
  title={Imgedit: A unified image editing dataset and benchmark},
  author={Ye, Yang and He, Xianyi and Li, Zongjian and Yuan, Shenghai and Yan, Zhiyuan and Hou, Bohan and Yuan, Li and others},
  journal={Advances in Neural Information Processing Systems},
  volume={38},
  year={2026}
}

@article{yi2026vde,
  title={VDE Bench: Evaluating The Capability of Image Editing Models to Modify Visual Documents},
  author={Yi, Hongzhu and Yang, Yujia and Wang, Yuanxiang and Li, Tong and Guan, Zhenyu and Zong, Tianyu and Chen, Jiahuan and Bao, Chenxi and Yang, Tiankun and Jin, Haopeng and others},
  journal={arXiv preprint arXiv:2602.00122},
  year={2026}
}

@article{jiang2026geditbench,
  title={GEditBench v2: A Human-Aligned Benchmark for General Image Editing},
  author={Jiang, Zhangqi and Sun, Zheng and Zeng, Xianfang and Yang, Yufeng and Zhang, Xuanyang and Wu, Yongliang and Cheng, Wei and Yu, Gang and Yang, Xu and Wen, Bihan},
  journal={arXiv preprint arXiv:2603.28547},
  year={2026}
}

@inproceedings{ku2024imagenhub,
  title={Imagenhub: Standardizing the evaluation of conditional image generation models},
  author={Ku, Max and Li, Tianle and Zhang, Kai and Lu, Yujie and Fu, Xingyu and Zhuang, Wenwen and Chen, Wenhu},
  booktitle={International Conference on Learning Representations},
  volume={2024},
  pages={46689--46722},
  year={2024}
}

@inproceedings{pan2025ice,
  title={Ice-bench: A unified and comprehensive benchmark for image creating and editing},
  author={Pan, Yulin and He, Xiangteng and Mao, Chaojie and Han, Zhen and Jiang, Zeyinzi and Zhang, Jingfeng and Liu, Yu},
  booktitle={2025 IEEE/CVF International Conference on Computer Vision (ICCV)},
  pages={16586--16596},
  year={2025},
  organization={IEEE}
}

@article{hua2025mmigbench,
  title={Mmigbench: Towards comprehensive and explainable evaluation of multi-modal image generation models},
  author={Hua, Hang and Zeng, Ziyun and Song, Yizhi and Tang, Yunlong and He, Liu and Aliaga, Daniel and Xiong, Wei and Luo, Jiebo},
  journal={arXiv preprint arXiv:2505.19415},
  volume={3},
  pages={3},
  year={2025}
}

@article{sani2026imagenworld,
  title={ImagenWorld: Stress-Testing Image Generation Models with Explainable Human Evaluation on Open-ended Real-World Tasks},
  author={Sani, Samin Mahdizadeh and Ku, Max and Jamali, Nima and Sani, Matina Mahdizadeh and Khoshtab, Paria and Sun, Wei-Chieh and Fazel, Parnian and Tam, Zhi Rui and Chong, Thomas and Chan, Edisy Kin Wai and others},
  journal={arXiv preprint arXiv:2603.27862},
  year={2026}
}

@article{wang2025unigenbench++,
  title={Unigenbench++: A unified semantic evaluation benchmark for text-to-image generation},
  author={Wang, Yibin and Li, Zhimin and Zang, Yuhang and Bu, Jiazi and Zhou, Yujie and Xin, Yi and He, Junjun and Wang, Chunyu and Lu, Qinglin and Jin, Cheng and others},
  journal={arXiv preprint arXiv:2510.18701},
  year={2025}
}

@article{duan2026moebius,
  title={Moebius: 0.2 B Lightweight Image Inpainting Framework with 10B-Level Performance},
  author={Duan, Kangsheng and Xu, Ziyang and Liu, Wenyu and Ruan, Xiaohu and Chen, Xiaoxin and Wang, Xinggang},
  journal={arXiv preprint arXiv:2606.19195},
  year={2026}
}

@inproceedings{yu2025anyedit,
  title={Anyedit: Mastering unified high-quality image editing for any idea},
  author={Yu, Qifan and Chow, Wei and Yue, Zhongqi and Pan, Kaihang and Wu, Yang and Wan, Xiaoyang and Li, Juncheng and Tang, Siliang and Zhang, Hanwang and Zhuang, Yueting},
  booktitle={2025 IEEE/CVF Conference on Computer Vision and Pattern Recognition (CVPR)},
  pages={26125--26135},
  year={2025},
  organization={IEEE}
}

@article{labs2025flux,
  title={Flux. 1 kontext: Flow matching for in-context image generation and editing in latent space},
  author={Labs, Black Forest and Batifol, Stephen and Blattmann, Andreas and Boesel, Frederic and Consul, Saksham and Diagne, Cyril and Dockhorn, Tim and English, Jack and English, Zion and Esser, Patrick and others},
  journal={arXiv preprint arXiv:2506.15742},
  year={2025}
}

@article{bai2026edit,
  title={Edit-Compass \& EditReward-Compass: A Unified Benchmark for Image Editing and Reward Modeling},
  author={Bai, Xuehai and Shi, Yang and Zhang, Yi-Fan and Zhu, Xuanyu and Wang, Yuran and Dai, Yifan and Liu, Xinyu and Ji, Yiyan and Gu, Xiaoling and Zhang, Yuanxing},
  journal={arXiv preprint arXiv:2605.13062},
  year={2026}
}

@inproceedings{wu2026omnigen2,
  title={Omnigen2: Towards instruction-aligned multimodal generation},
  author={Wu, Chenyuan and Wang, Jiahao and Zheng, Pengfei and Yan, Ruiran and Xiao, Shitao and Luo, Xin and Wang, Yueze and Li, Wanli and Jiang, Xiyan and Liu, Yexin and others},
  booktitle={Proceedings of the IEEE/CVF Conference on Computer Vision and Pattern Recognition},
  pages={21964--21975},
  year={2026}
}

@inproceedings{oshima2026multibanana,
  title={Multibanana: A challenging benchmark for multi-reference text-to-image generation},
  author={Oshima, Yuta and Miyake, Daiki and Matsutani, Kohsei and Iwasawa, Yusuke and Suzuki, Masahiro and Matsuo, Yutaka and Furuta, Hiroki},
  booktitle={Proceedings of the IEEE/CVF Conference on Computer Vision and Pattern Recognition},
  pages={448--460},
  year={2026}
}

@article{wu2026micon,
  title={MICON-Bench: Benchmarking and Enhancing Multi-Image Context Image Generation in Unified Multimodal Models},
  author={Wu, Mingrui and Liu, Hang and Ji, Jiayi and Sun, Xiaoshuai and Ji, Rongrong},
  journal={arXiv preprint arXiv:2602.19497},
  year={2026}
}

@inproceedings{pathiraja2025refedit,
  title={Refedit: A benchmark and method for improving instruction-based image editing model on referring expressions},
  author={Pathiraja, Bimsara and Patel, Maitreya and Singh, Shivam and Yang, Yezhou and Baral, Chitta},
  booktitle={2025 IEEE/CVF International Conference on Computer Vision (ICCV)},
  pages={15646--15656},
  year={2025},
  organization={IEEE}
}

@article{gui2025texteditbench,
  title={Texteditbench: Evaluating reasoning-aware text editing beyond rendering},
  author={Gui, Rui and Wan, Yang and Han, Haochen and Mao, Dongxing and Liu, Fangming and Li, Min and Wang, Alex Jinpeng},
  journal={arXiv preprint arXiv:2512.16270},
  year={2025}
}

@misc{gemini31pro2026,
  author       = {{Google}},
  title        = {Gemini 3.1 Pro},
  year         = {2026},
  howpublished = {Google AI Blog},
  url          = {https://blog.google/innovation-and-ai/models-and-research/gemini-models/gemini-3-1-pro/},
  note         = {Version: gemini-3.1-pro-preview}
}

@article{openai2024gpt4o,
  title={Gpt-4o system card},
  author={Hurst, Aaron and Lerer, Adam and Goucher, Adam P and Perelman, Adam and Ramesh, Aditya and Clark, Aidan and Ostrow, AJ and Welihinda, Akila and Hayes, Alan and Radford, Alec and others},
  journal={arXiv preprint arXiv:2410.21276},
  year={2024}
}

@misc{openai2026gptimage25,
  author       = {{OpenAI}},
  title        = {Introducing {ChatGPT Images 2.5}},
  year         = {2026},
  howpublished = {\url{https://openai.com/index/introducing-chatgpt-images-2-5/}},
  note         = {Accessed: 2026-09-26}
}

@misc{bytedance2026seedream5,
  author       = {{ByteDance Seed}},
  title        = {{Seedream 5.0 Pro}},
  year         = {2026},
  howpublished = {\url{https://seed.bytedance.com/en/seedream5_0_pro}},
  note         = {Official model page. Accessed: 2026-09-26}
}

@article{li2026qwen,
  title={Qwen-image-bench: From generation to creation in text-to-image evaluation},
  author={Li, Niantong and Hu, Guangzheng and Qiao, Weixu and Ba, Ying and Hong, Qichen and Shen, Shijun and Wang, Jinlin and Zhou, Fan and Kang, Jianye and Shang, Xin and others},
  journal={arXiv preprint arXiv:2605.28091},
  year={2026}
}

@inproceedings{ku2024viescore,
  title={Viescore: Towards explainable metrics for conditional image synthesis evaluation},
  author={Ku, Max and Jiang, Dongfu and Wei, Cong and Yue, Xiang and Chen, Wenhu},
  booktitle={Proceedings of the 62nd Annual Meeting of the Association for Computational Linguistics (Volume 1: Long Papers)},
  pages={12268--12290},
  year={2024}
}

@article{kamath2025geneval,
  title={Geneval 2: Addressing benchmark drift in text-to-image evaluation},
  author={Kamath, Amita and Chang, Kai-Wei and Krishna, Ranjay and Zettlemoyer, Luke and Hu, Yushi and Ghazvininejad, Marjan},
  journal={arXiv preprint arXiv:2512.16853},
  year={2025}
}

@inproceedings{chen2025sana,
  title={Sana-sprint: One-step diffusion with continuous-time consistency distillation},
  author={Chen, Junsong and Xue, Shuchen and Zhao, Yuyang and Yu, Jincheng and Paul, Sayak and Chen, Junyu and Cai, Han and Han, Song and Xie, Enze},
  booktitle={2025 IEEE/CVF International Conference on Computer Vision (ICCV)},
  pages={16185--16195},
  year={2025},
  organization={IEEE}
}

@article{xie2024sana,
  title={Sana: Efficient high-resolution image synthesis with linear diffusion transformers},
  author={Xie, Enze and Chen, Junsong and Chen, Junyu and Cai, Han and Tang, Haotian and Lin, Yujun and Zhang, Zhekai and Li, Muyang and Zhu, Ligeng and Lu, Yao and others},
  journal={arXiv preprint arXiv:2410.10629},
  year={2024}
}

@article{chen2025janus,
  title={Janus-Pro: Unified Multimodal Understanding and Generation with Data and Model Scaling},
  author={Chen, Xiaokang and Wu, Zhiyu and Liu, Xingchao and Pan, Zizheng and Liu, Wen and Xie, Zhenda and Yu, Xingkai and Ruan, Chong},
  journal={arXiv preprint arXiv:2501.17811},
  year={2025}
}

@article{xin2025lumina,
  title={Lumina-dimoo: An omni diffusion large language model for multi-modal generation and understanding},
  author={Xin, Yi and Qin, Qi and Luo, Siqi and Zhu, Kaiwen and Yan, Juncheng and Tai, Yan and Lei, Jiayi and Cao, Yuewen and Wang, Keqi and Wang, Yibin and others},
  journal={arXiv preprint arXiv:2510.06308},
  year={2025}
}

@article{team2025longcat,
  title={Longcat-image technical report},
  author={Team, Meituan LongCat and Ma, Hanghang and Tan, Haoxian and Huang, Jiale and Wu, Junqiang and He, Jun-Yan and Gao, Lishuai and Xiao, Songlin and Wei, Xiaoming and Ma, Xiaoqi and others},
  journal={arXiv preprint arXiv:2512.07584},
  year={2025}
}

@article{wang2026deepgen,
  title={Deepgen 1.0: A lightweight unified multimodal model for advancing image generation and editing},
  author={Wang, Dianyi and Li, Ruihang and Han, Feng and Ma, Chaofan and Song, Wei and Wang, Siyuan and Wang, Yibin and Xin, Yi and Liu, Hongjian and Zhang, Zhixiong and others},
  journal={arXiv preprint arXiv:2602.12205},
  year={2026}
}

@article{tian2026internvl,
  title={Internvl-u: Democratizing unified multimodal models for understanding, reasoning, generation and editing},
  author={Tian, Changyao and Yang, Danni and Chen, Guanzhou and Cui, Erfei and Wang, Zhaokai and Duan, Yuchen and Yin, Penghao and Chen, Sitao and Yang, Ganlin and Liu, Mingxin and others},
  journal={arXiv preprint arXiv:2603.09877},
  year={2026}
}

@article{alekseenko2026vibe,
  title={VIBE: Visual instruction based editor},
  author={Alekseenko, Grigorii and Gordeev, Aleksandr and Tolstykh, Irina and Suleimanov, Bulat and Dokholyan, Vladimir and Fedorov, Georgii and Yakubson, Sergey and Tsybina, Aleksandra and Chernyshov, Mikhail and Kuprashevich, Maksim},
  journal={arXiv preprint arXiv:2601.02242},
  year={2026}
}

@misc{blackforestlabs2026flux2klein,
  author       = {{Black Forest Labs}},
  title        = {{FLUX.2 [klein] 4B}},
  year         = {2026},
  howpublished = {\url{https://huggingface.co/black-forest-labs/FLUX.2-klein-4B}},
  note         = {Hugging Face model card}
}

@article{fu2026lance,
  title={Lance: Unified multimodal modeling by multi-task synergy},
  author={Fu, Fengyi and Huang, Mengqi and Wu, Shaojin and Jiang, Yunsheng and Huo, Yufei and Li, Hao and Song, Yinghang and Ding, Fei and Guo, Jianzhu and He, Qian and others},
  journal={arXiv preprint arXiv:2605.18678},
  year={2026}
}

@article{wang2025ovis,
  title={Ovis-u1 technical report},
  author={Wang, Guo-Hua and Zhao, Shanshan and Zhang, Xinjie and Cao, Liangfu and Zhan, Pengxin and Duan, Lunhao and Lu, Shiyin and Fu, Minghao and Chen, Xiaohao and Zhao, Jianshan and others},
  journal={arXiv preprint arXiv:2506.23044},
  year={2025}
}

@article{deng2025bagel,
  title   = {Emerging Properties in Unified Multimodal Pretraining},
  author  = {Deng, Chaorui and Zhu, Deyao and Li, Kunchang and Gou, Chenhui and Li, Feng and Wang, Zeyu and Zhong, Shu and Yu, Weihao and Nie, Xiaonan and Song, Ziang and Shi, Guang and Fan, Haoqi},
  journal = {arXiv preprint arXiv:2505.14683},
  year    = {2025}
}

@misc{google2026nanobanana2,
  author       = {{Google DeepMind}},
  title        = {Nano Banana 2 (Gemini 3.1 Flash Image)},
  year         = {2026},
  howpublished = {\url{https://deepmind.google/models/gemini-image/}},
  note         = {Accessed: 2026-10-06}
}

@inproceedings{cai-etal-2025-bayesian,
    title = "{B}ayesian Optimization for Controlled Image Editing via {LLM}s",
    author = "Cai, Chengkun  and
      Liu, Haoliang  and
      Zhao, Xu  and
      Jiang, Zhongyu  and
      Zhang, Tianfang  and
      Wu, Zongkai  and
      Lee, John  and
      Hwang, Jenq-Neng  and
      Li, Lei",
    editor = "Che, Wanxiang  and
      Nabende, Joyce  and
      Shutova, Ekaterina  and
      Pilehvar, Mohammad Taher",
    booktitle = "Findings of the Association for Computational Linguistics: ACL 2025",
    month = jul,
    year = "2025",
    address = "Vienna, Austria",
    publisher = "Association for Computational Linguistics",
    url = "https://aclanthology.org/2025.findings-acl.523/",
    doi = "10.18653/v1/2025.findings-acl.523",
    pages = "10045--10056",
    ISBN = "979-8-89176-256-5"
}

@inproceedings{feng2026vitaglyph,
  title={Vitaglyph: Vitalizing artistic typography with flexible dual-branch diffusion models},
  author={Feng, Kailai and Zhang, Yabo and Yu, Haodong and Ji, Zhilong and Bai, Jinfeng and Zhang, Hongzhi and Zuo, Wangmeng},
  booktitle={2026 IEEE/CVF Winter Conference on Applications of Computer Vision (WACV)},
  pages={8220--8230},
  year={2026},
  organization={IEEE}
}

@inproceedings{feng2024multi,
  title={Multi-attentional distance for zero-shot classification with text-to-image diffusion model},
  author={Feng, Kailai and Ni, Minheng and Jiang, Jiaxiu and Zhang, Zhilu and Zuo, Wangmeng},
  booktitle={2024 IEEE International Conference on Multimedia and Expo (ICME)},
  pages={1--6},
  year={2024},
  organization={IEEE}
}

@inproceedings{liu2026instancecontrol,
  title={InstanceControl: Controllable Complex Image Generation Without Instance Labeling},
  author={Liu, Xiaoyu and Wang, Huan and Li, Fan and Wang, Zhixin and Xu, Jiaqi and Liu, Ming and Zuo, Wangmeng},
  booktitle={European Conference on Computer Vision},
  pages={146--164},
  year={2026},
  organization={Springer}
}
\bibliographystyle{iclr2027_conference}

\clearpage
\appendix

\section{User Survey and Benchmark Construction}
\label{app:benchmark}

\subsection{User Survey}
\label{app:survey}

\paragraph{Survey overview.}
We collected 1,000 anonymous responses on mobile image AI.
Fig.~\ref{fig:survey_bmk_sta}(a) reports interest, use cases and priorities.
Interest in on-device image AI is 68\% Yes, 24\% Maybe and 8\% No.
Tab.~\ref{tab:app-survey-questionnaire} presents the questionnaire structure and one example scenario.
Q1 asks about current usage and interest, and Q2 asks about desired use cases.
Q3 contains 25 image-generation and editing scenarios.
Each scenario presents 10--15 requirements, including shared usability needs and task-specific requirements.
All respondents answer all 25 scenarios and select up to eight requirements per scenario.
Selection rates use all 1,000 respondents as the denominator.
For shared requirements, a respondent is counted once if they select the requirement in any scenario.
Selection rates describe the respondents rather than the wider population.
%

\begin{table}[p]
\centering
\caption{
Questionnaire structure and one of the 25 Q3 scenarios.
Q1 contains three single-choice questions.
Q2 allows up to eight selections.
Each Q3 scenario presents 10--15 requirements and allows up to eight selections.
}
\label{tab:app-survey-questionnaire}

\small
\setlength{\tabcolsep}{2pt}
\renewcommand{\arraystretch}{1}

\begin{tabularx}{\textwidth}{@{}p{0.25\textwidth}X@{}}
\toprule
\textbf{Topic} & \textbf{Question and response options} \\
\midrule

\textbf{Q1a. Current usage}
&
How often do you use AI tools on your phone,
such as chat assistants, image generators or photo editors?
\newline
\textit{Options:}
Daily;
Several times a week;
About once a week;
Less than once a week;
Never.
\\
\midrule

\textbf{Q1b. Image tools}
&
Would you like to generate or edit images using AI tools
on your phone?
\newline
\textit{Options:}
Yes;
Maybe;
No.
\\
\midrule

\textbf{Q1c. On-device interest}
&
Would you be interested in image AI that runs directly
on your phone without uploading your photos to a server?
\newline
\textit{Options:}
Yes;
Maybe;
No.
\\
\midrule

\textbf{Q2. Desired use cases}
&
Which of the following would you most like to use
AI image tools for on your phone?
Select up to eight options.
\newline
\textit{Options:}
 (1) Retouch portraits, such as removing blemishes or wrinkles.
 (2) Remove unwanted people or objects from photos.
 (3) Restore old, blurry or damaged photos.
 (4) Edit or translate text in images and documents.
 (5) Improve or creatively edit pet photos.
 (6) Make memes or stickers.
 (7) Create personal avatars or profile pictures.
 (8) Generate wallpapers.
 (9) Try different hairstyles, makeup or accessories.
 (10) Change clothing or preview outfit combinations.
 (11) Replace backgrounds in personal or product photos.
 (12) Turn photos into cartoons, paintings or other styles.
 (13) Create posters, invitations or greeting cards.
 (14) Improve product photos for online listings.
 (15) Preview furniture or decorations in a room.
 (16) Combine people or objects from several photos.

\\
\midrule

\textbf{Q3. Usability requirements}
&
Suppose you ask an AI tool to remove a small blemish
from your cheek in a portrait and leave the rest
of the photo unchanged.
What would matter to you when judging the result?
Select up to eight options.
\newline
\textit{Options:}
(1) Results are returned quickly.
(2) The requested blemish is removed.
(3) Photos remain private.
(4) Clothing, background and other people remain unchanged.
(5) The skin retains natural texture without visible artifacts.
(6) The tool requires little storage.
(7) The person remains recognisable, with facial features preserved.
(8) The tool works offline.
(9) Only the intended area is retouched.
(10) The overall photo remains clear and visually balanced.
(11) The tool does not slow down the phone.
(12) The original pose, framing and positions remain unchanged.
(13) The retouched area matches the surrounding skin tone and lighting.

\\

\bottomrule
\end{tabularx}
\end{table}

\subsection{Task Definitions and Coverage}
\label{app:tasks}

An \textit{operation} specifies the requested change, and a \textit{target} identifies the content involved.
Tables~\ref{tab:app-operation-definitions} and~\ref{tab:app-target-definitions} define both axes.

\paragraph{Operations.}
The operation axis contains 16 units organized into five groups.
Reference-editing units specify the available base and reference images.
Replacement substitutes content; adjustment changes its attributes.
Outpainting extends image boundaries; inpainting fills masked regions within them.
For reference editing, $B1$ denotes one base image and $Rk$ denotes $k$ reference images.

\begingroup
\small
\setlength{\tabcolsep}{3pt}
\renewcommand{\arraystretch}{1.08}
\setlength{\LTcapwidth}{\linewidth}
\begin{longtable}{@{}>{\raggedright\arraybackslash}p{0.16\linewidth}>{\raggedright\arraybackslash}p{0.15\linewidth}>{\raggedright\arraybackslash}p{\dimexpr0.69\linewidth-4\tabcolsep\relax}@{}}
\caption{Definitions of the 16 operation units.}\label{tab:app-operation-definitions}\\
\toprule
\textbf{Group} & \textbf{Operation} & \textbf{Definition} \\
\midrule
\endfirsthead
\toprule
\textbf{Group} & \textbf{Operation} & \textbf{Definition} \\
\midrule
\endhead
\bottomrule
\endfoot
Generation & T2I & Generate an image from text without an input image. \\
\midrule
\multirow[t]{4}{=}{Local editing} & Addition & Add specified content while retaining existing, unrelated content. \\*
 & Removal & Delete specified content and complete the exposed region. \\*
 & Replacement & Replace selected content with a specified alternative. \\*
 & Adjustment & Change attributes of existing content, such as color, size or pose. \\
\midrule
\multirow[t]{4}{=}{Global editing} & Stylization & Apply a visual style across the image. \\*
 & Background & Change the background while retaining the foreground subject. \\*
 & Restoration & Recover image clarity or appearance from degradation. \\*
 & Viewpoint & Change the camera view, or rectify document perspective. \\
\midrule
\multirow[t]{3}{=}{Understanding editing} & Semantic & Infer the required edit from meaning, context or an intended outcome. \\*
 & Outpainting & Extend content beyond the original image boundaries. \\*
 & Inpainting & Reconstruct missing content within supplied masks. \\
\midrule
\multirow[t]{4}{=}{Reference editing} & R1 & Generate an image from one reference, without a base image. \\*
 & B1+R1 & Edit a base image using content or attributes from one reference. \\*
 & B1+R2 & Edit a base image by combining two role-specific references. \\*
 & B1+R3 & Edit a base image by combining three role-specific references. \\
\end{longtable}
\endgroup

\paragraph{Targets.}
The eight categories identify the primary content to generate or modify, not every visible entity.
For example, changing a jacket targets Clothing \& Accessories; removing one person from a group targets Multi-Subjects.

\begingroup
\small
\setlength{\tabcolsep}{1pt}
\renewcommand{\arraystretch}{1}
\setlength{\LTcapwidth}{\linewidth}
\begin{longtable}{@{}>{\raggedright\arraybackslash}p{0.16\linewidth}>{\raggedright\arraybackslash}p{0.23\linewidth}>{\raggedright\arraybackslash}p{\dimexpr0.61\linewidth-4\tabcolsep\relax}@{}}
\caption{Definitions of the eight target categories.}\label{tab:app-target-definitions}\\
\toprule
\textbf{Group} & \textbf{Target} & \textbf{Definition} \\
\midrule
\endfirsthead
\toprule
\textbf{Group} & \textbf{Target} & \textbf{Definition} \\
\midrule
\endhead
\bottomrule
\endfoot
\multirow[t]{2}{=}{General Objects} & Concrete Objects & Discrete entities such as vehicles, furniture and everyday objects. \\*
 & Abstract Objects & Visual effects and relations, such as lighting, shadows and reflections. \\
\midrule
Animals & Animals & Animals, including pets, their appearance and their accessories. \\
\midrule
\multirow[t]{4}{=}{Portraits} & Body \& Limbs & Human body shape, limb configuration and pose. \\*
 & Face \& Hair & Facial features, expressions, makeup, hairstyles and identity. \\*
 & Clothing \& Accessories & Human garments and accessories, such as coats, glasses and jewelry. \\*
 & Multi-Subjects & Groups of people and edits to selected individuals within a group. \\
\midrule
Text \& Doc. & Text \& Documents & Text content, typography, layout and document appearance. \\
\end{longtable}
\endgroup

\paragraph{Task coverage and examples.}

Table~\ref{tab:app-task-examples} provides task definitions and illustrative instructions, not verbatim benchmark annotations.
N/A marks combinations excluded from the benchmark scope, such as adding or removing anatomical parts; it does not imply that such requests are impossible.
Included tasks unsupported by a model instead receive zero scores.
Each merged cell counts as one shared subtask.
The benchmark contains 8 generation, 28 local-editing, 7 global-editing, 5 understanding-editing and 16 reference-editing subtasks, totaling 64.

\begin{landscape}
\begingroup

\ifdefined\mogeOpWidth\else\newlength{\mogeOpWidth}\fi
\ifdefined\mogeCellWidth\else\newlength{\mogeCellWidth}\fi

\providecommand{\mogeTaskCount}[3]{%
  {\bfseries #1}\par
  {\itshape ``#2''}\unskip\nobreak\hfill
  \mbox{\scriptsize\normalfont $n=#3$}\par
}

\centering
\footnotesize
\setlength{\tabcolsep}{3pt}
\setlength{\arrayrulewidth}{0.35pt}
\renewcommand{\arraystretch}{1.05}
\setlength{\mogeOpWidth}{0.083\linewidth}
\setlength{\mogeCellWidth}{%
  \dimexpr(\linewidth-\mogeOpWidth-18\tabcolsep-10\arrayrulewidth)/8\relax
}

\captionof{table}{
Operation $\times$ Target task definitions and illustrative instructions (1/2).
Each ordinary or merged cell is one subtask; N/A cells are excluded.
$n$ gives the subtask sample count (64 subtasks; 1,150 samples).
$B1$ denotes one base image and $Rk$ denotes $k$ reference images.
}
\label{tab:app-task-examples}

\begin{adjustbox}{max width=\linewidth,max totalheight=0.82\textwidth}
\begin{tabular}{|>{\raggedright\arraybackslash}p{\mogeOpWidth}|*{8}{>{\raggedright\arraybackslash}p{\mogeCellWidth}|}}
\hline
\textbf{Operation}
& \multicolumn{2}{c|}{\textbf{General Objects}}
& \textbf{Animals}
& \multicolumn{4}{c|}{\textbf{Portraits}}
& \textbf{Text \& Docs.} \\
\cline{2-3}\cline{5-8}
& Concrete Objects
& Abstract Objects
& Animals
& Body \& Limbs
& Face \& Hair
& Clothing \& Accs.
& Multi-Subjects
& Text \& Docs. \\
\hline

\multicolumn{9}{|l|}{\textbf{Generation}} \\
\hline
\textbf{T2I}
& \mogeTaskCount{Object generation}{A bicycle beside a fence.}{20}
& \mogeTaskCount{Abstract visuals}{A blue abstract wallpaper.}{20}
& \mogeTaskCount{Animal generation}{A dog resting on grass.}{15}
& \mogeTaskCount{Full-body portrait}{A full-body portrait of a runner.}{20}
& \mogeTaskCount{Portrait close-up}{A close-up of a curly-haired woman.}{20}
& \mogeTaskCount{Outfit generation}{A person wearing a blue coat.}{10}
& \mogeTaskCount{Group portrait}{Four friends playing cards.}{15}
& \mogeTaskCount{Text-bearing image}{A poster titled Summer Fair.}{30} \\
\hline

\multicolumn{9}{|l|}{\textbf{Local Editing}} \\
\hline
\textbf{Addition}
& \mogeTaskCount{Add an object}{Add a vase to the table.}{20}
& \mogeTaskCount{Add an effect}{Add a sunbeam through the window.}{15}
& \mogeTaskCount{Add an animal}{Add a parrot to the chair.}{15}
& \textit{N/A}
& \mogeTaskCount{Add a facial feature}{Add a grey beard.}{30}
& \mogeTaskCount{Add an accessory}{Add a floral hairpin.}{25}
& \mogeTaskCount{Add a group member}{Add a person on the left.}{15}
& \mogeTaskCount{Add text}{Add OPEN to the sign.}{20} \\
\hline

\textbf{Removal}
& \mogeTaskCount{Remove an object}{Remove the cup.}{25}
& \mogeTaskCount{Remove an effect}{Remove the window glare.}{15}
& \mogeTaskCount{Remove an animal}{Remove the dog.}{10}
& \textit{N/A}
& \mogeTaskCount{Remove a facial detail}{Remove the cheek blemish.}{30}
& \mogeTaskCount{Remove an accessory}{Remove the sunglasses.}{20}
& \mogeTaskCount{Remove a group member}{Remove the leftmost person.}{25}
& \mogeTaskCount{Remove text}{Remove the poster title.}{20} \\
\hline

\textbf{Replace\-ment}
& \mogeTaskCount{Replace an object}{Replace the chair with a stool.}{20}
& \textit{N/A}
& \mogeTaskCount{Replace an animal}{Replace the dog with a cat.}{10}
& \textit{N/A}
& \mogeTaskCount{Replace a hairstyle}{Replace long hair with a bob.}{30}
& \mogeTaskCount{Replace clothing}{Replace the coat with a jacket.}{25}
& \mogeTaskCount{Replace a group member}{Replace the middle person with a woman.}{15}
& \mogeTaskCount{Replace text}{Replace SALE with OPEN.}{20} \\
\hline

\textbf{Adjustment}
& \mogeTaskCount{Adjust an attribute}{Make the car blue.}{20}
& \mogeTaskCount{Adjust illumination}{Make the lighting warmer.}{15}
& \mogeTaskCount{Adjust animal appearance}{Make the fur orange.}{10}
& \mogeTaskCount{Adjust body pose}{Raise the left arm.}{30}
& \mogeTaskCount{Adjust facial appearance}{Widen the smile slightly.}{30}
& \mogeTaskCount{Adjust an outfit}{Make the jacket green.}{20}
& \mogeTaskCount{Adjust within a group}{Raise the rightmost person's hand.}{15}
& \mogeTaskCount{Adjust typography}{Enlarge the title.}{15} \\
\hline

\multicolumn{9}{|l|}{\textbf{Global Editing}} \\
\hline
\textbf{Stylization}
& \multicolumn{8}{>{\raggedright\arraybackslash}p{\dimexpr 8\mogeCellWidth+14\tabcolsep+7\arrayrulewidth\relax}|}{%
  \mogeTaskCount{Apply an image-wide style}{Render the image in Lego style.}{30}
} \\
\hline

\textbf{Background}
& \multicolumn{8}{>{\raggedright\arraybackslash}p{\dimexpr 8\mogeCellWidth+14\tabcolsep+7\arrayrulewidth\relax}|}{%
  \mogeTaskCount{Change scene background}{Use a beach background; keep the person unchanged.}{35}
} \\
\hline
\end{tabular}
\end{adjustbox}

\par
\clearpage

\centering
\footnotesize

\captionof*{table}{
\tablename~\thetable:
Operation $\times$ Target task definitions and illustrative instructions (2/2, continued).
$n$ denotes the number of benchmark samples in each ordinary or merged task cell.
}

\begin{adjustbox}{max width=\linewidth,max totalheight=0.82\textwidth}
\begin{tabular}{|>{\raggedright\arraybackslash}p{\mogeOpWidth}|*{8}{>{\raggedright\arraybackslash}p{\mogeCellWidth}|}}
\hline
\textbf{Operation}
& \multicolumn{2}{c|}{\textbf{General Objects}}
& \textbf{Animals}
& \multicolumn{4}{c|}{\textbf{Portraits}}
& \textbf{Text \& Docs.} \\
\cline{2-3}\cline{5-8}
& Concrete Objects
& Abstract Objects
& Animals
& Body \& Limbs
& Face \& Hair
& Clothing \& Accs.
& Multi-Subjects
& Text \& Docs. \\
\hline

\multicolumn{9}{|l|}{\textbf{Global Editing (continued)}} \\
\hline
\textbf{Restoration}
& \multicolumn{3}{>{\raggedright\arraybackslash}p{\dimexpr 3\mogeCellWidth+4\tabcolsep+2\arrayrulewidth\relax}|}{%
  \mogeTaskCount{Restore scene clarity}{Denoise and sharpen this photo.}{30}
}
& \multicolumn{4}{>{\raggedright\arraybackslash}p{\dimexpr 4\mogeCellWidth+6\tabcolsep+3\arrayrulewidth\relax}|}{%
  \mogeTaskCount{Restore portrait quality}{Correct facial overexposure.}{25}
}
& \multicolumn{1}{>{\raggedright\arraybackslash}p{\mogeCellWidth}|}{%
  \mogeTaskCount{Restore document clarity}{Deblur this page.}{15}
} \\
\hline

\textbf{Viewpoint}
& \multicolumn{7}{>{\raggedright\arraybackslash}p{\dimexpr 7\mogeCellWidth+12\tabcolsep+6\arrayrulewidth\relax}|}{%
  \mogeTaskCount{Change camera view}{Show the scene from above.}{30}
}
& \multicolumn{1}{>{\raggedright\arraybackslash}p{\mogeCellWidth}|}{%
  \mogeTaskCount{Rectify a document}{Straighten this page.}{15}
} \\
\hline

\multicolumn{9}{|l|}{\textbf{Understanding Editing}} \\
\hline
\textbf{Semantic}
& \multicolumn{3}{>{\raggedright\arraybackslash}p{\dimexpr 3\mogeCellWidth+4\tabcolsep+2\arrayrulewidth\relax}|}{%
  \mogeTaskCount{Infer a contextual change}{Show the ice cream after sun exposure.}{30}
}
& \multicolumn{4}{>{\raggedright\arraybackslash}p{\dimexpr 4\mogeCellWidth+6\tabcolsep+3\arrayrulewidth\relax}|}{%
  \mogeTaskCount{Infer portrait appearance}{Show the person reacting to an unexpected gift.}{20}
}
& \multicolumn{1}{>{\raggedright\arraybackslash}p{\mogeCellWidth}|}{%
  \mogeTaskCount{Translate text}{Translate this title into English.}{20}
} \\
\hline

\textbf{Outpainting}
& \multicolumn{8}{>{\raggedright\arraybackslash}p{\dimexpr 8\mogeCellWidth+14\tabcolsep+7\arrayrulewidth\relax}|}{%
  \mogeTaskCount{Extend image boundaries}{Extend the scene on both sides.}{25}
} \\
\hline

\textbf{Inpainting}
& \multicolumn{8}{>{\raggedright\arraybackslash}p{\dimexpr 8\mogeCellWidth+14\tabcolsep+7\arrayrulewidth\relax}|}{%
  \mogeTaskCount{Complete masked regions}{Reconstruct the masked regions.}{25}
} \\
\hline

\multicolumn{9}{|l|}{\textbf{Reference Editing}} \\
\hline
\textbf{R1}
& \mogeTaskCount{Recreate an object}{Show this backpack on a desk.}{8}
& \textit{N/A}
& \mogeTaskCount{Recreate an animal}{Show this cat beside a window.}{5}
& \textit{N/A}
& \mogeTaskCount{Generate an identity-based portrait}{Create this person's headshot.}{10}
& \mogeTaskCount{Recreate an item}{Create a product photo of this jacket.}{5}
& \mogeTaskCount{Recreate a group}{Show this group in a park.}{4}
& \mogeTaskCount{Use a design reference}{Create a poster in this style.}{6} \\
\hline

\textbf{B1+R1}
& \mogeTaskCount{Transfer an object}{Replace the chair with this one.}{14}
& \mogeTaskCount{Transfer a style}{Apply the reference style.}{8}
& \mogeTaskCount{Transfer an animal}{Add this dog to the garden.}{7}
& \mogeTaskCount{Transfer a pose}{Copy the reference pose.}{8}
& \mogeTaskCount{Transfer a hairstyle}{Copy the reference hairstyle.}{14}
& \mogeTaskCount{Transfer an outfit}{Use the reference outfit.}{15}
& \mogeTaskCount{Insert a reference person}{Add this person to the group.}{8}
& \mogeTaskCount{Transfer a layout}{Copy the reference layout.}{8} \\
\hline

\textbf{B1+R2}
& \multicolumn{8}{>{\raggedright\arraybackslash}p{\dimexpr 8\mogeCellWidth+14\tabcolsep+7\arrayrulewidth\relax}|}{%
  \mogeTaskCount{Combine two references}{Use the hairstyle from reference 1 and coat from reference 2.}{12}
} \\
\hline

\textbf{B1+R3}
& \multicolumn{8}{>{\raggedright\arraybackslash}p{\dimexpr 8\mogeCellWidth+14\tabcolsep+7\arrayrulewidth\relax}|}{%
  \mogeTaskCount{Combine three references}{Use the hairstyle from reference 1, coat from reference 2 and background from reference 3.}{8}
} \\
\hline
\end{tabular}
\end{adjustbox}

\par
\endgroup
\end{landscape}
\section{Evaluation Protocol and Experimental Settings}
\label{app:protocol}

\subsection{Task-Specific Rubrics and Criterion Applicability}
\label{app:rubrics}

A \textit{criterion} specifies a quality requirement for an output.
MoI-Eval defines eight criteria under instruction following (IF), visual quality (VQ) and detail preservation (DP) (Tab.~\ref{tab:app-criterion-definitions}).
Preservation applies only to content or attributes not targeted by the instruction.

\begingroup
\small
\setlength{\tabcolsep}{3pt}
\renewcommand{\arraystretch}{1.08}
\setlength{\LTcapwidth}{\linewidth}
\begin{longtable}{@{}
>{\raggedright\arraybackslash}p{0.1\linewidth}
>{\raggedright\arraybackslash}p{0.15\linewidth}
>{\raggedright\arraybackslash}p{\dimexpr0.75\linewidth-4\tabcolsep\relax}
@{}}
\caption{Definitions of the eight evaluation criteria.}
\label{tab:app-criterion-definitions}\\
\toprule
\textbf{Group} & \textbf{Criterion} & \textbf{Definition} \\
\midrule
\endfirsthead
\toprule
\textbf{Group} & \textbf{Criterion} & \textbf{Definition} \\
\midrule
\endhead
\bottomrule
\endfoot

\multirow[t]{2}{=}{IF}
& Fidelity
& Satisfying the requested outcome and explicit constraints. \\*
& Selection
& Applying the change to the correct region or instance. \\

\midrule
\multirow[t]{3}{=}{VQ}
& Integrity
& Well-formed generated or edited content without defects or artifacts. \\*
& Coherence
& Consistent boundaries, lighting and appearance within the output and around edited regions. \\*
& Presentation
& Overall readability, composition and physical plausibility. \\

\midrule
\multirow[t]{3}{=}{DP}
& Context
& Preserving non-target objects and background details. \\*
& Structure
& Preserving framing, positions and spatial relations not targeted by the edit. \\*
& Identity
& Preserving human identity and appearance features not targeted by the edit. \\

\end{longtable}
\endgroup

\paragraph{Task-specific rubrics.}
Each entry $(o,t,c)$ defines criterion $c$ for operation $o$ and target $t$.
For example, Fidelity checks text removal or correct replacement text, depending on the operation.
For person removal, Identity checks the remaining people.
These requirements guide sample-grounded questions.

\paragraph{Criterion applicability.}
We manually label each rubric entry as:
\begin{itemize}
    \item \textit{Always applicable.} Required for every sample in the subtask, such as Fidelity in text replacement.
    \item \textit{Sample-dependent.} Determined by the instruction and input images. For example, Identity applies to object editing when people must remain unchanged.
    \item \textit{Not applicable.} Excluded from the subtask. For example, text-to-image generation has no source content to preserve, so DP is excluded.
\end{itemize}
We include always-applicable criteria and resolve sample-dependent criteria for each sample.
The selected criteria determine its questions and criterion groups.

Non-applicable criteria are excluded, not scored as zero.

\subsection{Sample-Grounded Evaluation and Scoring}
\label{app:questions}

\paragraph{Complete sample example.}
We use sample \texttt{local\_remove\_portrait\_multi\_010}.
The task is Removal $\times$ Multi-Subjects.
Its instruction is: ``Remove the leftmost person from the group photo and reconstruct the occluded background.''
The input shows a group of people on a street.
All eight criteria apply: the edit must remove the specified person, repair the exposed region and preserve the remaining scene and people.
Tab.~\ref{tab:app-complete-question-example} lists all 14 questions and their GPT-4o scores for the Ovis-U1-3B output.
Rubric summaries use the paper's criterion names.
Some question wording is edited for clarity; scores are reproduced from the original evaluation record.

\begingroup
\small
\setlength{\tabcolsep}{3pt}
\renewcommand{\arraystretch}{1.08}
\setlength{\LTcapwidth}{\linewidth}
\begin{longtable}{@{}>{\raggedright\arraybackslash}p{0.17\linewidth}>{\raggedright\arraybackslash}p{\dimexpr0.73\linewidth-4\tabcolsep\relax}r@{}}
\caption{A complete sample-grounded question set. Scores refer to one Ovis-U1-3B output evaluated by GPT-4o.}\label{tab:app-complete-question-example}\\
\toprule
Criterion & Rubric focus and questions & Scores \\
\midrule
\endfirsthead
\toprule
Criterion & Rubric focus and questions & Scores \\
\midrule
\endhead
\bottomrule
\endfoot
Fidelity & \textit{Remove the requested person.}\par (1) Is the leftmost person completely removed from the image without leaving visible remnants? & 0.6 \\
\midrule
Selection & \textit{Edit the correct person only.}\par (1) Did the removal target only the leftmost person (man in dark t-shirt), leaving the adjacent man in the white t-shirt untouched? & 1 \\
\midrule
Integrity & \textit{Reconstruct complete shapes and clean textures.}\par (1) Is the reconstructed background on the far left structurally complete without holes, repeated parts, or malformed shapes?\par (2) Are the textures in the repaired region on the left free of smearing, ringing, or synthetic artifacts? & 1, 0.6 \\
\midrule
Coherence & \textit{Match the surrounding geometry, colour and lighting.}\par (1) Does the reconstructed region on the left continue the surrounding street and background geometry without a visible seam?\par (2) Do the color and lighting in the filled area on the left match the neighboring image area? & 1, 0.6 \\
\midrule
Presentation & \textit{Maintain overall clarity and plausible composition.}\par (1) Is the final image clear and usable without awkward crop or weak focus?\par (2) Is the overall scene physically plausible and balanced in scale and composition? & 1, 1 \\
\midrule
Context & \textit{Preserve the unedited background.}\par (1) Are the buildings in the background unchanged in appearance?\par (2) Is the street surface outside the removed region preserved in color, texture, and lighting? & 1, 1 \\
\midrule
Structure & \textit{Preserve framing and spatial relations.}\par (1) Are the original framing, composition, and camera view of the street scene preserved?\par (2) Are the positions and spatial relationships of the remaining people in the group preserved? & 1, 1 \\
\midrule
Identity & \textit{Preserve the remaining people.}\par (1) Does each remaining person retain the same identity as in the input image?\par (2) Are the remaining people's clothing, poses, and accessories preserved? & 1, 1 \\
\end{longtable}
\endgroup

The criterion means are $0.6$ for Fidelity, $1$ for Selection, $0.8$ for Integrity and Coherence, and $1$ for the other criteria.
The group scores are $\mathrm{IF}=0.8$, $\mathrm{VQ}=0.867$ and $\mathrm{DP}=1$.
Using weights $(0.45,0.25,0.30)$ gives a sample quality score of $0.873$.

\paragraph{Aggregation weights.}
\label{app:aggregation}

Table~\ref{tab:app-aggregation-weights} lists operation-specific weights, fixed across models.
We exclude inapplicable groups and normalize the remaining weights for the sample-level geometric mean.
Within each group, applicable criteria receive equal weight regardless of question counts.

\begin{table}[t]
\centering
\small
\setlength{\tabcolsep}{6pt}
\renewcommand{\arraystretch}{1.08}
\caption{Criterion-group weights by operation.
A dash denotes an inapplicable group.}
\label{tab:app-aggregation-weights}
\begin{tabular}{@{}lrrr@{}}
\toprule
Operation & IF & VQ & DP \\
\midrule
Text-to-image                  & 0.60 & 0.40 & ---  \\
Addition, removal, replacement & 0.45 & 0.25 & 0.30 \\
Adjustment                     & 0.40 & 0.25 & 0.35 \\
Stylization, background editing & 0.45 & 0.35 & 0.20 \\
Restoration                    & 0.40 & 0.45 & 0.15 \\
Viewpoint editing              & 0.45 & 0.30 & 0.25 \\
Semantic editing               & 0.55 & 0.25 & 0.20 \\
Outpainting, inpainting         & 0.40 & 0.40 & 0.20 \\
Reference editing              & 0.50 & 0.25 & 0.25 \\
\bottomrule
\end{tabular}
\end{table}

\paragraph{Judge inputs and responses.}
Each judge call evaluates one question and returns a JSON score.
Editing inputs appear in their stored order, followed by the model output.
Text-to-image calls contain only the generated image.
The accompanying text provides the instruction, question and score conditions.
The following examples show judge prompts for base-image editing and text-to-image generation.
Braces mark sample-specific fields.
These examples do not enumerate all input configurations.

\begingroup
\lstset{
    basicstyle=\ttfamily\footnotesize, columns=fullflexible, breaklines=true, keepspaces=true, showstringspaces=false, frame=single, aboveskip=6pt, belowskip=6pt }

\begin{promptbox}{Editing system prompt example}
\raggedright
You are a strict visual judge for an image-editing benchmark.
Compare the ordered input image or images with the edited result and score only the stated atomic question.
Use the supplied score anchors exactly.
If the partial condition is N/A, return only 0 or 1.
Do not reward unrelated visual quality and do not speculate beyond visible evidence.
Output only one JSON object:
\texttt{\{"score": 0\}},
\texttt{\{"score": 0.6\}}, or
\texttt{\{"score": 1\}}.
\end{promptbox}

\paragraph{Base-image editing prompt example.}
For a base-image edit, \texttt{image\_description} states: ``Image 1 is the base input image; Image 2 is the edited result.''
For multiple inputs, it identifies the ordered reference images and final output.

\begin{promptbox}{Editing user prompt}
\raggedright
\texttt{\{image\_description\}}

\medskip
Editing instruction:\\
\texttt{"""}\\
\texttt{\{instruction\}}\\
\texttt{"""}

\medskip
Question to answer:\\
\texttt{"""}\\
\texttt{\{question\}}\\
\texttt{"""}

\medskip
Score 1 condition:\\
\texttt{\{success\_condition\}}

\medskip
Score 0.6 condition:\\
\texttt{\{partial\_condition\}}

\medskip
Score 0 condition:\\
\texttt{\{failure\_condition\}}

\medskip
Return only
\texttt{\{"score": 0\}},
\texttt{\{"score": 0.6\}}, or
\texttt{\{"score": 1\}}.
\end{promptbox}

\begin{promptbox}{Text-to-image system prompt}
\raggedright
You are a strict visual judge for a text-to-image benchmark.
Inspect the generated image against the prompt and score only the stated atomic question.
Use the supplied score anchors exactly.
If the partial condition is N/A, return only 0 or 1.
Do not speculate beyond visible evidence.
Output only one JSON object:
\texttt{\{"score": 0\}},
\texttt{\{"score": 0.6\}}, or
\texttt{\{"score": 1\}}.
\end{promptbox}

\paragraph{Text-to-image user prompt.}\mbox{}\par
\begin{promptbox}{Text-to-image user prompt}
\raggedright
The attached image is the GENERATED image.

\medskip
Generation prompt:\\
\texttt{"""}\\
\texttt{\{prompt\}}\\
\texttt{"""}

\medskip
Question to answer:\\
\texttt{"""}\\
\texttt{\{question\}}\\
\texttt{"""}

\medskip
Score 1 condition:\\
\texttt{\{success\_condition\}}

\medskip
Score 0.6 condition:\\
\texttt{\{partial\_condition\}}

\medskip
Score 0 condition:\\
\texttt{\{failure\_condition\}}

\medskip
Return only
\texttt{\{"score": 0\}},
\texttt{\{"score": 0.6\}}, or
\texttt{\{"score": 1\}}.
\end{promptbox}

\endgroup

\subsection{Additional Experimental Settings}
\label{app:configurations}

\paragraph{Output resolution.}
Text-to-image generation uses each model's default resolution; wallpaper prompts use a $9{:}16$ aspect ratio.
Non-outpainting editing requests use the base-image dimensions unless another canvas size is specified.
Outpainting requests use a 1:1 canvas by default unless the instruction specifies another size or aspect ratio.
Each pipeline retains its native preprocessing and resolution constraints.
Editing outputs are not resized after generation, except for judge preprocessing.

\paragraph{Judge image preprocessing.}
Both judges use identical preprocessing.
We resize input and output images to a maximum side length of 1,024 pixels with bicubic interpolation, preserving aspect ratios without upscaling.
Images are converted to RGB and encoded as JPEG at quality 88.
This affects only judge inputs, not saved model outputs.

\paragraph{Deployment settings.}
\label{app:deployment-protocol}
Deployment tests use the prompt: ``A cinematic portrait of a young woman surrounded by flowers, soft natural light.''
Within each model, the prompt, seed, resolution and sampling settings remain fixed across precision configurations.
The Android Janus-Pro deployment uses precomputed prompt embeddings, and its reported latency excludes prompt preparation.
Loading or offloading performed during generation remains part of measured end-to-end latency.
Failed configurations are excluded from successful-run averages.
Some precision configurations also change session partitioning or resource management, as noted in Tab.~\ref{tab:precision-comparison}; their timing differences cannot be attributed to precision alone.
The backbone and decoder precision experiments in Tab.~\ref{tab:precision-comparison} do not repeat the full benchmark quality evaluation, and recognizable outputs do not establish quality retention.

\subsection{Text-encoder quantization}
\label{sec:text-encoder-quantization}

We assess the effect of 4-bit NF4 text-encoder quantization on Ovis-U1-3B, DreamLite and SANA-600M.
For each model, we compare the original configuration with a configuration that quantizes only the text encoder, keeping the remaining inference settings unchanged.
Both configurations are evaluated using Gemini~3.1~Pro under the same MoI-Eval protocol.

\begin{table}[t]
\centering
\caption{
Effect of 4-bit NF4 text-encoder quantization on quality.
$\Delta Q$ is the quantized score minus the original score.
Coverage and tier transitions are shown from the original to the quantized configuration.
}
\label{tab:text-encoder-quantization}
\small
\setlength{\tabcolsep}{4pt}
\renewcommand{\arraystretch}{1.10}
\resizebox{\linewidth}{!}{%
\begin{tabular}{@{}lrrrrcc@{}}
\toprule
\multirow{2}{*}{Model}
& \multicolumn{2}{c}{Overall quality $Q$}
& \multirow{2}{*}{$\Delta Q$}
& \multirow{2}{*}{Relative change}
& \multirow{2}{*}{\shortstack{Subtasks with\\quality $\geq 0.6$}}
& \multirow{2}{*}{Q-tier} \\
\cmidrule(lr){2-3}
& Original & Text-NF4 & & & & \\
\midrule
Ovis-U1-3B
& 0.541734 & 0.538307 & $-0.003427$ & $-0.63\%$
& $24/64 \rightarrow 26/64$ & Q2 $\rightarrow$ Q2 \\
DreamLite
& 0.523777 & 0.520798 & $-0.002979$ & $-0.57\%$
& $27/64 \rightarrow 25/64$ & Q2 $\rightarrow$ Q2 \\
SANA-600M
& 0.102904 & 0.101979 & $-0.000925$ & $-0.90\%$
& $7/64 \rightarrow 7/64$ & Q1 $\rightarrow$ Q1 \\
\bottomrule
\end{tabular}%
}
\end{table}

Tab.~\ref{tab:text-encoder-quantization} shows small decreases in overall quality, with absolute changes below $0.0035$ and relative decreases below $1\%$ for all three models.
The number of subtasks meeting the $0.6$ quality threshold changes by at most two, and all quality tiers remain unchanged.
These results suggest that text-encoder quantization has a limited effect on benchmark-level quality for the tested models.

\section{Additional Quality Results}
\label{app:quality-results}

\subsection{Results Evaluated by Gemini}
\label{app:gemini-results}

\begin{table*}[t]
\centering
\caption{
Capability profiles evaluated by Gemini~3.1~Pro
(\texttt{gemini-3.1-p-priority}).
General and Text. denote general objects and text and documents.
IF, VQ and DP denote instruction following, visual quality and detail preservation.
Target scores use the same sample categories as the GPT-4o results.
Scores range from $0$ to $1$, with higher values indicating better performance.
}
\label{tab:app-gemini-profiles}

\small
\renewcommand{\arraystretch}{1.12}
\setlength{\tabcolsep}{2pt}

\resizebox{\textwidth}{!}{%
\begin{tabular}{@{}l*{12}{r}@{}}
\toprule
\multirow{2}{*}{Model}
& \multicolumn{5}{c}{Operation}
& \multicolumn{4}{c}{Target}
& \multicolumn{3}{c}{Criteria} \\
\cmidrule(lr){2-6}
\cmidrule(lr){7-10}
\cmidrule(lr){11-13}
& Gen. & Local & Global & Underst. & Ref.
& General & Animals & Portraits & Text.
& IF & VQ & DP \\
\midrule

\multicolumn{13}{@{}l}{\textit{Closed-source Models}} \\
\addlinespace[2pt]

Nano Banana 2 & 0.969 & 0.943 & 0.889 & 0.885 & 0.938 & 0.925 & 0.971 & 0.941 & 0.897 & 0.939 & 0.957 & 0.962 \\

GPT Image 2 & 0.981 & 0.968 & 0.927 & 0.935 & 0.967 & 0.954 & 0.988 & 0.967 & 0.934 & 0.962 & 0.981 & 0.964 \\

Seedream 5.0 Pro & 0.966 & 0.948 & 0.864 & 0.907 & 0.952 & 0.919 & 0.954 & 0.951 & 0.903 & 0.934 & 0.969 & 0.959 \\

\midrule
\multicolumn{13}{@{}l}{\textit{Open-weight Reference Models}} \\
\addlinespace[2pt]

Lumina-DiMOO-8B & 0.874 & 0.288 & 0.232 & 0.163 & 0.052 & 0.332 & 0.462 & 0.301 & 0.246 & 0.423 & 0.358 & 0.447 \\

BAGEL-7B-MoT & 0.838 & 0.696 & 0.533 & 0.467 & 0.137 & 0.578 & 0.725 & 0.641 & 0.444 & 0.671 & 0.642 & 0.772 \\

LongCat-Image & 0.911 & 0.000 & 0.000 & 0.000 & 0.000 & 0.106 & 0.182 & 0.109 & 0.147 & 0.121 & 0.118 & 0.000 \\

LongCat-Image-Edit & 0.000 & 0.879 & 0.721 & 0.654 & 0.191 & 0.646 & 0.638 & 0.670 & 0.484 & 0.640 & 0.695 & 0.832 \\

DeepGen 1.0 & 0.888 & 0.605 & 0.570 & 0.453 & 0.176 & 0.580 & 0.700 & 0.583 & 0.435 & 0.685 & 0.661 & 0.517 \\

InternVL-U-4B & 0.869 & 0.333 & 0.396 & 0.377 & 0.123 & 0.388 & 0.427 & 0.413 & 0.318 & 0.568 & 0.478 & 0.326 \\

VIBE-Image-Edit & 0.000 & 0.741 & 0.588 & 0.558 & 0.123 & 0.562 & 0.563 & 0.569 & 0.300 & 0.579 & 0.548 & 0.729 \\

OmniGen2 & 0.859 & 0.569 & 0.475 & 0.337 & 0.352 & 0.538 & 0.738 & 0.586 & 0.321 & 0.615 & 0.658 & 0.652 \\

FLUX.2 [klein] 4B & 0.900 & 0.795 & 0.805 & 0.669 & 0.700 & 0.812 & 0.902 & 0.811 & 0.598 & 0.791 & 0.847 & 0.931 \\

Lance-3B & 0.792 & 0.678 & 0.593 & 0.489 & 0.147 & 0.614 & 0.741 & 0.643 & 0.338 & 0.640 & 0.693 & 0.651 \\

\midrule
\multicolumn{13}{@{}l}{\textit{On-device Models}} \\
\addlinespace[2pt]

Ovis-U1-3B & 0.821 & 0.562 & 0.657 & 0.415 & 0.123 & 0.589 & 0.644 & 0.558 & 0.340 & 0.619 & 0.637 & 0.513 \\

Janus-Pro-1B & 0.610 & 0.000 & 0.000 & 0.000 & 0.000 & 0.088 & 0.149 & 0.078 & 0.036 & 0.089 & 0.076 & 0.000 \\

SANA-600M & 0.789 & 0.000 & 0.000 & 0.000 & 0.000 & 0.100 & 0.176 & 0.101 & 0.082 & 0.108 & 0.101 & 0.000 \\

SANA-Sprint-0.6B & 0.748 & 0.000 & 0.000 & 0.000 & 0.000 & 0.099 & 0.167 & 0.095 & 0.070 & 0.103 & 0.097 & 0.000 \\

DreamLite & 0.771 & 0.561 & 0.528 & 0.478 & 0.145 & 0.593 & 0.685 & 0.538 & 0.255 & 0.621 & 0.522 & 0.617 \\

Mobile-O-0.5B & 0.713 & 0.000 & 0.000 & 0.000 & 0.000 & 0.097 & 0.160 & 0.092 & 0.059 & 0.097 & 0.098 & 0.000 \\

\bottomrule
\end{tabular}%
}
\end{table*}

Gemini~3.1~Pro (\texttt{gemini-3.1-p-priority}) evaluates the same outputs with identical questions and scoring rules.
Table~\ref{tab:app-gemini-profiles} provides Gemini capability profiles beyond the overall results in Tab.~\ref{tab:quality_overall}.

\subsection{Detailed Capability Profiles}
\label{app:profiles}

\begin{table*}[t]
\centering
\caption{Scores for the eight evaluation criteria, evaluated by GPT-4o (\texttt{gpt-4o-2024-11-20}).
IF, VQ and DP denote instruction following, visual quality and detail preservation.
Scores range from $0$ to $1$, with higher values indicating better performance.
}
\label{tab:app-criterion-profiles}
\small
\renewcommand{\arraystretch}{1.12}
\setlength{\tabcolsep}{3pt}
\resizebox{\textwidth}{!}{%
\begin{tabular}{@{}l*{8}{r}@{}}
\toprule
\multirow{2}{*}{Model}
& \multicolumn{2}{c}{IF}
& \multicolumn{3}{c}{VQ}
& \multicolumn{3}{c}{DP} \\
\cmidrule(lr){2-3}
\cmidrule(lr){4-6}
\cmidrule(lr){7-9}
& Fidelity & Selection
& Integrity & Coherence & Presentation
& Context & Structure & Identity \\
\midrule
\multicolumn{9}{@{}l}{\textit{Closed-source Models}} \\
\addlinespace[2pt]
Nano Banana 2 & 0.941 & 0.952 & 0.980 & 0.956 & 0.978 & 0.976 & 0.964 & 0.986 \\
GPT Image 2 & 0.963 & 0.969 & 0.989 & 0.975 & 0.987 & 0.976 & 0.979 & 0.984 \\
Seedream 5.0 Pro & 0.945 & 0.961 & 0.983 & 0.959 & 0.978 & 0.970 & 0.979 & 0.984 \\
\midrule
\multicolumn{9}{@{}l}{\textit{Open-weight Reference Models}} \\
\addlinespace[2pt]
Lumina-DiMOO-8B~\cite{xin2025lumina} & 0.479 & 0.419 & 0.460 & 0.431 & 0.525 & 0.622 & 0.645 & 0.528 \\
BAGEL-7B-MoT~\cite{deng2025bagel} & 0.688 & 0.670 & 0.683 & 0.635 & 0.729 & 0.789 & 0.782 & 0.774 \\
LongCat-Image~\cite{team2025longcat} & 0.120 & 0.000 & 0.121 & 0.123 & 0.117 & 0.000 & 0.000 & 0.000 \\
LongCat-Image-Edit~\cite{team2025longcat} & 0.653 & 0.809 & 0.712 & 0.683 & 0.718 & 0.867 & 0.848 & 0.848 \\
DeepGen 1.0~\cite{wang2026deepgen} & 0.761 & 0.678 & 0.743 & 0.715 & 0.774 & 0.694 & 0.725 & 0.652 \\
InternVL-U-4B~\cite{tian2026internvl} & 0.667 & 0.457 & 0.552 & 0.528 & 0.571 & 0.415 & 0.515 & 0.538 \\
VIBE-Image-Edit~\cite{alekseenko2026vibe} & 0.597 & 0.725 & 0.637 & 0.607 & 0.658 & 0.820 & 0.822 & 0.790 \\
OmniGen2~\cite{wu2026omnigen2} & 0.684 & 0.580 & 0.722 & 0.645 & 0.726 & 0.711 & 0.748 & 0.743 \\
FLUX.2 [klein] 4B~\cite{blackforestlabs2026flux2klein} & 0.813 & 0.803 & 0.884 & 0.865 & 0.900 & 0.946 & 0.964 & 0.940 \\
Lance-3B~\cite{fu2026lance} & 0.681 & 0.703 & 0.744 & 0.726 & 0.745 & 0.788 & 0.742 & 0.772 \\
\midrule
\multicolumn{9}{@{}l}{\textit{On-device Models}} \\
\addlinespace[2pt]
Ovis-U1-3B~\cite{wang2025ovis} & 0.708 & 0.609 & 0.695 & 0.684 & 0.733 & 0.698 & 0.783 & 0.649 \\
Janus-Pro-1B~\cite{chen2025janus} & 0.088 & 0.000 & 0.072 & 0.082 & 0.082 & 0.000 & 0.000 & 0.000 \\
SANA-600M~\cite{xie2024sana} & 0.106 & 0.000 & 0.103 & 0.112 & 0.102 & 0.000 & 0.000 & 0.000 \\
SANA-Sprint-0.6B~\cite{chen2025sana} & 0.104 & 0.000 & 0.098 & 0.108 & 0.098 & 0.000 & 0.000 & 0.000 \\
DreamLite~\cite{feng2026dreamlite} & 0.664 & 0.604 & 0.629 & 0.596 & 0.682 & 0.709 & 0.828 & 0.732 \\
Mobile-O-0.5B~\cite{shaker2026mobile} & 0.097 & 0.000 & 0.104 & 0.108 & 0.098 & 0.000 & 0.000 & 0.000 \\
\bottomrule
\end{tabular}%
}
\end{table*}

Table~\ref{tab:app-criterion-profiles} reports all 19 models at three decimal places, complementing the selected models in the main text.

\subsection{Additional Qualitative Results}
\label{app:qualitative}

\begin{figure*}[t]
\centering
\includegraphics[width=\textwidth]{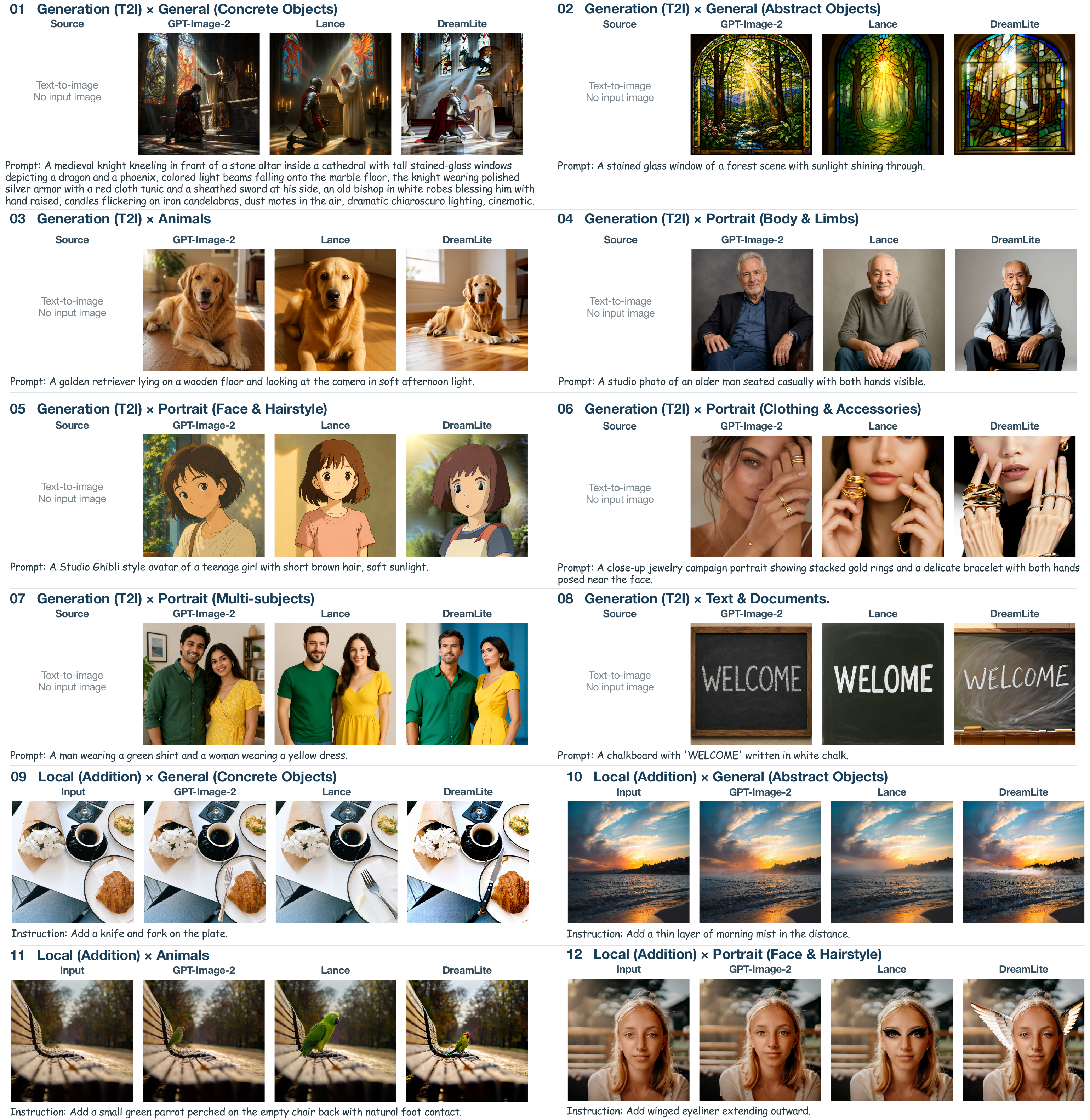}
\caption{
Additional qualitative comparisons for generation and local addition (examples 1--12).
}
\label{fig:app-qualitative-1}
\end{figure*}

\begin{figure*}[t]
\centering
\includegraphics[width=\textwidth]{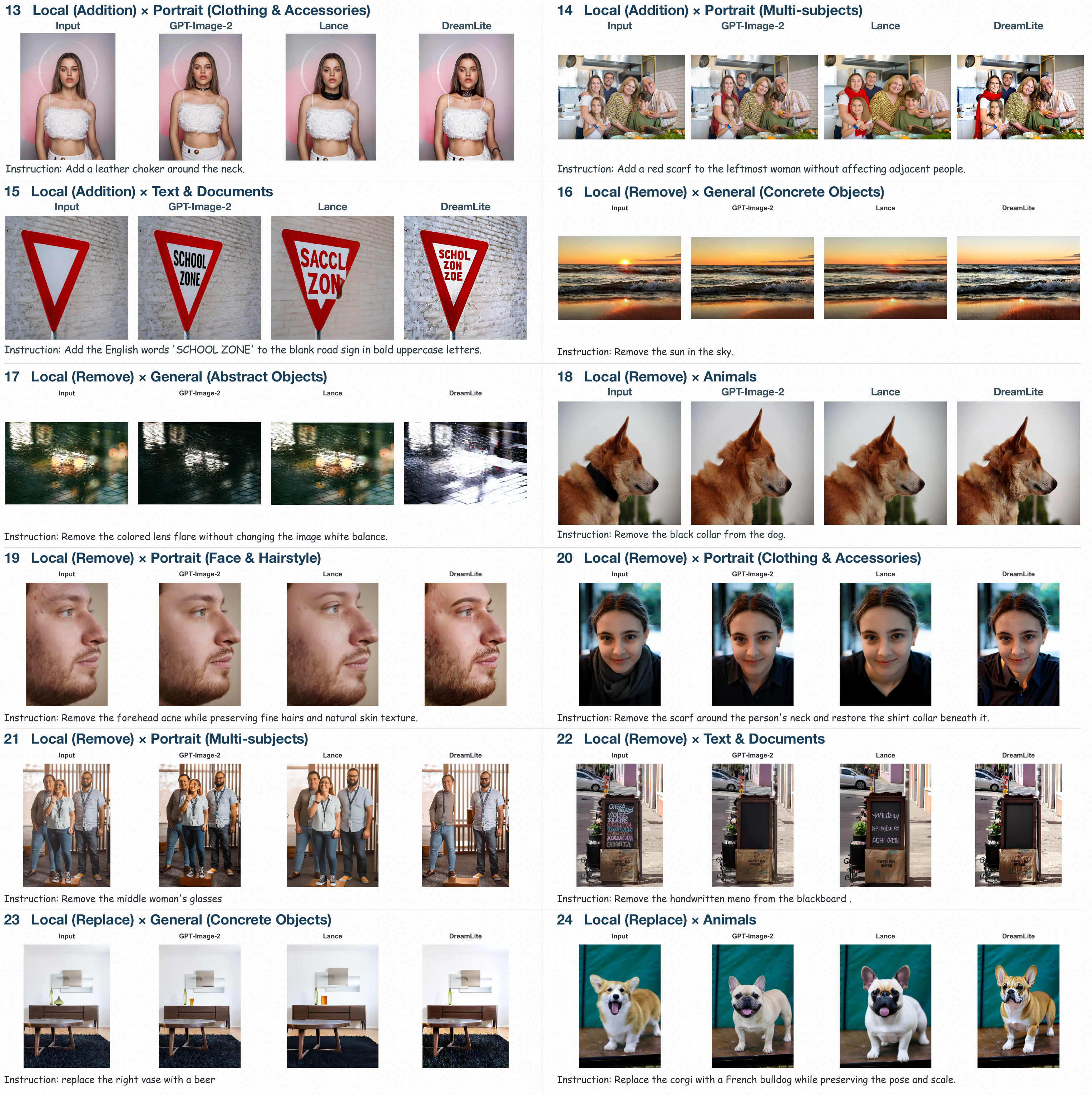}
\caption{
Additional qualitative comparisons for local edit (examples 13--24).
}
\label{fig:app-qualitative-2}
\end{figure*}

\begin{figure*}[t]
\centering
\includegraphics[width=\textwidth]{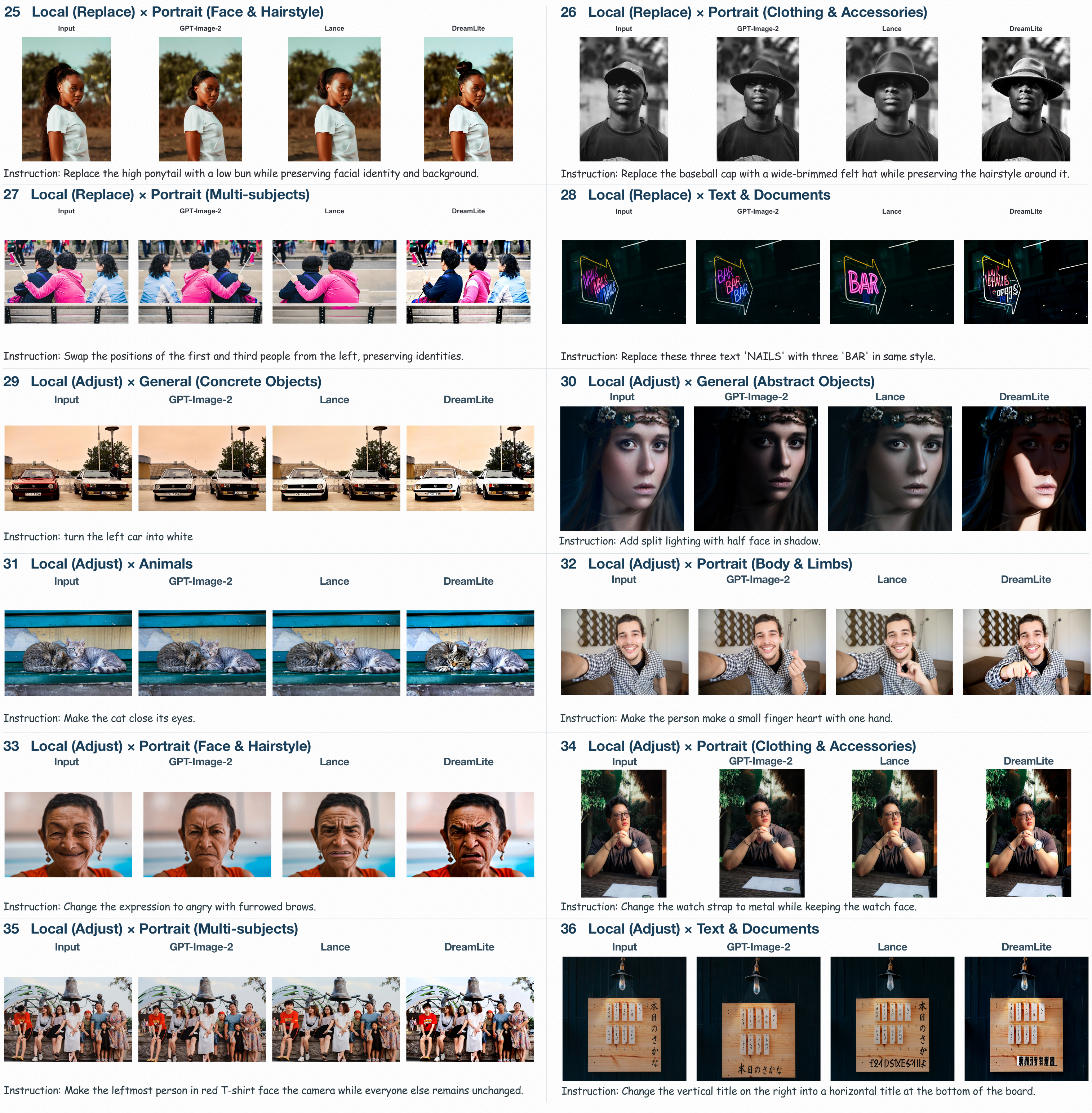}
\caption{
Additional qualitative comparisons for local edit (examples 25--36).
}
\label{fig:app-qualitative-3}
\end{figure*}

\begin{figure*}[t]
\centering
\includegraphics[width=\textwidth]{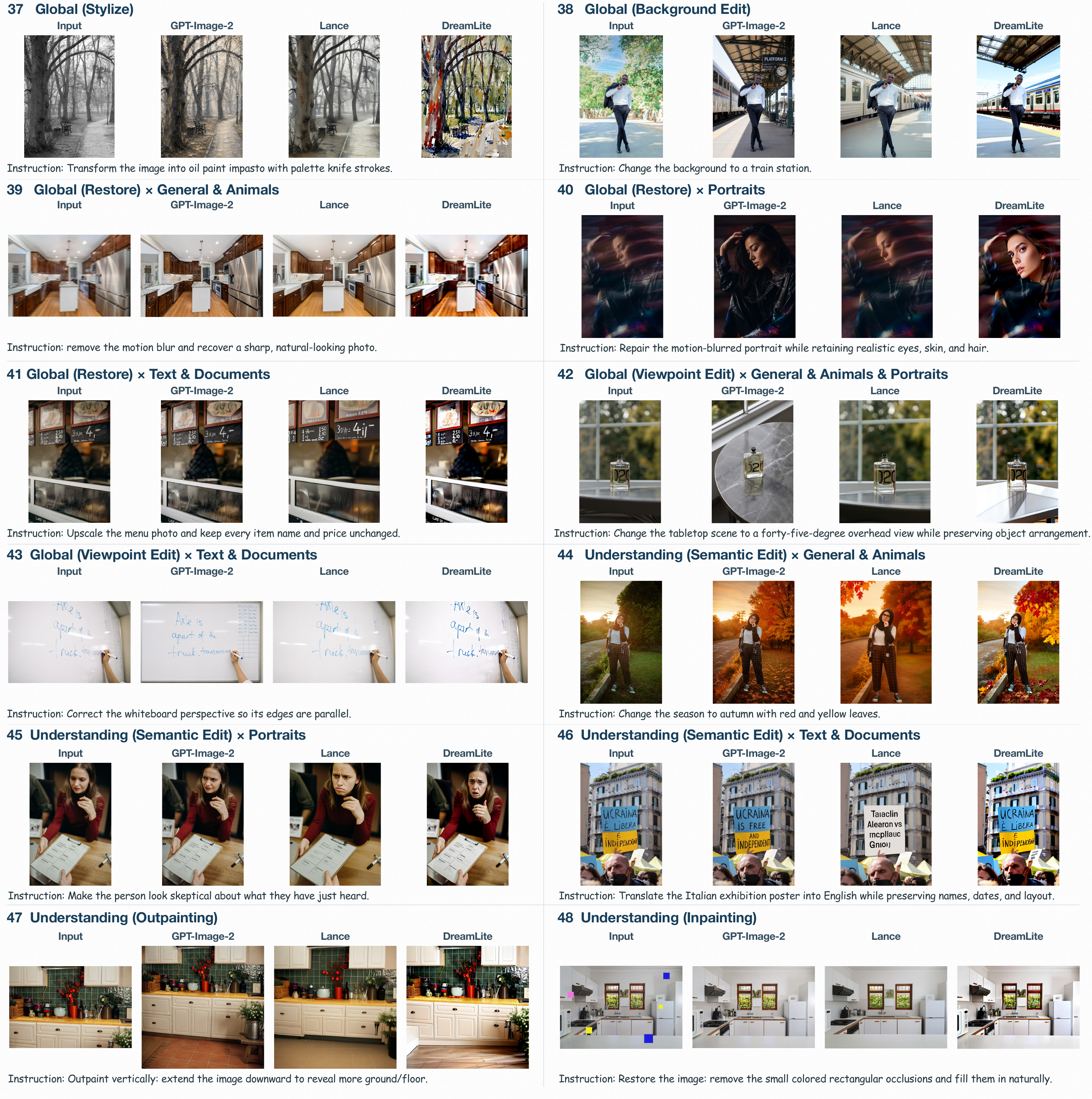}
\caption{
Additional qualitative comparisons for global edit and understanding edit (examples 37--48).
}
\label{fig:app-qualitative-4}
\end{figure*}

\begin{figure*}[t]
\centering
\includegraphics[width=\textwidth]{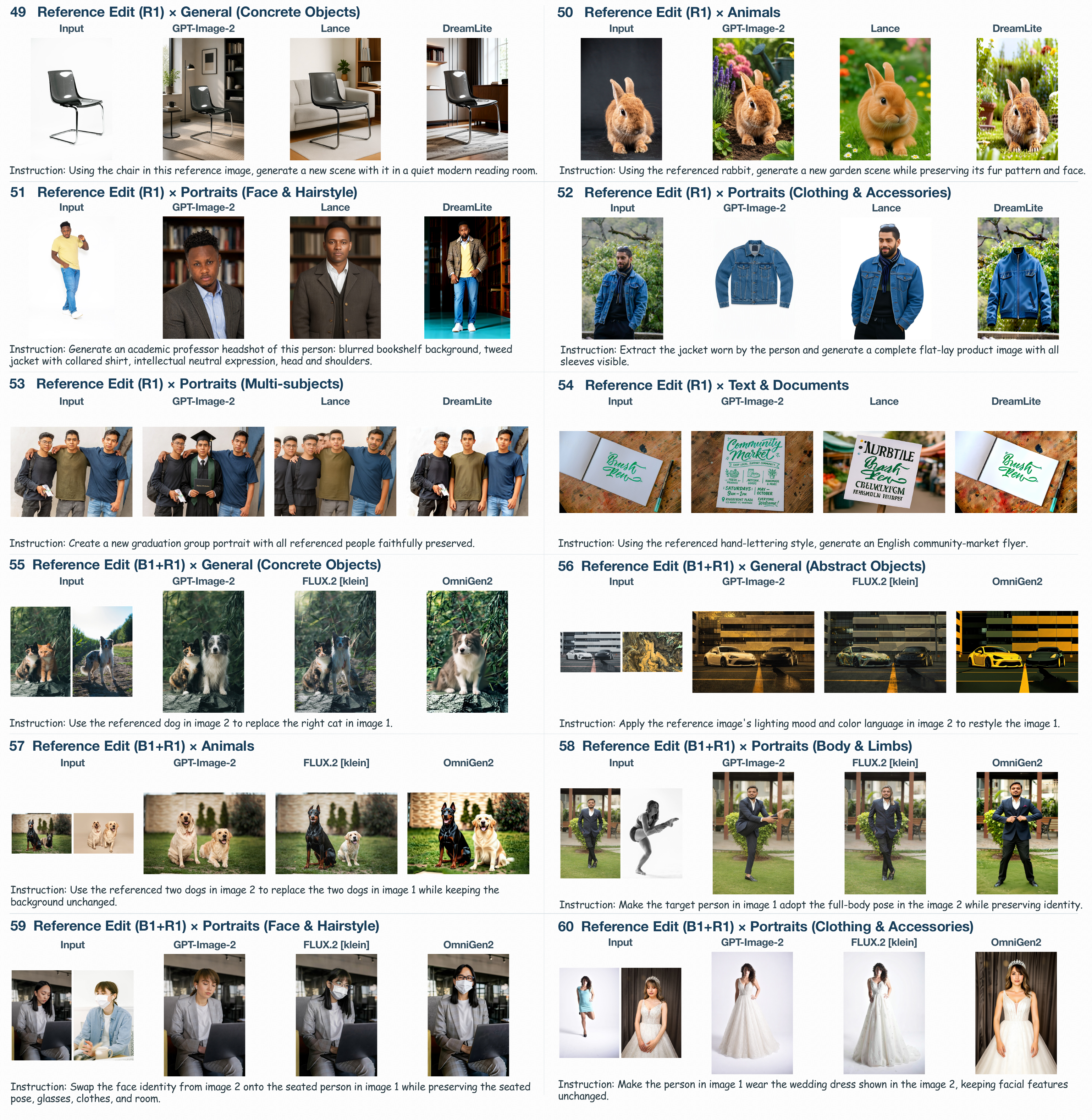}
\caption{
Additional qualitative comparisons for reference edit (examples 49--60).
}
\label{fig:app-qualitative-5}
\end{figure*}

\begin{figure*}[t]
\centering
\includegraphics[width=\textwidth]{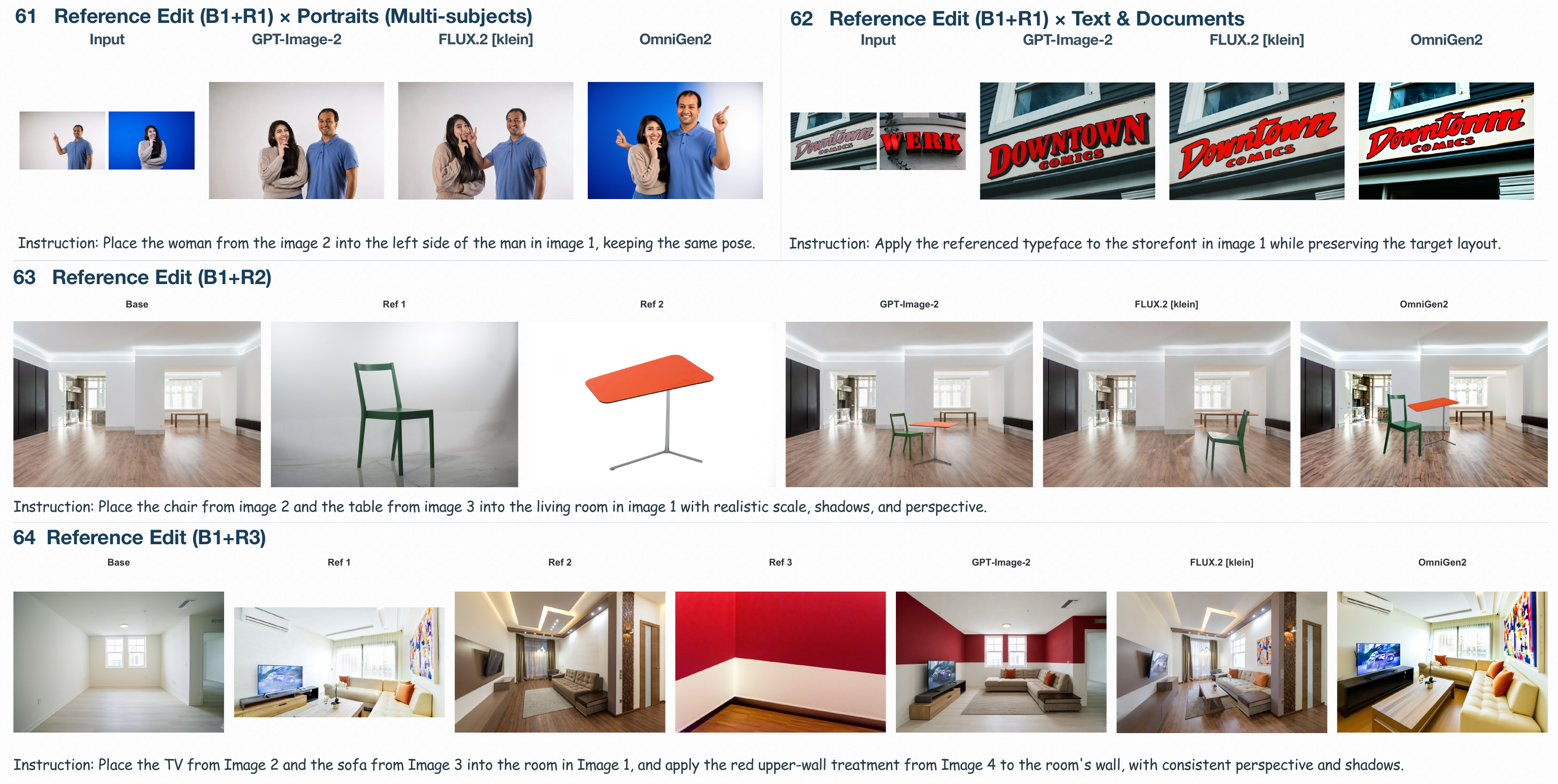}
\caption{
Additional qualitative comparisons for reference edit (examples 61--64).
}
\label{fig:app-qualitative-6}
\end{figure*}

Figs.~\ref{fig:app-qualitative-1},~\ref{fig:app-qualitative-2}, ~\ref{fig:app-qualitative-3}, ~\ref{fig:app-qualitative-4}, ~\ref{fig:app-qualitative-5}, ~\ref{fig:app-qualitative-6} compare GPT Image 2, FLUX.2, OmniGen2, Lance and DreamLite across generation and editing tasks.
DreamLite captures the main content in several generation examples, including the animal and two-person scenes.
Precise editing remains less consistent.

\section{Evaluation Reliability and Sensitivity}
\label{app:reliability}

\subsection{Cross-Judge Consistency}
\label{app:cross-judge}

\paragraph{Paired results.}
We compare 19 models with valid results from both judges, using the same outputs, questions and score anchors.
Sample-level comparisons retain outputs scored successfully by both judges, including valid zero scores.
Excluding preset zeros for unsupported tasks leaves 15,716 paired scores.

\paragraph{Ranking and score agreement.}
Overall quality rankings have a Spearman correlation of $\rho=0.968$, including unsupported-task zeros in $Q$.
Paired sample scores have $\rho=0.790$ and a mean absolute difference of $0.139$.
Gemini scores are lower by $0.051$ on average across the 19 models.

\paragraph{Quality-tier agreement.}
Both judges yield the same tiers under the final thresholds in Tab.~\ref{tab:quality_tiers}.
Because cross-judge agreement informed threshold calibration, this tier agreement is not an independent validation of judge reliability.

\subsection{Agreement with Human Judgments}
\label{app:human}
\paragraph{Human evaluation.}
We sample 500 outputs per editing-capable model across its supported generation and editing tasks, and use all 150 generation outputs for each generation-only model.
Unsupported tasks and execution failures are excluded.
Two annotators independently evaluate the selected outputs.
Neither participated in benchmark construction or evaluation-question review.
Model names and VLM judge scores are hidden from both annotators.
They inspect the instructions, input images and model outputs, and answer the same questions as the VLM judges using the same $0$, $0.6$ and $1$ anchors.
Sample scores follow the same aggregation procedure as the automated evaluation.

\paragraph{Agreement.}
The annotators' sample scores have a Spearman correlation of $\rho=0.904$.
We average their sample scores to form a human reference.

GPT-4o and Gemini~3.1~Pro achieve correlations of $\rho=0.779$ and $\rho=0.852$ with this reference on the same outputs.
This comparison concerns rubric-based scores rather than user preference.

\subsection{Sensitivity to Criterion-Group Weights}
\label{app:weight-sensitivity}

We reaggregate the existing GPT-4o judgments using equal criterion-group weights and by increasing or decreasing each default group weight by $25\%$, one group at a time.
Weights are renormalized over applicable groups, while question scores, criterion applicability and sample inclusion remain fixed.
We recompute $Q$ and $C_{0.6}$ and assign tiers using the final thresholds in Tab.~\ref{tab:quality_tiers}.
Criterion-group scores remain unchanged because the perturbations affect only their aggregation into sample quality.

\begin{table}[t]
\centering
\caption{
Sensitivity to criterion-group weights under GPT-4o.
Changes are relative to the default weights across 19 models.
Same tier counts unchanged assignments under the final quality thresholds.
}
\label{tab:app-weight-sensitivity}
\small
\setlength{\tabcolsep}{5pt}
\renewcommand{\arraystretch}{1.08}
\begin{tabular}{@{}lrrrr@{}}
\toprule
Setting & Mean $|\Delta Q|$ & Max. $|\Delta Q|$
& Spearman $\rho$ & Same tier \\
\midrule
Default       & 0.0000 & 0.0000 & 1.000 & 19/19 \\
Equal weights & 0.0016 & 0.0052 & 1.000 & 19/19 \\
IF $+25\%$    & 0.0009 & 0.0038 & 1.000 & 19/19 \\
IF $-25\%$    & 0.0011 & 0.0045 & 1.000 & 19/19 \\
VQ $+25\%$    & 0.0007 & 0.0029 & 1.000 & 19/19 \\
VQ $-25\%$    & 0.0009 & 0.0035 & 1.000 & 19/19 \\
DP $+25\%$    & 0.0008 & 0.0023 & 1.000 & 19/19 \\
DP $-25\%$    & 0.0009 & 0.0030 & 1.000 & 19/19 \\
\bottomrule
\end{tabular}
\end{table}

All tested settings preserve both the overall quality ranking and the quality tiers of all 19 models (Tab.~\ref{tab:app-weight-sensitivity}).
The largest absolute change in $Q$ is $0.0052$.
These results indicate stability under the tested weight perturbations.

\subsection{Sensitivity to Tier Thresholds}
\label{app:tier-sensitivity}

\paragraph{Quality thresholds.}
Starting from Tab.~\ref{tab:quality_tiers}, we shift all minimum requirements for $Q$, $C_{0.6}$ and criterion-group scores by $\delta\in\{-0.05,-0.02,0.02,0.05\}$.
The subtask quality cutoff used to compute $C_{0.6}$ remains fixed at $0.6$, and all scores remain unchanged.
Tier assignments use the stored scores before rounding to the precision displayed in the paper.
Between 15 and 19 models retain their default tiers (Tab.~\ref{tab:app-tier-sensitivity}).
Lowering thresholds by $0.05$ moves LongCat-Image-Edit, DeepGen, Lance and Ovis-U1 from Q2 to Q3; a reduction of $0.02$ moves only DeepGen.
Increasing thresholds by $0.05$ moves Nano Banana 2 and Seedream from Q4 to Q3 because their IF scores fall below the resulting $0.95$ group requirement.
All changes are limited to one tier.

\paragraph{Deployment thresholds.}
We multiply all D3 and D4 limits for size, latency and peak memory by $\alpha\in\{0.8,0.9,1.1,1.2\}$, keeping execution status fixed.
The analysis uses the six iPhone~17~Pro baseline configurations in Tab.~\ref{tab:deployment-results}.
All six models retain their tiers under a $10\%$ budget reduction or a $10$--$20\%$ increase.
A $20\%$ reduction moves Mobile-O and DreamLite from D3 to D2 because their model sizes exceed the reduced D3 storage limit of $1638.4$~MiB.
No evaluated configuration reaches D4 under any tested setting.
These results distinguish stability to aggregation weights from sensitivity to the resource and quality requirements used to define tiers.

\begin{table}[t]
\centering
\caption{
Sensitivity to tier thresholds relative to Tab.~\ref{tab:quality_tiers}.
Quality requirements are shifted by $\delta$ and deployment limits are scaled by $\alpha$.
Unchanged counts are relative to the default tiers.
}
\label{tab:app-tier-sensitivity}
\small
\setlength{\tabcolsep}{5pt}
\renewcommand{\arraystretch}{1.12}
\begin{tabular}{@{}lc>{\raggedright\arraybackslash}p{0.57\textwidth}@{}}
\toprule
Setting & Unchanged & Tier changes \\
\midrule
\multicolumn{3}{@{}l}{\textit{Q-tier: 19 models, GPT-4o}} \\
\addlinespace[2pt]

$\delta=-0.05$ & 15/19
& LongCat-Image-Edit, DeepGen 1.0, Lance-3B and Ovis-U1-3B:
$Q2\rightarrow Q3$ \\

$\delta=-0.02$ & 18/19
& DeepGen 1.0: $Q2\rightarrow Q3$ \\

$\delta=+0.02$ & 19/19 & None \\

$\delta=+0.05$ & 17/19
& Nano Banana 2 and Seedream 5.0 Pro: $Q4\rightarrow Q3$ \\

\midrule
\multicolumn{3}{@{}l}{
\textit{D-tier: six on-device models, iPhone 17 Pro}
} \\
\addlinespace[2pt]

$\alpha=0.80$ & 4/6
& Mobile-O-0.5B and DreamLite: $D3\rightarrow D2$ \\

$\alpha=0.90$ & 6/6 & None \\
$\alpha=1.10$ & 6/6 & None \\
$\alpha=1.20$ & 6/6 & None \\

\bottomrule
\end{tabular}
\end{table}

\end{document}